\documentclass{article}

\PassOptionsToPackage{numbers, compress}{natbib}
\usepackage[preprint]{neurips_2026}

\usepackage[utf8]{inputenc} % allow utf-8 input
\usepackage[T1]{fontenc}    % use 8-bit T1 fonts
\usepackage{hyperref}       % hyperlinks
\usepackage{url}            % simple URL typesetting
\usepackage{booktabs}       % professional-quality tables
\usepackage{amsfonts}       % blackboard math symbols
\usepackage{nicefrac}       % compact symbols for 1/2, etc.
\usepackage{microtype}      % microtypography
\usepackage{xcolor}         % colors
\usepackage{wrapfig} 
\usepackage{graphicx}
\usepackage[utf8]{inputenc} % allow utf-8 input
\usepackage[T1]{fontenc}    % use 8-bit T1 fonts
\usepackage{hyperref}       % hyperlinks
\usepackage{url}            % simple URL typesetting
\usepackage{booktabs}       % professional-quality tables
\usepackage{amsfonts}       % blackboard math symbols
\usepackage{nicefrac}       % compact symbols for 1/2, etc.
\usepackage{microtype}      % microtypography
\usepackage{xcolor}         % colors
\usepackage{hyperref}
\usepackage{url}
\usepackage[table]{xcolor}
\usepackage{booktabs}
\usepackage{multicol}
\usepackage[numbers]{natbib} 
\usepackage{xspace}
\usepackage{pifont}
\usepackage{amssymb}
\usepackage{bbding}
\usepackage{pifont}
\usepackage{makecell}
\usepackage{multirow}
\usepackage{wrapfig}
\usepackage{enumitem}

\usepackage{adjustbox}
\usepackage{makecell}
\usepackage{colortbl}
\usepackage{pgf}
\usepackage{threeparttable}
\usepackage{amsmath} 
\usepackage{algorithm}
\usepackage{algpseudocode}

\usepackage{graphicx, booktabs, multirow, array, amsmath, amssymb, listings, pifont, natbib}   
\usepackage[table]{xcolor}                                                                                
\definecolor{backcolour}{rgb}{0.95,0.95,0.92}  
\definecolor{codegreen}{rgb}{0,0.6,0}
\definecolor{codegray}{rgb}{0.5,0.5,0.5}
\definecolor{codepurple}{rgb}{0.58,0,0.82}

\usepackage{tcolorbox}                       
\tcbuselibrary{breakable, skins}  
\usepackage[table]{xcolor}
\usepackage{booktabs}
\usepackage{pifont}      % \ding
\usepackage{amssymb}     % \checkmark
\usepackage{makecell}    % \makecell
\usepackage{graphicx}    % \resizebox
   
\newcommand{\bench}{\textsc{GeoDrive-Bench}\xspace}
\newcommand{\method}{\textsc{DriveOPD}\xspace}
\definecolor{mydarkyellow}{RGB}{216, 214, 196}  
\definecolor{mylightyellow}{RGB}{245, 245, 240}

\newif\ifdraft\drafttrue

\title{\bench: Benchmarking Region-Specific Multimodal Reasoning in Autonomous Driving}

\author{
Yingzi Ma\textsuperscript{1},
Chaowei Xiao\textsuperscript{2},
Ming Jiang\textsuperscript{1},
\\[2pt]
\textsuperscript{1}University of Wisconsin-Madison \quad
\textsuperscript{2}Johns Hopkins University 
}

\begin{document}

\maketitle

\begin{abstract}
Vision-language models (VLMs) for autonomous driving have shown promising performance, but their ability to handle region-specific traffic rules remains underexplored, raising uncertainties about their deployment across diverse global settings. We therefore introduce \bench, a novel benchmark that enables the systematic investigation of VLMs'  geo-culturally grounded driving reasoning. We curated 5,053 human-validated multiple-choice QA pairs across six countries covering diverse driving cultures. Specifically, we emphasize four driving tasks: perception, prediction, planning, and region reasoning. Each question requires models to infer the correct driving behavior from visual evidence and local traffic conventions without explicit country labels. Beyond evaluation, we further design a distillation algorithm that injects region-specific traffic-rule knowledge into the internal representations of VLMs, enabling models to better align visual scene understanding with local driving policies. Experiments on nine state-of-the-art VLMs show substantial performance variations across geo-driving cultures for each task, while our proposed baseline models exhibit improved geo-cultural reasoning across regions. These results suggest that current VLMs still lack robust region-aware driving intelligence and highlight \bench as a diagnostic and training-oriented testbed for deployable autonomous driving foundation models. Our code, data, and models are available at \url{https://github.com/gray311/CulturalDrive-Bench}.
\end{abstract}

\begin{figure}[h]
    \centering
        \vspace{-1em}
    \includegraphics[
        width=0.9\linewidth,
        trim=0cm 0cm 0cm 0cm,
        clip
    ]{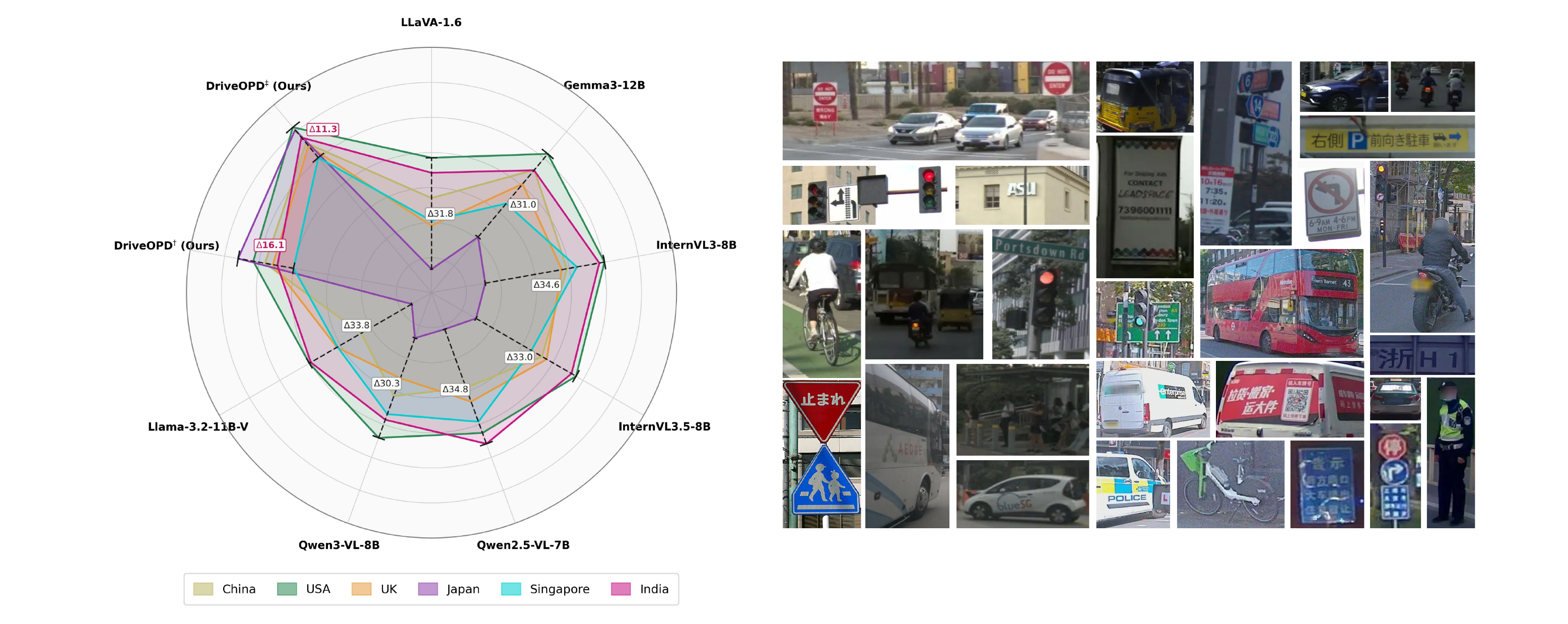}
        \vspace{-1em}
    \caption{\textbf{Overview of \bench.} \textit{Left}: radar visualization of per-country accuracy across representative VLMs, where each polygon corresponds to one country and each axis denotes a model. 
The results show that current VLMs exhibit highly imbalanced performance across country-specific scenarios, even when evaluated on the same driving tasks.\textit{Right}: region-specific visual
cues (signs, license plates, signals, vehicles) collected across the six
countries covered by \bench, which models must leverage to infer regional
context without explicit labels.}
    \label{fig:benchmark_distribution}
    \vspace{-1em}
\end{figure}

\section{Introduction}
\label{sec:intro}

Vision-language models (VLMs) have recently emerged as a promising
foundation for autonomous driving end-to-end systems, demonstrating strong
capabilities in scene understanding, behavior prediction, high-level
decision making, and low-level planning~\citep{nvidia2025alpamayor1bridgingreasoningaction, chi2025impromptu, ma2024dolphins,sima2024drivelm,xu2024drivegpt4,zhou2025autovla, zeng2025futuresightdrive,luo2025adathinkdrive,zhou2025opendrivevla}. To support the rapid progress of VLMs in this domain, researchers have developed a diverse set of driving-oriented
benchmarks~\citep{sima2024drivelm, caesar2020nuscenes, qian2024nuscenes, xie2025vlms, hao2025driveaction, xu2025wod}, covering a broad spectrum of tasks, from scene- and object-level perception to driver behavior reasoning and inference, from common driving scenarios to open-world corner cases, and from evaluating model effectiveness to assessing robustness. 

% By
% unifying visual perception with natural language reasoning, VLMs offer a
% flexible interface for tasks that were traditionally handled by modular
% pipelines, and have motivated a growing body of driving-oriented
% benchmarks~\citep{sima2024drivelm, caesar2020nuscenes, qian2024nuscenes, xie2025vlms, hao2025driveaction, xu2025wod}.
% These benchmarks have played a central role in measuring progress,
% providing standardized question--answer formats for perception,
% prediction, and planning under diverse driving conditions.

Despite remarkable progress, existing benchmarks primarily focus on common driving behavior prediction in limited countries, which % share a common blind spot: they are constructed from geographically homogeneous data, typically drawn from a specific dataset (e.g., nuScenes~\cite{caesar2020nuscenes}) collected in one or two regions such as the United States and Singapore (seeTable~\ref{tab:comparison_benchmark}). As a result, current evaluations
implicitly assume a \emph{universal} set of driving rules and
conventions, and reward models that learn perception-centric shortcuts
rather than rule-aware reasoning. In practice, however, driving is
deeply shaped by local context~\cite{li2024driving,dong2024towards,wang2025impact}. Traffic regulations, road-sign systems,
right-of-way conventions, and pedestrian behaviors differ substantially
across countries, and the \emph{same} visual scene can correspond to
\emph{different} correct actions depending on the region. We argue that evaluating driving VLMs requires a new axis of analysis:
\textbf{culturally and regionally grounded reasoning}. A competent
driving foundation model must (i)~recognize regional cues from visual context
alone, such as road signs, license plates, lane markings, and vehicle conventions; (ii)~retrieve or apply the corresponding local traffic
rules; and (iii)~integrate both into perception, prediction, and
planning decisions. Crucially, these abilities should be tested
\emph{without} explicit region labels, since a model that merely looks
up rules given a country name is not demonstrating visual--cultural
reasoning. To our knowledge, no existing benchmark jointly addresses
traffic-rule grounding, cultural reasoning, multi-country coverage, multiple task types, and
human-verified evaluation at scale.

\begin{table*}[h]
\small
\vspace{-1.5em}
\caption{\textbf{Comparison between \bench and existing autonomous driving benchmarks/datasets.}
\textit{Modality}: S.I.~= single image, M.I.~= multi-image, M.V.~= multi-view.
\textit{Tasks}: \textbf{P}erception, p\textbf{R}ediction, p\textbf{L}anning, r\textbf{E}gion understanding, \textbf{S}patial/numerical, \textbf{C}orner cases, \textbf{O}pen-loop planning.
\textit{Rule} denotes explicit grounding in local traffic regulations; \textit{Culture} denotes reasoning about region-specific driving conventions.
\textit{\#\,QA} reports the size of the test set.}
\label{tab:comparison_benchmark}
\centering

\resizebox{\textwidth}{!}{
\begin{tabular}{l|c|c|c|c|c|c|c}
\toprule
\rowcolor{mydarkyellow}
Benchmark &
Modality &
\makecell{Human\\Verif.} &
Tasks &
\makecell{Traffic\\Rule} &
\makecell{Cultural\\Reasoning} &
Countries &
\makecell{\#\,QA\\(test)} \\
\midrule
DriveMLLM~\citep{guo2024surds}              & S.I. & \checkmark & P, S         & \ding{55}  & \ding{55}  & SG, US              & 4.6K   \\
\rowcolor{mylightyellow}
NuScenes-QA~\citep{qian2024nuscenes}        & M.V. & \ding{55}  & P            & \ding{55}  & \ding{55}  & SG, US              & 83K    \\
DriveLM~\citep{sima2024drivelm}             & M.V. & \checkmark & P, Pr, Pl    & \ding{55}  & \ding{55}  & SG, US, Sim         & 15K    \\
\rowcolor{mylightyellow}
NuScenes-SpatialQA~\citep{tian2025nuscenes} & M.V. & \ding{55}  & P, S         & \ding{55}  & \ding{55}  & SG, US              & 4K     \\
CODA-LM~\citep{chen2025automated}           & S.I. & \checkmark & P, Pl, C     & \ding{55}  & \ding{55}  & CN                  & 4.8K   \\
\rowcolor{mylightyellow}
DriveLMM-o1~\citep{ishaq2025drivelmm}       & M.I. & \checkmark & P, Pr, Pl    & \ding{55}  & \ding{55}  & SG, US              & 4.6K   \\
OmniDrive~\citep{wang2025omnidrive}         & M.V. & \checkmark & P, Pr, Pl, O & \checkmark  & \ding{55}  & SG, US              & 35K    \\
\rowcolor{mylightyellow}
DriveBench~\citep{xie2025vlms}              & M.V. & \ding{55}  & P, Pr, Pl    & \ding{55}  & \ding{55}  & SG, US              & 20K    \\
DriveAction~\citep{hao2025driveaction}      & M.I. & \checkmark & P, Pl        & \ding{55}  & \ding{55}  & CN                  & 16K    \\
\rowcolor{mylightyellow}
LLaDA-AV~\citep{li2024driving}              & M.I. & \ding{55}  & O            & \checkmark & \checkmark & SG, US, CN, NL & --     \\
\midrule
\textbf{\bench} & \textbf{M.I.} & \checkmark & \textbf{P, Pr, Pl, E} & \checkmark & \checkmark & \textbf{SG, US, CN, JP, UK, IND} & \textbf{5{,}053} \\
    \bottomrule
\end{tabular}
}

\vspace{-1em}
\end{table*}

% For instance,
% turning right on a red light is generally prohibited in Japan, the
% United Kingdom, and Singapore, but may be allowed in China and the
% United States under certain conditions. A model that has only seen one
% regional distribution during training or evaluation cannot be expected
% to behave safely when deployed elsewhere, yet this cross-regional
% robustness is largely invisible to today's benchmarks.

To fill this gap, we introduce \bench, a benchmark for assessing the
geo-cultural policy awareness of VLMs in autonomous driving, together
with \method (\textbf{Dri}ve \textbf{O}n-\textbf{P}olicy
\textbf{D}istillation), a rule-conditioned self-distillation algorithm
that internalizes regional traffic knowledge directly into VLM
parameters. \bench is built from six public driving datasets spanning
six countries with distinct traffic systems---covering both left- and
right-hand traffic and both Asian and Western regulatory traditions,
and selected as the largest set of regions for which public driving
datasets with sufficient scene diversity are available---China~\citep{mao2021one},
Japan~\citep{arai2025covla}, Singapore~\citep{caesar2020nuscenes}, the
United Kingdom~\citep{marcu2023lingoqa}, India~\citep{varma2019idd},
and the United States~\citep{xu2025wod}, and is curated through a
two-stage scenario-mining pipeline that combines handbook-derived
keywords, Grounding DINO~\citep{liu2024grounding} filtering, and
VLM-based~\citep{qwen3technicalreport} semantic refinement. Each scene
is paired with multiple-choice questions across four task
categories---\emph{Perception}, \emph{Prediction}, \emph{Planning},
and \emph{Region}, the last of which explicitly diagnoses whether
region-specific knowledge has been internalized rather than merely
retrieved when visually cued---whose options are deliberately
constructed as culture-dependent distractors, so that the same question
may admit different correct answers under different regional rules
(for instance, turning right on red is generally prohibited in Japan,
the UK, and Singapore, yet permitted under most conditions in China
and the US). To ensure reliability, we apply counterfactual filtering,
calibrate an automatic verifier against human expert review at 91\%
agreement, and retain 5{,}053 human-validated QA pairs. Crucially, and
in contrast to prior work, \bench emphasizes \emph{implicit}
geographic inference: the country is never revealed to the model,
since a model that merely looks up rules given a country name is not
demonstrating visual--cultural reasoning. \method addresses this
challenge directly: a single VLM serves as both teacher (conditioned
on an \emph{anonymized} country handbook) and student
(conditioned only on the scene and question), and the student is
trained on its own on-policy rollouts to match the teacher's
rule-grounded output distribution, so that regional traffic knowledge
is absorbed into model parameters, and rule-grounded decisions emerge
at inference time without any handbook lookup.

We evaluate a broad set of state-of-the-art VLMs on \bench under three
prompting settings---direct, free-form reasoning, and rule-given (the
relevant rule supplied as context)---and benchmark \method against
them. The experiments yield three consistent findings. \textbf{First},
every open-source VLM exhibits a substantial country-level imbalance
under three prompting settings, with accuracy varying by tens of points across
regions for the same task category, and our error analysis attributes the dominant share of these errors to a \emph{Cultural Rule Gap} rather than visual misperception, showing that the bottleneck lies in grounded local knowledge, not perception. \textbf{Second}, supplying
the relevant rule at inference recovers much of this gap, yet the baseline
VLMs degrade sharply when the rule is mismatched or buried in a long
multi-country document, indicating that they treat rule prompts as
authoritative without verifying them against the scene. \textbf{Third},
\method matches or surpasses the \emph{Rule-Given} accuracy of its
base models under direct prompting, sharply reduce cross-country
variance, and remains stable under noisy or unfiltered rule contexts,
showing that culturally grounded reasoning can be partially
internalized into parameters rather than left to test-time prompting.

Our contributions are threefold. \textbf{(1)} We identify
\textbf{geo-culturally grounded driving reasoning} as a critical
missing capability for autonomous-driving VLMs, where correct decisions
require jointly grounding visual evidence and region-specific traffic
conventions. \textbf{(2)} We construct \bench, the first large-scale,
human-verified benchmark for cultural driving reasoning, with 5{,}053
multiple-choice QA pairs from six countries covering perception,
prediction, planning, and region reasoning, evaluated without exposing
country labels to the model. \textbf{(3)} We propose \method, an
on-policy rule-conditioned self-distillation algorithm that
internalizes regional traffic knowledge into VLM parameters, removing
the need for handbook retrieval at inference time. Evaluations show
that \method substantially mitigates the country-level imbalance
afflicting current VLMs and delivers markedly improved region-aware
driving reasoning.

\vspace{-1em}
\section{Related Work}
\label{sec:related}

\noindent\textbf{Vision-Language Models for End-to-End Driving.}
Recent VLM progress has motivated a shift from modular autonomous-driving pipelines toward end-to-end systems that jointly perform perception, reasoning, and action prediction~\citep{hu2023planning, jiang2023vad}. Early efforts such as DriveGPT4~\citep{xu2024drivegpt4} and LMDrive~\citep{shao2024lmdrive} showed that VLMs can interpret driving scenes and generate control signals or natural-language explanations from multi-view inputs. Subsequent work extends this paradigm in several directions: DriveVLM~\citep{tian2024drivevlm} integrates chain-of-thought reasoning into planning, OmniDrive~\citep{wang2025omnidrive} and DriveLM~\citep{sima2024drivelm} unify perception, prediction, and planning under a graph-structured VQA formulation, and vision-language-action (VLA) models such as Impromptu-VLA~\citep{chi2025impromptu}, Alpamayo-R1~\citep{nvidia2025alpamayor1bridgingreasoningaction}, and CoVLA~\citep{arai2025covla} couple visual reasoning directly with low-level action outputs. \citet{li2024driving} further explores cross-region policy adaptation by prompting LLMs with regional traffic descriptions. Despite this progress, most systems are trained and evaluated on a narrow set of geographic regions, leaving it unclear whether their reasoning transfers across different traffic rules and conventions. Our work complements this line by providing a diagnostic benchmark that specifically probes region-aware reasoning in driving VLMs. \looseness=-1

\noindent\textbf{Driving Datasets and Benchmarks for VLMs.}
Existing benchmarks evaluate VLMs in driving contexts along several axes: NuScenes-QA~\citep{qian2024nuscenes}, NuScenes-MQA~\citep{inoue2024nuscenes}, DriveLM~\citep{sima2024drivelm}, and NuInstruct~\citep{ding2024holistic} cover perception and multi-task VQA over nuScenes~\citep{caesar2020nuscenes}; DriveMLLM~\citep{guo2024surds} and NuScenes-SpatialQA~\citep{tian2025nuscenes} target spatial and numerical reasoning; DriveBench~\citep{xie2025vlms} and \citet{meng2025your} emphasize reliability and safety; and CODA-LM~\citep{chen2025automated}, DriveAction~\citep{hao2025driveaction}, and CarScenes~\citep{he2025carscenes} broaden coverage to corner cases, human-like decisions, and safety-critical semantics. However, nearly all are built on data from a small number of regions—most often the United States and Singapore—and therefore implicitly assume a uniform set of traffic rules. The closest work to ours, LLaDA-AV~\citep{li2024driving}, considers cross-regional policy adaptation but provides only a small-scale open-loop planning evaluation with explicit region labels. In contrast, \bench offers large-scale, human-verified multiple-choice evaluation across six countries and four task-relevant topics, and explicitly requires models to \emph{infer} regional context from visual cues rather than receive it as input (Table~\ref{tab:comparison_benchmark}). \looseness=-1
\section{\bench Construction}

\vspace{-0.5em}
% In this section, we describe the data construction pipeline used to collect diverse driving scenarios across countries and to build our benchmark. Our objective is to evaluate the robustness of high-level driving capabilities in existing vision-language models (VLMs).

% \ming{The overarching goal of \bench is to enable the systematic assessment of VLMs’ geo-cultural driving reasoning capabilities in autonomous driving. To achieve this goal, the key challenges in data curation include: (1)..., (2)..., etc. To address these challenges, we propose a two-stage pipeline: (1) scene collection -- briefly describe the major objective; and (2) QA pair generation -- briefly describe the major objective.}

\bench aims to systematically assess VLMs' geo-cultural driving
reasoning, since traffic rules differ substantially across
countries---particularly between Eastern and Western
regions---and a VLM's suitability as a foundation model for
vision-language-action driving systems hinge on whether its
perception, prediction, and planning remain consistent across regional
contexts~\citep{nvidia2025alpamayor1bridgingreasoningaction,
chi2025impromptu}. Curating such data poses two challenges:
(1) \emph{surfacing culturally divergent scenarios} where regional
rules genuinely change the correct action, and the region is inferable
from visual cues alone, and (2) \emph{ensuring verifiability at scale}
by grounding every QA pair in specific visual evidence and a specific
rule clause. We address them with a two-stage pipeline: \emph{scene
collection}~(Section~\ref{sec:collection}) mines visually similar but
rule-divergent scenarios across countries, and \emph{QA pair
generation}~(Section~\ref{sec:qa_generation}) converts the mined
scenarios into verifiable multiple-choice questions grounded in both
visual evidence and local rules.\looseness=-1

\subsection{Scene Collection} \label{sec:collection}

We collect data from six public datasets covering six countries: CoVLA (Japan)~\cite{arai2025covla}, ONCE (China)~\cite{mao2021one}, nuScenes (Singapore)~\cite{caesar2020nuscenes}, Waymo (United States)~\cite{xu2025wod}, LingoQA (United Kingdom)~\cite{marcu2023lingoqa}, and IDD (India)~\cite{varma2019idd}.
In particular, we focus on extracting scenes that are visually similar but require different driving decisions across countries, aiming to test whether a model can reason beyond visual similarity and account for local driving conventions. For example, at a signalized intersection, turning right on red is generally prohibited in Japan, the United Kingdom, and Singapore, but may be allowed in China and the United States depending on traffic conditions.

\begin{wrapfigure}{r}{0.58\linewidth}
    \vspace{-10pt}
    \centering
    \includegraphics[
        width=\linewidth,
        trim=20 165 20 20,
        clip
    ]{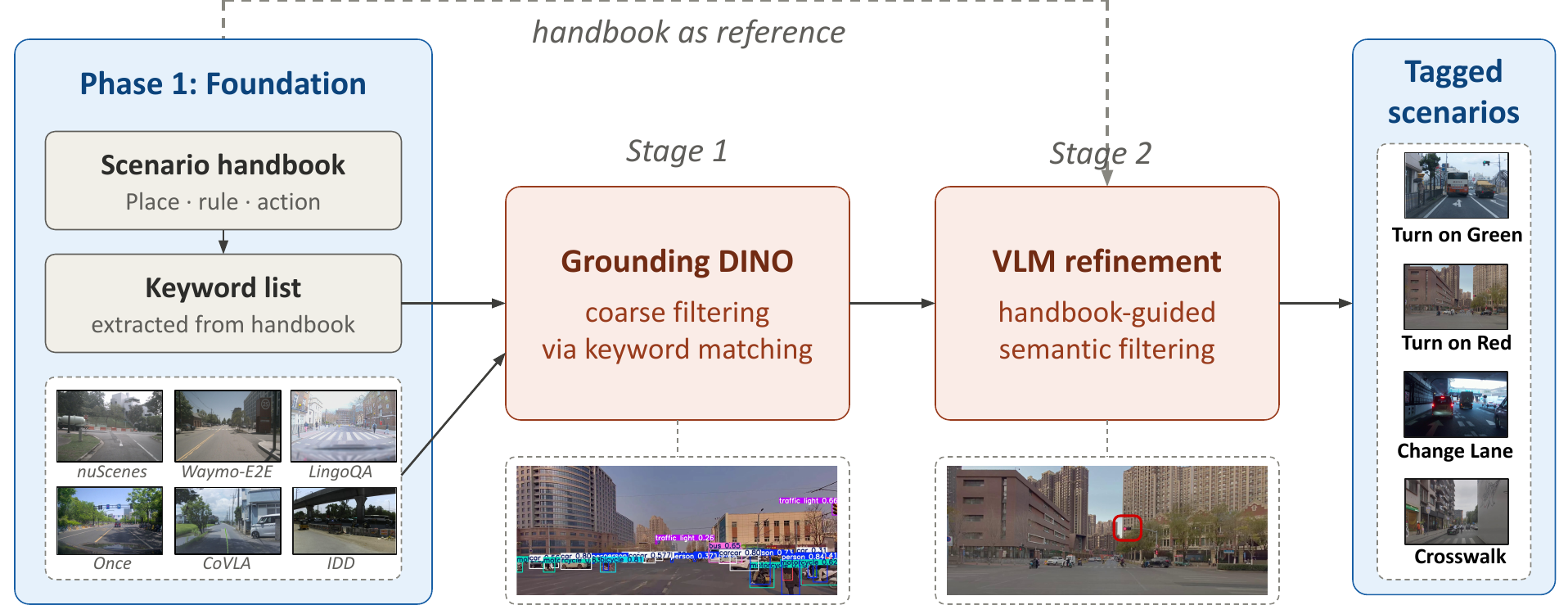}
    \vspace{-16pt}
    \caption{Overview of our scenario collection pipeline.}
    \label{fig:scenario_collection}
    \vspace{-10pt}
\end{wrapfigure}

% To capture culturally relevant driving scenarios, we manually define 13 categories of culture-specific traffic situations, drawing on crowdsourced traffic regulations from Wikipedia and prior studies on cross-country driving behavior~\cite{li2024driving}. \ming{need 1-2 more sentences to describe the details of your decisions on these 13 categories.} These categories include cases such as turning, yielding, overtaking, and stopping, where local rules may lead to different actions. The full list is provided in the appendix. \ming{Is the outcome of this manual definition a culturally relevant traffic rule handbook covering 13 categories across 6 considered countries? If so, you might want to say it explicitly.}

To make ``cultural relevance'' an operational criterion rather than an intuitive judgment, we manually define 13 categories of culture-specific traffic situations, drawing on crowdsourced traffic regulations from Wikipedia and prior studies on cross-country driving behavior~\cite{li2024driving}. A category is retained only when national traffic codes diverge along at least one of three axes: the legality of the maneuver itself (e.g., turn-on-red), the strictness of enforcement and prevailing compliance norms (e.g., pedestrian crossings, bus lanes), or the configuration of regulatory infrastructure (e.g., box junctions, HOV lanes, contraflow lanes). This filter excludes maneuvers governed only by universal driving common sense; the full list is in the appendix. The output of this step is a \emph{traffic rule handbook} that records, for each (category, country) pair, a short scene description, the governing rule, and the expected ego action, and serves as the shared supervision signal for both scene mining and QA verification (Section~\ref{sec:qa_generation}).

Mining culturally relevant scenes requires balancing recall against precision: a keyword-based detector over-retrieves, while running a large VLM over every frame is prohibitive at scale. We therefore adopt a two-stage cascade (Figure~\ref{fig:scenario_collection}). Stage~1 applies Grounding DINO~\cite{liu2024grounding}, an open-vocabulary detector chosen because it accepts free-form text queries without per-category retraining, to retain frames whose detections match at least one of a list of \emph{visual entity keywords} extracted from the handbook's scenario descriptions (e.g., \texttt{traffic light}, \texttt{stop line}, \texttt{turn arrow} for turn-on-red). Since keyword matches still admit false positives, Stage~2 re-ranks the survivors with Qwen3-VL-235B-A22B-Thinking~\cite{bai2025qwen3}. To make this re-ranking reliable, we feed the model privileged information unavailable to the benchmarked models—the country label, the matching handbook entry, and the expected action—so that the mining task is deliberately easier than the benchmark task itself. To standardize temporal granularity and limit redundancy, we sample at 2\,Hz (1\,Hz for LingoQA) and retain a single representative frame within each short temporal window.

\subsection{Culture-relevant Driving Question-Answer Generation} \label{sec:qa_generation}

Our goal is to identify VLM backbones suitable for VLA systems that operate across countries, so we focus on high-level driving-related VQA. Following prior driving benchmarks~\cite{sima2024drivelm, qian2024nuscenes, xie2025vlms}, we adopt the standard \textbf{Perception} / \textbf{Prediction} / \textbf{Planning} decomposition. We additionally introduce a \textbf{Region} task in which the model must first infer the country from incidental visual clues (signs, license plates, vehicle styles, road markings) and then answer a question about that country's traffic rule whose answer need not be directly visible, diagnosing whether region-specific knowledge has been internalized rather than only applied when visually cued. We construct QA pairs through a three-step pipeline that separates visual grounding, rule application, and quality control: (1) extracting a \emph{structured state} from each scene; (2) generating verifiable multiple-choice questions with culture-dependent distractors so that surface visual features alone cannot resolve the answer; and (3) filtering with counterfactual checks and human-calibrated verification. The details of each step are described below. \looseness=-1

\noindent\textbf{(1) Structured State Extraction.}
Generating questions directly from raw frames conflates ``what is in the image'' with ``what rule applies,'' making downstream errors hard to localize. We therefore first extract an explicit structured state recording the scene facts on which any traffic rule in our handbook could plausibly be conditioned. The schema mirrors the conditioning structure of typical traffic codes: (i) \textit{road layout} (intersection type, lane configuration, road geometry), determining which rule subset applies; (ii) \textit{traffic controls} (traffic lights, stop signs, lane markings, crosswalks), carrying the explicit regulatory signals; (iii) \textit{dynamic agents} (vehicles, pedestrians, cyclists with positions and motion states), identifying the entities whose right-of-way must be resolved; and (iv) \textit{interaction cues} (relative positions, conflict relationships, intended ego maneuver), triggering rule application. We instantiate the state with Qwen3-VL-235B-A22B-Thinking~\cite{bai2025qwen3}, conditioned on the scene frames, ground-truth bounding boxes, the country identity, and the corresponding handbook—providing all four signals together anchors the state in concrete detections and explicit rule context, which substantially reduced hallucinated state variables in our preliminary study.

\noindent\textbf{(2) Verifiable Multi-Choice QA Construction.}
We then generate QA pairs from the structured state and the country-specific traffic handbook. Our main goal is to make both questions and answers as verifiable as possible. Therefore, each QA must be grounded in observable visual evidence together with an applicable local rule. Following prior work~\cite{xie2025vlms,hao2025driveaction}, we avoid open-ended formulations that are hard to evaluate and formulate all samples as multiple-choice questions to enable standardized evaluation. This design also allows us to introduce culture-dependent distractors: under the same scene, different options may be correct in different countries. For example, an action that is legal in the United States may be prohibited in Japan. As a result, the model must jointly reason over visual evidence and local traffic rules, rather than rely on generic driving priors. In addition, the question wording is designed to avoid leaking decisive cues, so that the correct answer cannot be inferred from text alone without inspecting the scene.

\begin{figure}[t]
    \centering
    \includegraphics[
        width=1.0\linewidth,
        trim=0cm 0cm 0cm 0cm,
        clip
    ]{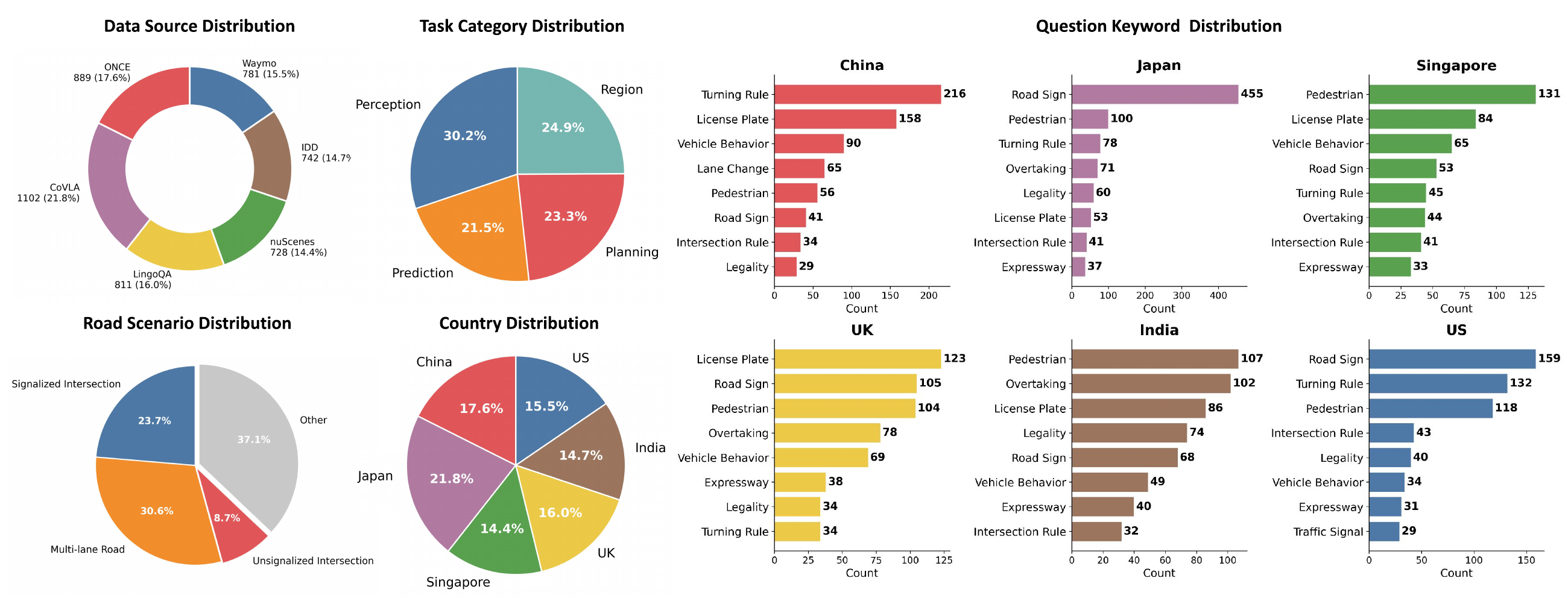}
    \vspace{-12pt}
    \caption{Distribution of \bench{} across datasets, task categories, countries, scenarios, and region-specific topics.}
    \vspace{-2em}
    \label{fig:benchmark_distribution}
\end{figure}

\noindent\textbf{(3) Counterfactual Filtering and Human Verification.}
To ensure that each sample is genuinely culture-related, we apply counterfactual verification by keeping the scene fixed while substituting the traffic rules of other countries. Starting from roughly 17K candidate QA pairs generated by the initial pipeline, this step, together with basic quality filtering, reduces the set to about 5.8K samples.
To further improve quality, we follow previous works~\cite{chi2025impromptu,meng2025your,he2025carscenes} and randomly sample 10\% of these QA pairs for expert review, and use Claude 4.6 Sonnet~\cite{anthropic2025claude46} to check the same subset. By iteratively refining the verification prompt, we achieve 91\% agreement between Claude and human experts. We then apply the calibrated verifier to the full candidate set. Among all reviewed samples, 84.5\% are marked as correct, 11.0\% as incorrect, and 4.3\% as ambiguous, with only a negligible number discarded due to parsing or API errors. The most common failure source is misjudging traffic-light or signal state (42\%), followed by wrong road-sign identification (18\%), incorrect lane count or lane-marking interpretation (14\%), and pedestrian or crosswalk confusion (11\%). The remaining errors mainly involve license-plate color, turning rules, object identification, and roundabout, parking, or overtaking cases. Finally, we retain \textbf{5,053} QA pairs in the final benchmark.

\subsection{Data Statistics}

Figure~\ref{fig:benchmark_distribution} shows the composition of
\bench{}: 5{,}053 human-verified QA pairs from six public driving
datasets spanning China, Japan, Singapore, the UK, India, and the US.
The data are broadly balanced across countries (Japan 21.8\%, others
14--18\%) and across tasks (perception 30.2\%, region 24.9\%, planning
23.3\%, prediction 21.5\%), supporting controlled evaluation across
task types and regional traffic systems. The benchmark also spans
diverse road environments---multi-lane roads, signalized and
unsignalized intersections, and other long-tail scenarios---and the
keyword distributions reveal clear country-specific patterns: Japan is
dominated by road-sign questions; India by pedestrian and overtaking
questions; the US by road signs, turning rules, and pedestrians; China
by turning rules, license plates, vehicle behavior, and lane changes;
and Singapore and the UK by pedestrian, license-plate, and road-sign
questions, indicating that \bench{} captures regionally distinctive
traffic cues for geo-culturally grounded driving evaluation.

\subsection{Culture-Aware Driving Baseline} 
\label{sec:baseline}

Section~\ref{sec:main_results} will show that providing the relevant traffic handbook substantially improves accuracy on culture-related driving questions, which raises the question of whether such region-specific knowledge can be internalized into model parameters so that the model behaves appropriately across regions without handbook prompts at inference time. Inspired by prior \textbf{on-policy self-distillation} methods~\cite{shenfeld2026self,kong2025sdpo}, we propose \method, in which a single VLM serves as both teacher and student: the teacher receives the country-specific handbook while the student does not, and the student is trained to imitate the teacher's rule-grounded responses on its own sampled trajectories. To prevent the teacher from leaking the country label as a textual shortcut, all country names in the handbook are replaced with the placeholder ``\textit{this country},'' so that the teacher's advantage comes purely from the rule content rather than from explicit regional identifiers. This keeps supervision within the model's own reachable policy space and avoids reliance on an external oracle. Formally, given visual input $x$, question $q$, and country-specific rule $r_c$, we define
\begin{equation}
y^{T} = \pi_\theta(\cdot \mid x, q, r_c), \qquad y^{S} = \pi_\theta(\cdot \mid x, q),
\end{equation}
and distill along student-sampled trajectories by minimizing
\begin{equation}
\mathcal{L}_{\text{distill}} = \mathbb{E}_{(x, q, c) \sim \mathcal{D}} \left[ \mathrm{KL}\Big(\pi_\theta(\cdot \mid x, q, r_c) \;\|\; \pi_\theta(\cdot \mid x, q)\Big) \right].
\end{equation}
Because the teacher conditions on $r_c$ while the student does not, optimization pushes the student to recover rule-grounded decisions directly from visual and linguistic context. We construct $\mathcal{D}$ jointly from all six countries in \bench, exposing the student to heterogeneous regional systems—left- vs.\ right-hand traffic, country-specific sign conventions, and different right-of-way rules—so that regional traffic knowledge is absorbed into the parameters rather than supplied as external context at test time.
\section{Experiments}

% This section empirically investigates whether current vision-language models can perform culture-aware driving reasoning on \bench{}. Specifically, we ask three questions: \textbf{(1)} Can current VLMs achieve balanced performance across countries, or do they exhibit substantial region-specific biases? \textbf{(2)} Does explicit traffic-rule grounding improve culture-aware driving QA compared with direct prompting and free-form reasoning across perception, prediction, planning, and region understanding tasks? \textbf{(3)} When performance varies across countries, what are the primary sources of failure: missing region-specific traffic knowledge, flawed or excessive reasoning, or errors in perception and visual grounding?

\subsection{Experimental Setting}

\noindent\textbf{Models.}
We evaluate a diverse set of open-source vision-language models (VLMs) on \bench{}, focusing on models that can serve as practical backbones for vision-language-action (VLA) systems. We intentionally exclude proprietary closed-source models, since their parameters, training data, and adaptation interfaces are not accessible, making them less suitable for controlled analysis of VLA-oriented fine-tuning and knowledge internalization. Our model suite covers a broad range of recent open-source VLM families and parameter scales, including \textbf{LLaVA-1.6-7B}~\citep{liu2024llavanext}, \textbf{Qwen2.5-VL-7B}~\citep{bai2025qwen2}, \textbf{Qwen3-VL-8B}~\citep{bai2025qwen3}, \textbf{InternVL3-8B}~\citep{zhu2025internvl3}, \textbf{InternVL3.5-8B}~\citep{wang2025internvl3}, \textbf{Llama-3.2-11B-Vision}~\citep{grattafiori2024llama}, and \textbf{Gemma3-12B}~\citep{gemmateam2025gemma3technicalreport}. This selection spans different vision encoders, language backbones, and training recipes, allowing us to assess whether cross-region reasoning failures are consistent across current open VLM architectures rather than tied to a single model family. For each model, we report results under three prompting settings: (1) \emph{Direct}, where the model receives only the scene and question and produces an answer in a single forward pass without intermediate reasoning; (2) \emph{Reasoning} where the model is instructed to generate a free-form chain of thought before providing a final answer; and (3) \emph{Rule-Given}, where a question-relevant rule clause from the country handbook~(see Appendix.~\ref{sec:appendix-handbook}) is included in the prompt. We further evaluate our on-policy distillation baseline, \textsc{DriveOPD}, which is built on top of open-source VLM backbones and trained with the 6-country rule-conditioned supervision described in Section~\ref{sec:baseline}.

\noindent\textbf{Implementation Details.}
All inference is performed with \texttt{vLLM}~\citep{kwon2023efficient} on
4\,$\times$\,NVIDIA A800 (80\,GB) GPUs, with the output length as 512 tokens. For \textsc{DriveOPD}, we fine-tune the student for 1 epoch with a
learning rate of $1\times10^{-5}$,
optimizing a forward-KL distillation loss against a rule-conditioned
frozen teacher (see details in Appendix.~\ref{sec:appendix-impl}).

\subsection{Main Results} \label{sec:main_results}

%% Color definitions for heatmap
\definecolor{PercBlue}{HTML}{3B7DD8}
\definecolor{PredTeal}{HTML}{1B8C65}
\definecolor{PlanAmber}{HTML}{D48A1A}
\definecolor{RegPurple}{HTML}{6B5CD0}
\definecolor{OverGray}{HTML}{7A7A76}
\begin{table*}[t]
\centering
\scriptsize
\renewcommand{\arraystretch}{1.0}
\setlength{\tabcolsep}{3pt}
\caption{\bench accuracy (\%) across three settings per model. Each task category is split by 6 countries; Background color encodes task category (\textcolor{PercBlue}{\rule{8pt}{8pt}}~Perception, \textcolor{PredTeal}{\rule{8pt}{8pt}}~Prediction, \textcolor{PlanAmber}{\rule{8pt}{8pt}}~Planning, \textcolor{RegPurple}{\rule{8pt}{8pt}}~Region, \textcolor{OverGray}{\rule{8pt}{8pt}}~Overall); within each row, darker shading indicates relatively higher accuracy. The best score across models is in \textbf{bold}.}
\label{tab:results_full_all}
\resizebox{\linewidth}{!}{
\begin{tabular}{llcccccc|cccccc|cccccc|cccccc|c}
\toprule
\multirow{2}{*}{\textbf{Model}} & \multirow{2}{*}{\textbf{Setting}} & \multicolumn{6}{c|}{\textbf{Perception}} & \multicolumn{6}{c|}{\textbf{Prediction}} & \multicolumn{6}{c|}{\textbf{Planning}} & \multicolumn{6}{c|}{\textbf{Region}} & \multirow{2}{*}{\textbf{Overall}} \\
\cmidrule(lr){3-8} \cmidrule(lr){9-14} \cmidrule(lr){15-20} \cmidrule(lr){21-26}
 &  & \textbf{CN} & \textbf{US} & \textbf{UK} & \textbf{JP} & \textbf{SG} & \textbf{IND} & \textbf{CN} & \textbf{US} & \textbf{UK} & \textbf{JP} & \textbf{SG} & \textbf{IND} & \textbf{CN} & \textbf{US} & \textbf{UK} & \textbf{JP} & \textbf{SG} & \textbf{IND} & \textbf{CN} & \textbf{US} & \textbf{UK} & \textbf{JP} & \textbf{SG} & \textbf{IND} & \\
\midrule
\multirow{3}{*}{LLaVA-1.6-7B} & Direct & \cellcolor{PercBlue!18} 24.7 & \cellcolor{PercBlue!41} 64.5 & \cellcolor{PercBlue!6} 4.3 & \cellcolor{PercBlue!5} 2.0 & \cellcolor{PercBlue!26} 37.6 & \cellcolor{PercBlue!41} 63.9 & \cellcolor{PredTeal!29} 44.1 & \cellcolor{PredTeal!33} 50.0 & \cellcolor{PredTeal!28} 42.2 & \cellcolor{PredTeal!16} 21.4 & \cellcolor{PredTeal!35} 52.7 & \cellcolor{PredTeal!37} 57.4 & \cellcolor{PlanAmber!50} 78.2 & \cellcolor{PlanAmber!36} 55.3 & \cellcolor{PlanAmber!50} 78.1 & \cellcolor{PlanAmber!54} 85.0 & \cellcolor{PlanAmber!55} 86.5 & \cellcolor{PlanAmber!44} 69.5 & \cellcolor{RegPurple!17} 22.4 & \cellcolor{RegPurple!24} 34.3 & \cellcolor{RegPurple!20} 27.6 & \cellcolor{RegPurple!20} 28.6 & \cellcolor{RegPurple!18} 24.8 & \cellcolor{RegPurple!24} 35.7 & \cellcolor{OverGray!28} 42.2 \\
 & Reasoning & \cellcolor{PercBlue!17} 23.2 & \cellcolor{PercBlue!50} 66.0 & \cellcolor{PercBlue!20} 26.5 & \cellcolor{PercBlue!5} 6.7 & \cellcolor{PercBlue!27} 36.6 & \cellcolor{PercBlue!44} 59.0 & \cellcolor{PredTeal!42} 55.9 & \cellcolor{PredTeal!48} 63.4 & \cellcolor{PredTeal!38} 50.9 & \cellcolor{PredTeal!23} 31.4 & \cellcolor{PredTeal!46} 60.9 & \cellcolor{PredTeal!48} 64.2 & \cellcolor{PlanAmber!52} 69.3 & \cellcolor{PlanAmber!53} 69.7 & \cellcolor{PlanAmber!45} 59.4 & \cellcolor{PlanAmber!55} 72.2 & \cellcolor{PlanAmber!42} 55.2 & \cellcolor{PlanAmber!44} 58.9 & \cellcolor{RegPurple!21} 28.6 & \cellcolor{RegPurple!28} 37.1 & \cellcolor{RegPurple!20} 27.1 & \cellcolor{RegPurple!21} 28.6 & \cellcolor{RegPurple!12} 16.2 & \cellcolor{RegPurple!28} 37.6 & \cellcolor{OverGray!32} 43.2 \\
 & Rule-Given & \cellcolor{PercBlue!5} 26.8 & \cellcolor{PercBlue!35} 60.9 & \cellcolor{PercBlue!11} 34.4 & \cellcolor{PercBlue!23} 47.1 & \cellcolor{PercBlue!16} 39.8 & \cellcolor{PercBlue!36} 61.5 & \cellcolor{PredTeal!39} 65.6 & \cellcolor{PredTeal!47} 74.2 & \cellcolor{PredTeal!35} 60.2 & \cellcolor{PredTeal!53} 81.0 & \cellcolor{PredTeal!41} 67.5 & \cellcolor{PredTeal!34} 59.1 & \cellcolor{PlanAmber!49} 76.6 & \cellcolor{PlanAmber!55} 82.4 & \cellcolor{PlanAmber!40} 66.3 & \cellcolor{PlanAmber!52} 79.1 & \cellcolor{PlanAmber!46} 73.0 & \cellcolor{PlanAmber!30} 55.0 & \cellcolor{RegPurple!19} 43.3 & \cellcolor{RegPurple!34} 59.5 & \cellcolor{RegPurple!19} 43.3 & \cellcolor{RegPurple!24} 49.0 & \cellcolor{RegPurple!15} 38.6 & \cellcolor{RegPurple!22} 46.7 & \cellcolor{OverGray!32} 57.1 \\
\midrule
\multirow{3}{*}{Llama-3.2-11B-V} & Direct & \cellcolor{PercBlue!22} 43.2 & \cellcolor{PercBlue!55} 81.7 & \cellcolor{PercBlue!14} 33.6 & \cellcolor{PercBlue!5} 22.4 & \cellcolor{PercBlue!29} 51.6 & \cellcolor{PercBlue!47} 72.7 & \cellcolor{PredTeal!15} 34.4 & \cellcolor{PredTeal!33} 55.9 & \cellcolor{PredTeal!16} 36.6 & \cellcolor{PredTeal!13} 32.9 & \cellcolor{PredTeal!36} 59.2 & \cellcolor{PredTeal!34} 56.8 & \cellcolor{PlanAmber!12} 31.0 & \cellcolor{PlanAmber!48} 73.9 & \cellcolor{PlanAmber!41} 65.8 & \cellcolor{PlanAmber!42} 67.4 & \cellcolor{PlanAmber!44} 69.3 & \cellcolor{PlanAmber!44} 69.5 & \cellcolor{RegPurple!10} 29.0 & \cellcolor{RegPurple!17} 37.6 & \cellcolor{RegPurple!19} 40.0 & \cellcolor{RegPurple!13} 32.4 & \cellcolor{RegPurple!5} 23.3 & \cellcolor{RegPurple!18} 38.1 & \cellcolor{OverGray!24} 45.5 \\
 & Reasoning & \cellcolor{PercBlue!28} 42.6 & \cellcolor{PercBlue!55} 83.8 & \cellcolor{PercBlue!32} 49.4 & \cellcolor{PercBlue!5} 6.5 & \cellcolor{PercBlue!37} 57.0 & \cellcolor{PercBlue!46} 71.2 & \cellcolor{PredTeal!36} 55.4 & \cellcolor{PredTeal!36} 55.4 & \cellcolor{PredTeal!36} 55.3 & \cellcolor{PredTeal!22} 33.3 & \cellcolor{PredTeal!36} 55.6 & \cellcolor{PredTeal!38} 58.5 & \cellcolor{PlanAmber!28} 43.2 & \cellcolor{PlanAmber!42} 64.4 & \cellcolor{PlanAmber!45} 68.4 & \cellcolor{PlanAmber!47} 72.2 & \cellcolor{PlanAmber!43} 66.3 & \cellcolor{PlanAmber!47} 72.2 & \cellcolor{RegPurple!22} 32.9 & \cellcolor{RegPurple!26} 39.0 & \cellcolor{RegPurple!23} 34.8 & \cellcolor{RegPurple!17} 26.2 & \cellcolor{RegPurple!20} 30.0 & \cellcolor{RegPurple!27} 41.0 & \cellcolor{OverGray!31} 47.0 \\
 & Rule-Given & \cellcolor{PercBlue!17} 45.8 & \cellcolor{PercBlue!55} 82.2 & \cellcolor{PercBlue!19} 48.6 & \cellcolor{PercBlue!51} 79.0 & \cellcolor{PercBlue!38} 66.7 & \cellcolor{PercBlue!47} 75.1 & \cellcolor{PredTeal!39} 67.7 & \cellcolor{PredTeal!46} 73.7 & \cellcolor{PredTeal!34} 62.1 & \cellcolor{PredTeal!52} 80.0 & \cellcolor{PredTeal!33} 61.5 & \cellcolor{PredTeal!39} 67.0 & \cellcolor{PlanAmber!45} 73.3 & \cellcolor{PlanAmber!50} 78.2 & \cellcolor{PlanAmber!36} 64.2 & \cellcolor{PlanAmber!46} 73.8 & \cellcolor{PlanAmber!51} 78.5 & \cellcolor{PlanAmber!32} 60.3 & \cellcolor{RegPurple!14} 43.3 & \cellcolor{RegPurple!24} 52.9 & \cellcolor{RegPurple!19} 48.6 & \cellcolor{RegPurple!5} 34.3 & \cellcolor{RegPurple!15} 44.8 & \cellcolor{RegPurple!21} 50.0 & \cellcolor{OverGray!35} 63.6 \\
\midrule
\multirow{3}{*}{Qwen2.5-VL-7B} & Direct & \cellcolor{PercBlue!41} 74.7 & \cellcolor{PercBlue!55} \textbf{97.0} & \cellcolor{PercBlue!47} 85.0 & \cellcolor{PercBlue!5} 16.4 & \cellcolor{PercBlue!43} 79.0 & \cellcolor{PercBlue!50} 90.2 & \cellcolor{PredTeal!33} 62.9 & \cellcolor{PredTeal!36} 66.7 & \cellcolor{PredTeal!28} 54.0 & \cellcolor{PredTeal!22} 45.2 & \cellcolor{PredTeal!29} 56.2 & \cellcolor{PredTeal!36} 67.6 & \cellcolor{PlanAmber!41} 75.2 & \cellcolor{PlanAmber!36} 67.0 & \cellcolor{PlanAmber!35} 65.8 & \cellcolor{PlanAmber!42} 77.5 & \cellcolor{PlanAmber!41} 75.5 & \cellcolor{PlanAmber!42} 76.2 & \cellcolor{RegPurple!19} 40.5 & \cellcolor{RegPurple!24} 47.1 & \cellcolor{RegPurple!13} 30.0 & \cellcolor{RegPurple!11} 26.7 & \cellcolor{RegPurple!13} 30.5 & \cellcolor{RegPurple!24} 48.1 & \cellcolor{OverGray!30} 57.9 \\
 & Reasoning & \cellcolor{PercBlue!35} 63.2 & \cellcolor{PercBlue!55} 92.9 & \cellcolor{PercBlue!34} 62.1 & \cellcolor{PercBlue!5} 18.6 & \cellcolor{PercBlue!40} 72.0 & \cellcolor{PercBlue!50} 86.3 & \cellcolor{PredTeal!28} 53.8 & \cellcolor{PredTeal!30} 56.5 & \cellcolor{PredTeal!31} 57.8 & \cellcolor{PredTeal!15} 33.8 & \cellcolor{PredTeal!33} 60.4 & \cellcolor{PredTeal!32} 59.7 & \cellcolor{PlanAmber!19} 39.9 & \cellcolor{PlanAmber!37} 67.6 & \cellcolor{PlanAmber!35} 64.2 & \cellcolor{PlanAmber!40} 71.7 & \cellcolor{PlanAmber!44} 76.7 & \cellcolor{PlanAmber!43} 76.2 & \cellcolor{RegPurple!22} 44.8 & \cellcolor{RegPurple!16} 35.2 & \cellcolor{RegPurple!11} 29.0 & \cellcolor{RegPurple!7} 21.9 & \cellcolor{RegPurple!14} 33.3 & \cellcolor{RegPurple!21} 43.8 & \cellcolor{OverGray!27} 51.8 \\
 & Rule-Given & \cellcolor{PercBlue!27} 64.7 & \cellcolor{PercBlue!52} 90.4 & \cellcolor{PercBlue!45} 83.4 & \cellcolor{PercBlue!44} 82.4 & \cellcolor{PercBlue!38} 75.8 & \cellcolor{PercBlue!55} \textbf{93.2} & \cellcolor{PredTeal!32} 70.4 & \cellcolor{PredTeal!34} 72.0 & \cellcolor{PredTeal!26} 64.0 & \cellcolor{PredTeal!42} 80.0 & \cellcolor{PredTeal!31} 68.6 & \cellcolor{PredTeal!27} 64.8 & \cellcolor{PlanAmber!42} 79.9 & \cellcolor{PlanAmber!41} 79.8 & \cellcolor{PlanAmber!28} 66.3 & \cellcolor{PlanAmber!43} 81.8 & \cellcolor{PlanAmber!34} 71.8 & \cellcolor{PlanAmber!16} 53.6 & \cellcolor{RegPurple!18} 56.2 & \cellcolor{RegPurple!22} 59.5 & \cellcolor{RegPurple!12} 49.5 & \cellcolor{RegPurple!5} 41.9 & \cellcolor{RegPurple!13} 51.0 & \cellcolor{RegPurple!21} 58.6 & \cellcolor{OverGray!32} 70.3 \\
\midrule
\multirow{3}{*}{Qwen3-VL-8B} & Direct & \cellcolor{PercBlue!36} 65.8 & \cellcolor{PercBlue!55} 96.4 & \cellcolor{PercBlue!47} 85.0 & \cellcolor{PercBlue!5} 15.6 & \cellcolor{PercBlue!39} 72.0 & \cellcolor{PercBlue!48} 85.9 & \cellcolor{PredTeal!39} 71.0 & \cellcolor{PredTeal!34} 62.9 & \cellcolor{PredTeal!38} 69.6 & \cellcolor{PredTeal!25} 49.0 & \cellcolor{PredTeal!40} 72.2 & \cellcolor{PredTeal!43} 78.4 & \cellcolor{PlanAmber!41} 74.3 & \cellcolor{PlanAmber!40} 72.9 & \cellcolor{PlanAmber!34} 63.1 & \cellcolor{PlanAmber!47} 84.0 & \cellcolor{PlanAmber!35} 65.6 & \cellcolor{PlanAmber!30} 57.0 & \cellcolor{RegPurple!22} 44.3 & \cellcolor{RegPurple!30} 56.2 & \cellcolor{RegPurple!14} 30.5 & \cellcolor{RegPurple!13} 30.0 & \cellcolor{RegPurple!14} 31.0 & \cellcolor{RegPurple!24} 46.7 & \cellcolor{OverGray!31} 58.8 \\
 & Reasoning & \cellcolor{PercBlue!29} 54.2 & \cellcolor{PercBlue!55} 95.4 & \cellcolor{PercBlue!33} 60.9 & \cellcolor{PercBlue!5} 15.2 & \cellcolor{PercBlue!36} 66.1 & \cellcolor{PercBlue!50} 88.3 & \cellcolor{PredTeal!29} 54.8 & \cellcolor{PredTeal!29} 53.8 & \cellcolor{PredTeal!30} 56.5 & \cellcolor{PredTeal!20} 40.0 & \cellcolor{PredTeal!36} 65.7 & \cellcolor{PredTeal!39} 69.9 & \cellcolor{PlanAmber!28} 53.5 & \cellcolor{PlanAmber!41} 74.5 & \cellcolor{PlanAmber!17} 35.8 & \cellcolor{PlanAmber!44} 78.1 & \cellcolor{PlanAmber!41} 73.0 & \cellcolor{PlanAmber!19} 39.1 & \cellcolor{RegPurple!22} 42.9 & \cellcolor{RegPurple!17} 34.8 & \cellcolor{RegPurple!13} 29.0 & \cellcolor{RegPurple!15} 31.9 & \cellcolor{RegPurple!13} 29.0 & \cellcolor{RegPurple!16} 34.3 & \cellcolor{OverGray!27} 50.5 \\
 & Rule-Given & \cellcolor{PercBlue!9} 56.3 & \cellcolor{PercBlue!55} 95.9 & \cellcolor{PercBlue!38} 81.0 & \cellcolor{PercBlue!53} 94.5 & \cellcolor{PercBlue!32} 76.3 & \cellcolor{PercBlue!49} 90.7 & \cellcolor{PredTeal!32} \textbf{75.8} & \cellcolor{PredTeal!37} 80.6 & \cellcolor{PredTeal!25} \textbf{70.2} & \cellcolor{PredTeal!41} 84.3 & \cellcolor{PredTeal!34} 77.5 & \cellcolor{PredTeal!38} 81.2 & \cellcolor{PlanAmber!40} 83.2 & \cellcolor{PlanAmber!44} 86.7 & \cellcolor{PlanAmber!32} 76.5 & \cellcolor{PlanAmber!45} 87.7 & \cellcolor{PlanAmber!43} 85.9 & \cellcolor{PlanAmber!36} 79.5 & \cellcolor{RegPurple!22} 67.1 & \cellcolor{RegPurple!25} 70.0 & \cellcolor{RegPurple!20} 65.7 & \cellcolor{RegPurple!6} 52.9 & \cellcolor{RegPurple!5} 51.9 & \cellcolor{RegPurple!23} 68.6 & \cellcolor{OverGray!34} 77.7 \\
\midrule
\multirow{3}{*}{Gemma3-12B} & Direct & \cellcolor{PercBlue!15} 48.4 & \cellcolor{PercBlue!37} 71.1 & \cellcolor{PercBlue!6} 39.5 & \cellcolor{PercBlue!14} 46.9 & \cellcolor{PercBlue!27} 60.2 & \cellcolor{PercBlue!54} 88.3 & \cellcolor{PredTeal!26} 59.7 & \cellcolor{PredTeal!27} 60.2 & \cellcolor{PredTeal!21} 54.0 & \cellcolor{PredTeal!8} 41.4 & \cellcolor{PredTeal!22} 55.6 & \cellcolor{PredTeal!40} 73.3 & \cellcolor{PlanAmber!55} 88.4 & \cellcolor{PlanAmber!52} 86.2 & \cellcolor{PlanAmber!41} 74.3 & \cellcolor{PlanAmber!49} 82.9 & \cellcolor{PlanAmber!42} 75.5 & \cellcolor{PlanAmber!34} 67.5 & \cellcolor{RegPurple!19} 51.9 & \cellcolor{RegPurple!34} 67.6 & \cellcolor{RegPurple!18} 51.0 & \cellcolor{RegPurple!5} 37.6 & \cellcolor{RegPurple!6} 39.5 & \cellcolor{RegPurple!11} 43.8 & \cellcolor{OverGray!27} 60.1 \\
 & Reasoning & \cellcolor{PercBlue!33} 62.6 & \cellcolor{PercBlue!50} 82.2 & \cellcolor{PercBlue!40} 70.0 & \cellcolor{PercBlue!12} 37.0 & \cellcolor{PercBlue!23} 50.0 & \cellcolor{PercBlue!52} 85.4 & \cellcolor{PredTeal!28} 55.9 & \cellcolor{PredTeal!33} 61.8 & \cellcolor{PredTeal!27} 55.3 & \cellcolor{PredTeal!9} 32.9 & \cellcolor{PredTeal!33} 61.5 & \cellcolor{PredTeal!35} 64.8 & \cellcolor{PlanAmber!55} 88.1 & \cellcolor{PlanAmber!51} 83.5 & \cellcolor{PlanAmber!29} 56.7 & \cellcolor{PlanAmber!43} 74.3 & \cellcolor{PlanAmber!41} 71.8 & \cellcolor{PlanAmber!32} 60.3 & \cellcolor{RegPurple!19} 44.8 & \cellcolor{RegPurple!31} 60.0 & \cellcolor{RegPurple!29} 56.7 & \cellcolor{RegPurple!5} 27.6 & \cellcolor{RegPurple!10} 34.8 & \cellcolor{RegPurple!24} 51.0 & \cellcolor{OverGray!30} 58.5 \\
 & Rule-Given & \cellcolor{PercBlue!15} 55.8 & \cellcolor{PercBlue!33} 72.6 & \cellcolor{PercBlue!5} 46.6 & \cellcolor{PercBlue!53} 91.3 & \cellcolor{PercBlue!16} 57.0 & \cellcolor{PercBlue!47} 85.9 & \cellcolor{PredTeal!33} 73.1 & \cellcolor{PredTeal!34} 73.7 & \cellcolor{PredTeal!14} 55.3 & \cellcolor{PredTeal!45} 83.3 & \cellcolor{PredTeal!30} 70.4 & \cellcolor{PredTeal!39} 77.8 & \cellcolor{PlanAmber!55} 92.4 & \cellcolor{PlanAmber!51} 88.8 & \cellcolor{PlanAmber!35} 74.3 & \cellcolor{PlanAmber!52} \textbf{89.8} & \cellcolor{PlanAmber!52} \textbf{90.2} & \cellcolor{PlanAmber!45} 83.4 & \cellcolor{RegPurple!22} 62.4 & \cellcolor{RegPurple!22} 62.4 & \cellcolor{RegPurple!25} 65.2 & \cellcolor{RegPurple!5} 46.7 & \cellcolor{RegPurple!11} 52.4 & \cellcolor{RegPurple!16} 56.7 & \cellcolor{OverGray!32} 72.2 \\
\midrule
\multirow{3}{*}{InternVL3.5-8B} & Direct & \cellcolor{PercBlue!46} 80.5 & \cellcolor{PercBlue!55} 91.9 & \cellcolor{PercBlue!44} 77.1 & \cellcolor{PercBlue!10} 31.9 & \cellcolor{PercBlue!43} 76.3 & \cellcolor{PercBlue!48} 82.4 & \cellcolor{PredTeal!38} 69.9 & \cellcolor{PredTeal!36} 66.7 & \cellcolor{PredTeal!38} 68.9 & \cellcolor{PredTeal!19} 43.3 & \cellcolor{PredTeal!39} 70.4 & \cellcolor{PredTeal!42} 75.0 & \cellcolor{PlanAmber!29} 57.4 & \cellcolor{PlanAmber!41} 72.9 & \cellcolor{PlanAmber!42} 74.3 & \cellcolor{PlanAmber!52} 88.8 & \cellcolor{PlanAmber!41} 73.0 & \cellcolor{PlanAmber!44} 77.5 & \cellcolor{RegPurple!14} 36.2 & \cellcolor{RegPurple!24} 51.0 & \cellcolor{RegPurple!19} 43.8 & \cellcolor{RegPurple!10} 31.9 & \cellcolor{RegPurple!5} 23.8 & \cellcolor{RegPurple!15} 37.6 & \cellcolor{OverGray!31} 59.9 \\
 & Reasoning & \cellcolor{PercBlue!45} 77.4 & \cellcolor{PercBlue!55} 90.9 & \cellcolor{PercBlue!32} 59.3 & \cellcolor{PercBlue!5} 22.0 & \cellcolor{PercBlue!38} 68.3 & \cellcolor{PercBlue!47} 81.0 & \cellcolor{PredTeal!28} 53.8 & \cellcolor{PredTeal!28} 53.8 & \cellcolor{PredTeal!35} 64.6 & \cellcolor{PredTeal!12} 32.4 & \cellcolor{PredTeal!30} 56.8 & \cellcolor{PredTeal!38} 67.6 & \cellcolor{PlanAmber!28} 54.8 & \cellcolor{PlanAmber!42} 73.9 & \cellcolor{PlanAmber!39} 69.0 & \cellcolor{PlanAmber!45} 78.1 & \cellcolor{PlanAmber!39} 69.9 & \cellcolor{PlanAmber!40} 71.5 & \cellcolor{RegPurple!19} 42.4 & \cellcolor{RegPurple!27} 52.9 & \cellcolor{RegPurple!19} 41.4 & \cellcolor{RegPurple!9} 28.1 & \cellcolor{RegPurple!5} 23.3 & \cellcolor{RegPurple!22} 46.7 & \cellcolor{OverGray!28} 54.6 \\
 & Rule-Given & \cellcolor{PercBlue!36} 77.4 & \cellcolor{PercBlue!53} 93.4 & \cellcolor{PercBlue!33} 74.7 & \cellcolor{PercBlue!55} 94.5 & \cellcolor{PercBlue!26} 68.8 & \cellcolor{PercBlue!47} 87.3 & \cellcolor{PredTeal!27} 69.9 & \cellcolor{PredTeal!33} 74.7 & \cellcolor{PredTeal!23} 65.8 & \cellcolor{PredTeal!44} 84.8 & \cellcolor{PredTeal!29} 71.6 & \cellcolor{PredTeal!34} 76.1 & \cellcolor{PlanAmber!51} 91.1 & \cellcolor{PlanAmber!50} 90.4 & \cellcolor{PlanAmber!34} 75.9 & \cellcolor{PlanAmber!42} 83.4 & \cellcolor{PlanAmber!42} 82.8 & \cellcolor{PlanAmber!38} 80.1 & \cellcolor{RegPurple!16} 60.0 & \cellcolor{RegPurple!33} 75.2 & \cellcolor{RegPurple!24} \textbf{67.1} & \cellcolor{RegPurple!7} 51.9 & \cellcolor{RegPurple!5} 49.5 & \cellcolor{RegPurple!25} 67.6 & \cellcolor{OverGray!35} 76.8 \\
\midrule
\multirow{3}{*}{InternVL3-8B} & Direct & \cellcolor{PercBlue!43} 75.3 & \cellcolor{PercBlue!55} 93.4 & \cellcolor{PercBlue!48} 84.2 & \cellcolor{PercBlue!5} 17.6 & \cellcolor{PercBlue!46} 80.1 & \cellcolor{PercBlue!50} 87.3 & \cellcolor{PredTeal!37} 67.2 & \cellcolor{PredTeal!42} 73.7 & \cellcolor{PredTeal!35} 64.0 & \cellcolor{PredTeal!34} 62.9 & \cellcolor{PredTeal!39} 70.4 & \cellcolor{PredTeal!46} 80.1 & \cellcolor{PlanAmber!36} 64.7 & \cellcolor{PlanAmber!48} 83.0 & \cellcolor{PlanAmber!42} 73.8 & \cellcolor{PlanAmber!49} 84.5 & \cellcolor{PlanAmber!46} 80.4 & \cellcolor{PlanAmber!45} 78.8 & \cellcolor{RegPurple!25} 48.6 & \cellcolor{RegPurple!19} 39.0 & \cellcolor{RegPurple!15} 33.8 & \cellcolor{RegPurple!11} 27.1 & \cellcolor{RegPurple!15} 33.3 & \cellcolor{RegPurple!21} 43.3 & \cellcolor{OverGray!33} 61.0 \\
 & Reasoning & \cellcolor{PercBlue!37} 68.4 & \cellcolor{PercBlue!55} 96.4 & \cellcolor{PercBlue!38} 70.0 & \cellcolor{PercBlue!5} 15.2 & \cellcolor{PercBlue!42} 75.3 & \cellcolor{PercBlue!49} 87.3 & \cellcolor{PredTeal!32} 59.7 & \cellcolor{PredTeal!38} 69.4 & \cellcolor{PredTeal!33} 61.5 & \cellcolor{PredTeal!24} 46.2 & \cellcolor{PredTeal!38} 70.4 & \cellcolor{PredTeal!42} 75.6 & \cellcolor{PlanAmber!32} 59.7 & \cellcolor{PlanAmber!39} 71.3 & \cellcolor{PlanAmber!32} 59.4 & \cellcolor{PlanAmber!44} 78.6 & \cellcolor{PlanAmber!40} 72.4 & \cellcolor{PlanAmber!36} 66.2 & \cellcolor{RegPurple!26} 50.0 & \cellcolor{RegPurple!23} 45.2 & \cellcolor{RegPurple!19} 39.0 & \cellcolor{RegPurple!17} 34.8 & \cellcolor{RegPurple!17} 36.2 & \cellcolor{RegPurple!24} 46.7 & \cellcolor{OverGray!30} 57.4 \\
 & Rule-Given & \cellcolor{PercBlue!30} 73.7 & \cellcolor{PercBlue!54} 94.9 & \cellcolor{PercBlue!42} 84.6 & \cellcolor{PercBlue!55} 95.8 & \cellcolor{PercBlue!41} \textbf{83.3} & \cellcolor{PercBlue!44} 86.3 & \cellcolor{PredTeal!27} 71.0 & \cellcolor{PredTeal!38} 80.6 & \cellcolor{PredTeal!13} 58.4 & \cellcolor{PredTeal!44} \textbf{86.7} & \cellcolor{PredTeal!37} \textbf{79.9} & \cellcolor{PredTeal!32} 75.0 & \cellcolor{PlanAmber!41} 83.5 & \cellcolor{PlanAmber!46} 88.3 & \cellcolor{PlanAmber!33} 75.9 & \cellcolor{PlanAmber!39} 81.8 & \cellcolor{PlanAmber!39} 82.2 & \cellcolor{PlanAmber!28} 72.2 & \cellcolor{RegPurple!19} 63.8 & \cellcolor{RegPurple!19} 63.3 & \cellcolor{RegPurple!18} 62.4 & \cellcolor{RegPurple!5} 50.5 & \cellcolor{RegPurple!13} \textbf{58.6} & \cellcolor{RegPurple!24} 68.6 & \cellcolor{OverGray!34} 77.2 \\
\midrule
\midrule
\multirow{3}{*}{\method$^\dagger$} & Direct & \cellcolor{PercBlue!44} \textbf{82.1} & \cellcolor{PercBlue!54} 95.9 & \cellcolor{PercBlue!46} 85.8 & \cellcolor{PercBlue!55} \textbf{97.2} & \cellcolor{PercBlue!42} 79.6 & \cellcolor{PercBlue!49} 90.2 & \cellcolor{PredTeal!38} 74.7 & \cellcolor{PredTeal!37} 72.6 & \cellcolor{PredTeal!27} 59.0 & \cellcolor{PredTeal!38} 74.8 & \cellcolor{PredTeal!31} 65.1 & \cellcolor{PredTeal!37} 72.7 & \cellcolor{PlanAmber!51} \textbf{92.7} & \cellcolor{PlanAmber!48} 88.8 & \cellcolor{PlanAmber!39} 75.4 & \cellcolor{PlanAmber!45} 84.0 & \cellcolor{PlanAmber!45} 83.4 & \cellcolor{PlanAmber!51} \textbf{92.1} & \cellcolor{RegPurple!23} 54.3 & \cellcolor{RegPurple!33} 67.6 & \cellcolor{RegPurple!24} 54.8 & \cellcolor{RegPurple!5} 28.1 & \cellcolor{RegPurple!21} 51.0 & \cellcolor{RegPurple!28} 60.5 & \cellcolor{OverGray!39} 75.7 \\
 & Reasoning & \cellcolor{PercBlue!27} 61.1 & \cellcolor{PercBlue!39} 75.1 & \cellcolor{PercBlue!48} 86.2 & \cellcolor{PercBlue!55} 93.5 & \cellcolor{PercBlue!28} 62.4 & \cellcolor{PercBlue!44} 80.5 & \cellcolor{PredTeal!32} 67.2 & \cellcolor{PredTeal!34} 69.4 & \cellcolor{PredTeal!26} 59.0 & \cellcolor{PredTeal!37} 72.9 & \cellcolor{PredTeal!29} 63.3 & \cellcolor{PredTeal!27} 60.8 & \cellcolor{PlanAmber!47} 85.1 & \cellcolor{PlanAmber!44} 80.9 & \cellcolor{PlanAmber!26} 59.9 & \cellcolor{PlanAmber!44} 81.3 & \cellcolor{PlanAmber!38} 74.2 & \cellcolor{PlanAmber!22} 54.3 & \cellcolor{RegPurple!19} 51.4 & \cellcolor{RegPurple!28} 62.4 & \cellcolor{RegPurple!20} 52.4 & \cellcolor{RegPurple!5} 33.8 & \cellcolor{RegPurple!13} 44.3 & \cellcolor{RegPurple!26} 60.0 & \cellcolor{OverGray!33} 68.4 \\
 & Rule-Given & \cellcolor{PercBlue!24} 62.6 & \cellcolor{PercBlue!40} 78.7 & \cellcolor{PercBlue!44} 83.0 & \cellcolor{PercBlue!55} 93.7 & \cellcolor{PercBlue!24} 63.4 & \cellcolor{PercBlue!46} 84.9 & \cellcolor{PredTeal!31} 70.4 & \cellcolor{PredTeal!36} 74.7 & \cellcolor{PredTeal!20} 59.0 & \cellcolor{PredTeal!39} 77.6 & \cellcolor{PredTeal!24} 63.3 & \cellcolor{PredTeal!28} 66.5 & \cellcolor{PlanAmber!44} 83.2 & \cellcolor{PlanAmber!50} 89.4 & \cellcolor{PlanAmber!16} 54.5 & \cellcolor{PlanAmber!44} 83.4 & \cellcolor{PlanAmber!44} 83.4 & \cellcolor{PlanAmber!19} 58.3 & \cellcolor{RegPurple!18} 56.7 & \cellcolor{RegPurple!30} 69.0 & \cellcolor{RegPurple!17} 55.7 & \cellcolor{RegPurple!5} 43.3 & \cellcolor{RegPurple!18} 57.1 & \cellcolor{RegPurple!23} 62.4 & \cellcolor{OverGray!33} 71.6 \\
\midrule
\multirow{3}{*}{\method$^\ddagger$} & Direct & \cellcolor{PercBlue!25} 66.8 & \cellcolor{PercBlue!51} 90.9 & \cellcolor{PercBlue!51} 90.5 & \cellcolor{PercBlue!55} 94.1 & \cellcolor{PercBlue!34} 75.8 & \cellcolor{PercBlue!41} 81.5 & \cellcolor{PredTeal!28} 69.9 & \cellcolor{PredTeal!37} 78.0 & \cellcolor{PredTeal!25} 67.1 & \cellcolor{PredTeal!41} 81.4 & \cellcolor{PredTeal!18} 60.9 & \cellcolor{PredTeal!42} \textbf{83.0} & \cellcolor{PlanAmber!51} 90.8 & \cellcolor{PlanAmber!49} 89.4 & \cellcolor{PlanAmber!39} \textbf{79.7} & \cellcolor{PlanAmber!49} 89.3 & \cellcolor{PlanAmber!44} 84.7 & \cellcolor{PlanAmber!52} 91.4 & \cellcolor{RegPurple!22} 64.8 & \cellcolor{RegPurple!26} 68.6 & \cellcolor{RegPurple!23} 65.2 & \cellcolor{RegPurple!5} 48.6 & \cellcolor{RegPurple!8} 51.9 & \cellcolor{RegPurple!19} 61.9 & \cellcolor{OverGray!36} 77.3 \\
 & Reasoning & \cellcolor{PercBlue!24} 66.8 & \cellcolor{PercBlue!45} 85.3 & \cellcolor{PercBlue!53} \textbf{91.7} & \cellcolor{PercBlue!55} 93.3 & \cellcolor{PercBlue!36} 77.4 & \cellcolor{PercBlue!41} 82.0 & \cellcolor{PredTeal!29} 71.5 & \cellcolor{PredTeal!40} \textbf{81.2} & \cellcolor{PredTeal!25} 68.3 & \cellcolor{PredTeal!37} 78.6 & \cellcolor{PredTeal!25} 68.0 & \cellcolor{PredTeal!34} 76.1 & \cellcolor{PlanAmber!49} 88.4 & \cellcolor{PlanAmber!47} 86.7 & \cellcolor{PlanAmber!27} 69.5 & \cellcolor{PlanAmber!42} 82.9 & \cellcolor{PlanAmber!47} 87.1 & \cellcolor{PlanAmber!44} 84.1 & \cellcolor{RegPurple!27} 69.5 & \cellcolor{RegPurple!32} 73.8 & \cellcolor{RegPurple!20} 63.8 & \cellcolor{RegPurple!5} 50.5 & \cellcolor{RegPurple!7} 52.9 & \cellcolor{RegPurple!28} 70.5 & \cellcolor{OverGray!36} 77.1 \\
 & Rule-Given & \cellcolor{PercBlue!16} 65.3 & \cellcolor{PercBlue!44} 86.3 & \cellcolor{PercBlue!51} 91.3 & \cellcolor{PercBlue!55} 93.7 & \cellcolor{PercBlue!30} 75.8 & \cellcolor{PercBlue!39} 82.0 & \cellcolor{PredTeal!25} 72.0 & \cellcolor{PredTeal!36} 80.1 & \cellcolor{PredTeal!13} 62.7 & \cellcolor{PredTeal!44} 85.7 & \cellcolor{PredTeal!24} 71.0 & \cellcolor{PredTeal!33} 77.8 & \cellcolor{PlanAmber!49} 89.4 & \cellcolor{PlanAmber!50} \textbf{90.4} & \cellcolor{PlanAmber!31} 76.5 & \cellcolor{PlanAmber!44} 86.1 & \cellcolor{PlanAmber!42} 84.7 & \cellcolor{PlanAmber!39} 82.1 & \cellcolor{RegPurple!24} \textbf{71.0} & \cellcolor{RegPurple!33} \textbf{78.1} & \cellcolor{RegPurple!18} 66.7 & \cellcolor{RegPurple!8} \textbf{59.0} & \cellcolor{RegPurple!5} 56.7 & \cellcolor{RegPurple!24} \textbf{71.0} & \cellcolor{OverGray!34} \textbf{78.6} \\
\bottomrule
\end{tabular}}
\footnotesize
DriveOPD$^\dagger$ built on Qwen2.5-VL-7B and DriveOPD$^\ddagger$ built on InternVL3-8B.

\vspace{-1em}
\end{table*}
\begin{table}[t]
\centering
\small

\setlength{\tabcolsep}{4pt}
\caption{Cross-country standard deviation of overall accuracy (in \%) on \bench. For each (setting, model) pair, we report the standard deviation across six countries.}
\vspace{-8pt}
\label{tab:country_std}
\resizebox{\linewidth}{!}{
\begin{tabular}{lccccccccc}
\toprule
\textbf{Setting} & \textbf{LLaVA-1.6-7B} & \textbf{Gemma3-12B} & \textbf{InternVL3-8B} & \textbf{InternVL3.5-8B} & \textbf{Qwen2.5-VL-7B} & \textbf{Qwen3-VL-8B} & \textbf{Llama-3.2-11B-V} & \textbf{DriveOPD$^{\dagger}$} & \textbf{DriveOPD$^{\ddagger}$} \\

\midrule
Direct & 10.41 & 7.79 & 10.85 & 8.68 & 12.03 & 11.45 & 10.91 & \textbf{4.48} & 4.89 \\
Reasoning & 10.48 & 10.15 & 11.40 & 10.87 & 11.48 & 9.87 & 11.50 & 5.14 & \textbf{3.76} \\
Rule-Given & 6.13 & 6.90 & 4.04 & 5.82 & \textbf{3.47} & 4.98 & 5.73 & 5.56 & 4.58 \\
\bottomrule
\end{tabular}}
\vspace{-2em}
\end{table}

\noindent\textbf{Unbalanced region-aware reasoning.}
Table~\ref{tab:results_full_all} shows that current VLMs achieve reasonable
aggregate accuracy on \bench but are highly unbalanced across countries.
Under Direct prompting, InternVL3-8B leads open-source base models at
61.0\%, yet this average masks large regional gaps: Qwen2.5-VL-7B reaches
97.0\% on U.S. perception versus only 16.4\% on Japan, and Qwen3-VL-8B
shows a parallel 96.4\% / 15.6\% split. Table~\ref{tab:country_std}
confirms that this is systematic—cross-country standard deviations of
overall accuracy fall in the 8--12\% range for every base model under
Direct prompting. VLMs thus do not fail uniformly: they perform well in
regions closer to common pretraining priors and degrade sharply in
culturally distinctive ones, so high aggregate accuracy does not imply
robust region-aware reasoning.\looseness=-1

\noindent\textbf{Effect of prompting settings.}
Comparing Direct, Reasoning, and Rule-Given prompting reveals that the
bottleneck is grounded local traffic knowledge, not reasoning tokens.
Rule-Given prompting improves most models, especially in culturally
sensitive cases—Japan perception jumps from 16.4\% to 82.4\% for
Qwen2.5-VL-7B and from 17.6\% to 95.8\% for InternVL3-8B—indicating that
the errors are not purely visual: models often observe the relevant cue
(e.g., the inverted-triangle Japanese stop sign) but fail to associate
it with the correct local rule. In contrast, free-form Reasoning is not
reliably beneficial and frequently hurts (e.g., Qwen3-VL-8B's overall
accuracy drops from 58.8\% to 50.5\%, and its UK planning collapses from
63.1\% to 35.8\%), suggesting that unconstrained chain-of-thought
amplifies incorrect regional priors when it is not anchored to the
appropriate rule.\looseness=-1

\noindent\textbf{Effectiveness of our baselines.}
Overall, \method substantially improves region-aware reasoning without rule
snippets at inference time: \method$^\dagger$ raises Qwen2.5-VL-7B from
57.9\% to 75.7\%, and \method$^\ddagger$ raises InternVL3-8B from
61.0\% to 77.3\%. The gains concentrate on culture-dependent cases rather
than already-easy regions—Qwen2.5-VL-7B's Japan perception rises from
16.4\% to 97.2\% after \method training—and Table~\ref{tab:country_std}
shows the cross-country standard deviation drops from 8--12\% to under
5\%. Notably, \method under Direct prompting matches or surpasses the
Rule-Given performance of its base models (75.7\% vs.\ 70.3\% for the
Qwen2.5-VL-7B family, 77.3\% vs.\ 77.2\% for the InternVL3-8B family),
suggesting that rule-conditioned self-distillation internalizes
region-specific traffic knowledge into model parameters and is therefore
robust to test-time conditions where external rule prompts may be
unavailable, incomplete, or noisy.\looseness=-1

\begin{wraptable}{r}{0.45\textwidth}
\centering
\vspace{-1.8em}
\footnotesize  
\renewcommand{\arraystretch}{0.95}
\setlength{\tabcolsep}{4pt}
\caption{Robustness to image perturbation. For each base VLM, we report overall accuracy (\%) under two input conditions (\emph{Normal}: original image; \emph{No Image}: image removed) crossed with three prompting settings. \looseness=-1 }
\label{tab:image_perturbation}
\resizebox{\linewidth}{!}{
\begin{tabular}{lccc}
\toprule
\textbf{Setting} & \textbf{Direct} & \textbf{Reasoning} & \textbf{Rule-Given} \\
\midrule
\multicolumn{4}{c}{\textit{Gemma3-12B}} \\
\midrule
Normal   & \textbf{60.12} & \textbf{58.54} & \textbf{72.17} \\
No Image & 45.52$_{\color{red}-14.60}$ & 41.66$_{\color{red}-16.88}$ & 66.38$_{\color{red}-5.79}$ \\
\midrule
\multicolumn{4}{c}{\textit{InternVL3-8B}} \\
\midrule
Normal   & \textbf{61.01} & \textbf{57.37} & \textbf{77.18} \\
No Image & 48.53$_{\color{red}-12.48}$ & 44.07$_{\color{red}-13.30}$ & 65.47$_{\color{red}-11.71}$ \\
\midrule
\multicolumn{4}{c}{\textit{Qwen3-VL-8B}} \\
\midrule
Normal   & \textbf{58.82} & \textbf{50.50} & \textbf{77.66} \\
No Image & 47.16$_{\color{red}-11.66}$ & 37.01$_{\color{red}-13.49}$ & 64.60$_{\color{red}-13.06}$ \\
\bottomrule
\end{tabular}}
\vspace{-12pt}
\end{wraptable}

\noindent\textbf{Image Perturbation.}
We examine whether models rely on visual grounding or exploit language
and driving-action priors. Table~\ref{tab:image_perturbation} shows
that removing the image consistently reduces accuracy across all
models, confirming that visual evidence is important for \bench{}; yet
the \emph{No Image} setting still yields non-trivial performance,
especially under \emph{Rule-Given} prompting. Consistent with prior
observations~\citep{xie2025vlms}, models can often produce plausible
prediction and planning answers by defaulting to conservative actions
(e.g., slowing or yielding) without a detailed visual understanding.
Overall, \bench{} requires visual grounding, but also reveals that
current driving VLMs' prediction and planning performance is partially
supported by conservative action priors rather than precise
geo-cultural scene understanding.\looseness=-1

\begin{figure}[!t]
    \centering
          \vspace{-1.em}
    \includegraphics[
        width=1.0\linewidth,
        trim=0cm 0.0cm 0cm 0cm,
        clip
    ]{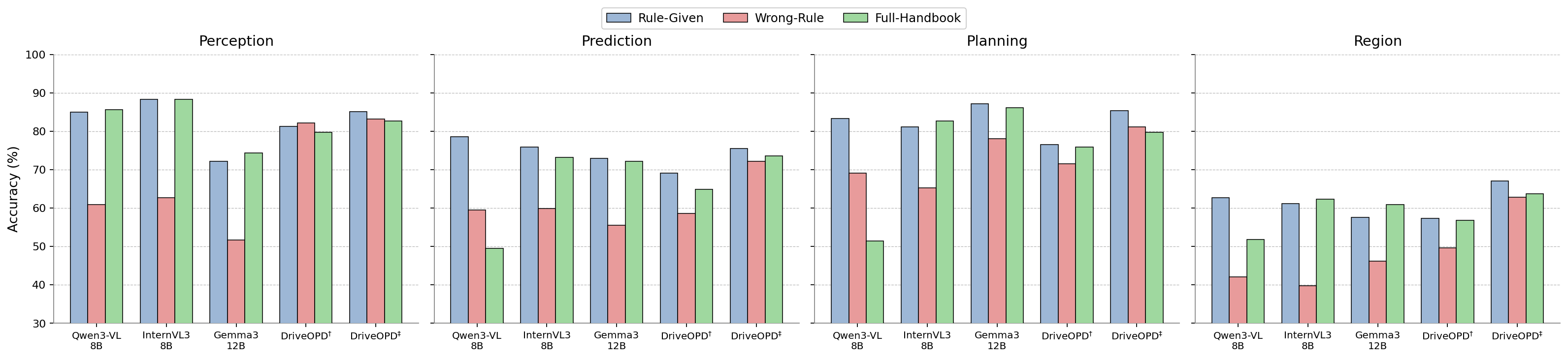}
        \vspace{-2.em}
    \caption{\textbf{Ablation on traffic rule injection} across the four
task categories. \textit{Rule-Given} provides the correct rule,
\textit{Wrong-Rule} injects a mismatched rule, and
\textit{Full-Handbook} provides the full multi-country handbook.}
\label{fig:ablation_rule}
\vspace{-2.em}
\end{figure}

\noindent\textbf{Traffic Rule Diagnosis.}
The strong gains from \emph{Rule-Given} prompting raise a follow-up question: do models actually reason from the provided rule, or simply benefit from any rule-shaped context regardless of whether it matches the scene? To disentangle these, we probe how models use traffic-rule information under three settings—\emph{Rule-Given}, \emph{Wrong-Rule}, and \emph{Full-Handbook} (Fig.~\ref{fig:ablation_rule}). Baseline VLMs drop noticeably when the correct rule is replaced with a mismatched one, especially on perception, prediction, and planning, indicating that they are sensitive to external rule prompts but cannot verify whether the provided rule matches the visual and regional context; \emph{Full-Handbook} likewise fails to consistently outperform concise \emph{Rule-Given} prompting, since the model must first retrieve the relevant country-specific rule from a long multi-country document. \method variants behave more stably: a smaller gap between \emph{Rule-Given} and \emph{Wrong-Rule} together with competitive \emph{Full-Handbook} performance suggests that rule-conditioned self-distillation internalizes geo-specific traffic knowledge into the model, rather than relying on test-time rule snippets, making it robust to noisy, mismatched, or non-preselected rule contexts.

\begin{figure}[!t]
    \centering
    \includegraphics[
        width=1.0\linewidth,
        trim=0cm 0.5cm 0cm 0cm,
        clip
    ]{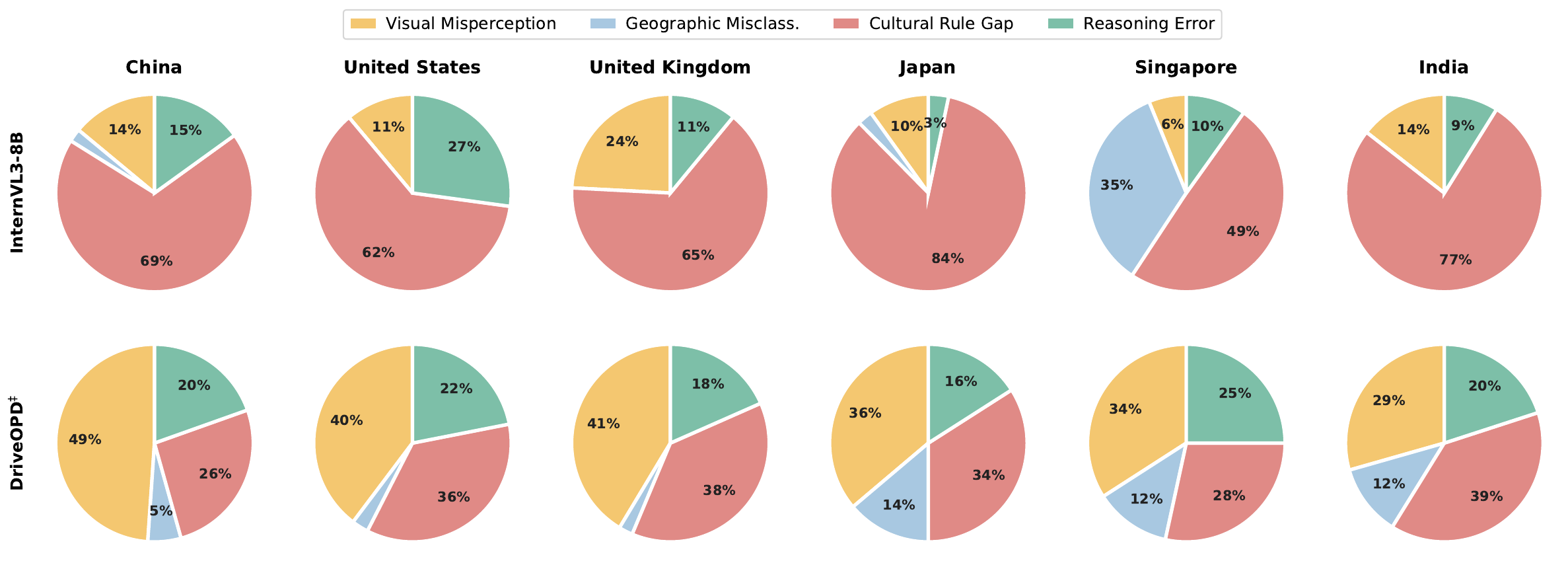}
    \vspace{-1em}
    \caption{Country-wise distribution of error types for InternVL3 and our \method$^\ddagger$ under the reasoning setting. Each pie chart shows the proportion of four major error categories within a country: Visual Misperception, Geographic Misclassification, Cultural Rule Gap, and Reasoning Error. \looseness=-1 }
\label{fig:error_analysis}
    \vspace{-1.5em}
\end{figure}

\noindent\textbf{Error Analysis.} In addition, we conducted an error analysis to identify and compare the primary vulnerabilities of existing VLMs with those of our proposed baselines. We manually categorize around 500 unique error cases from InternVL3-8B and its \method$^{\ddagger}$ variant—sampled to include at least 60 from each country and each task category—into four types: \emph{Visual Misperception}, \emph{Geographic Misclassification}, \emph{Cultural Rule Gap}, and \emph{Reasoning Error}. As shown in Fig.~\ref{fig:error_analysis}, InternVL3-8B is dominated by \emph{Cultural Rule Gap}, accounting for 84\% and 77\% of sampled errors in Japan and India, indicating that the base model often recognizes the scene but fails to apply the correct region-specific rule. After \method training, \emph{Cultural Rule Gap} drops substantially (e.g., 84\%→34\% in Japan, 69\%→26\% in China), and the remaining errors shift toward \emph{Visual Misperception} and \emph{Reasoning Error}—once regional rule knowledge is internalized, fine-grained visual understanding becomes the more prominent bottleneck. \looseness=-1

% \bench thus diagnoses not only whether models fail, but why they fail across regions.

\subsection{Case Study}
\begin{figure}[h]
    \centering
    \vspace{-1em}
    \includegraphics[
        width=1.0\linewidth,
        trim=0cm 0cm 0cm 0cm,
        clip
    ]{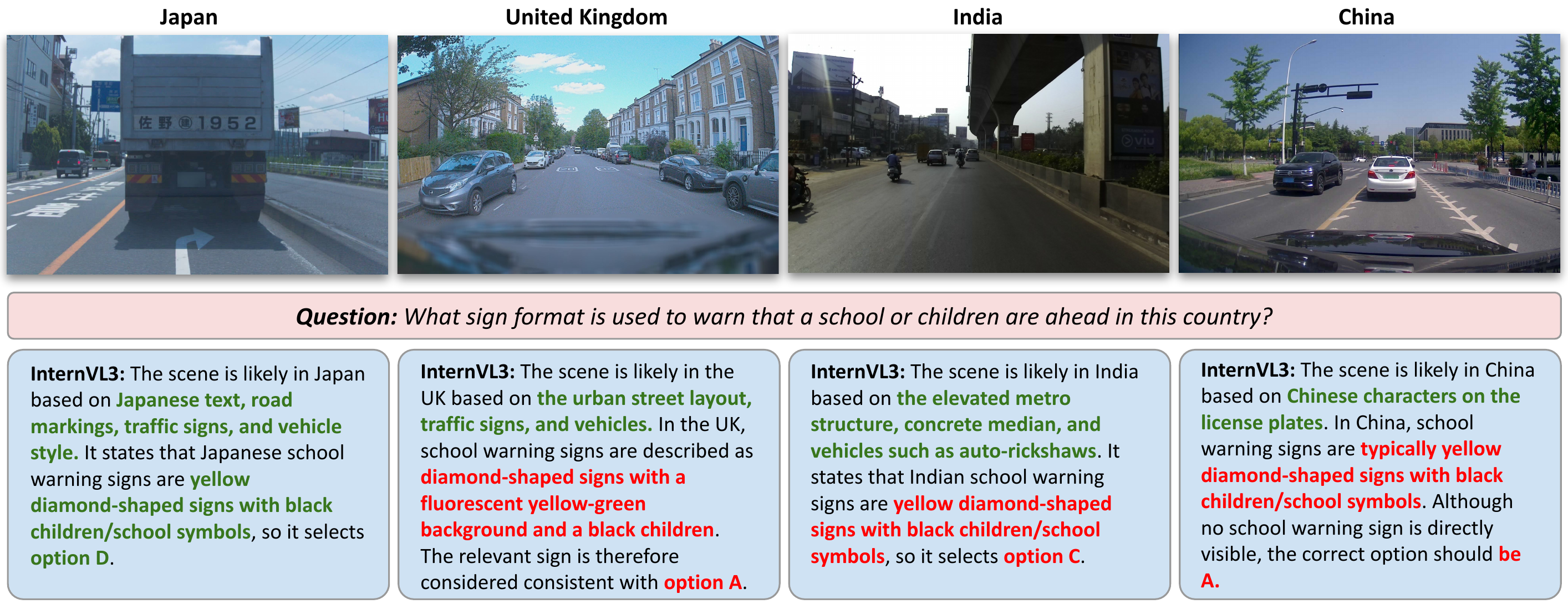 }
      \vspace{-2em}
    \caption{Illustration of the reasoning processes of InternVL3 for school-warning sign questions .}
\label{fig:case_study}
    \vspace{-1em}
\end{figure}

Figure~\ref{fig:case_study} shows a region reasoning case study of
InternVL3 on a school-warning sign question across four countries.
The model identifies each country from salient cues---Japanese text,
UK street layout, Indian auto-rickshaws, Chinese license plates---yet
defaults to a generic yellow-diamond template, failing in the UK,
India, and China. This decoupling of recognition from rule grounding
motivates explicit internalization of regional rules as in
\textsc{DriveOPD}. More examples can be found in Appendix.~\ref{sec:appendix-cases}\looseness=-1

\section{Conclusion and Limitation Discussion}

\vspace{-3mm}
In this paper, we introduce \bench{}, a novel human-verified benchmark for evaluating geo-culturally grounded multimodal reasoning in autonomous driving. Extensive experiments show that current VLMs exhibit substantial country-level imbalance and often fail to apply region-specific traffic rules, while our \method{} baseline improves region-aware reasoning by internalizing traffic-rule knowledge.
A limitation of \bench{} is that it currently covers six countries and is constrained by the availability of public driving datasets. In addition, \bench{} evaluates high-level reasoning through multiple-choice QA and does not yet directly measure low-level planning behavior across countries under region-specific rules. Future work will extend the benchmark toward cross-country low-level planning evaluation with richer traffic conditions and closed-loop settings.

% \newpage
% \input{checklist.tex}

\clearpage
{\small
\bibliographystyle{plainnat}
\bibliography{references}
}

\clearpage
\appendix
\section{Overview}
% TODO: 1-paragraph roadmap of the appendix (mirror the enumerate below).

Our appendix includes the following sections:
\begin{enumerate}
    \item \textbf{Section~\ref{sec:appendix-impl}: Additional Implementation Details.}
          Prompt templates for every evaluation setting and the full training
          recipe of \method (algorithm, data, hyperparameters, compute). \looseness=-1
    \item \textbf{Section~\ref{sec:appendix-results}: Additional Results.}
          Full image-perturbation table, per-category rule-context ablation,
          and error-type analysis on the Qwen2.5-VL family.
    \item \textbf{Section~\ref{sec:appendix-construction}: Benchmark Construction Details.}
          The 13 culture-specific traffic categories, the 20-section
          per-country traffic-rule handbook, the counterfactual verification
          protocol, and the annotation tool used by human reviewers.
    \item \textbf{Section~\ref{sec:appendix-cases}: Extended Case Studies.}
          Additional qualitative comparisons between base VLMs and
          DRIVEOPD across countries.
    \item \textbf{Section~\ref{sec:appendix-broader-impact}: Broader Impact.}
           Discussion of the broader implications of \bench.
\end{enumerate}

% =============================================================================
\section{Additional Implementation Details}
\label{sec:appendix-impl}
% =============================================================================

\subsection{\method Training Details}
\label{sec:appendix-driveopd}

We instantiate \method on top of two open-source VLM backbones, Qwen2.5-VL-7B~\cite{bai2025qwen2} and InternVL3-8B~\cite{zhu2025internvl3}, yielding the two checkpoints denoted as \method$^{\dagger}$ and \method$^{\ddagger}$ in the main paper. Both teacher and student are initialized from the \emph{same} pre-trained weights $\theta_0$. The student is conditioned only on the visual scene $x$ and the question $q$, while the teacher additionally conditions on the country-specific traffic handbook $H_c$. To prevent the country identity from leaking through textual cues, every occurrence of a country name in $H_c$ is replaced by the generic phrase ``this country'' during both training and evaluation, so that the teacher's advantage stems purely from rule \emph{content} rather than explicit regional identifiers. Only the student parameters $\theta_S$ are updated; the teacher is held fixed at $\theta_0$ and serves as a stationary rule-grounded oracle. The student matches the teacher distribution along its own on-policy rollouts via a forward KL loss (Eq.~(2) in the main paper).

This design offers three practical advantages over retrieval- or prompt-based alternatives that supply the handbook at inference time. \textbf{First, knowledge internalization.} Regional traffic conventions are absorbed directly into the student's parameters, so the model can produce rule-grounded decisions from visual and linguistic context alone, without depending on an external handbook lookup at deployment. \textbf{Second, reduced inference cost.} Country handbooks routinely span thousands of tokens; prepending them at test time inflates prompt length, latency, and memory footprint, which is particularly problematic for on-vehicle deployment under tight compute budgets. By internalizing rule knowledge during training, \method matches---and in many cases surpasses---the Rule-Given accuracy of its base model under Direct prompting, eliminating this overhead entirely. \textbf{Third, robustness to imperfect rule context.} A model that depends on prompt-time handbook injection is brittle to retrieval errors: as we show in the Wrong-Rule and Full-Handbook diagnoses (Fig.~4), baseline VLMs degrade sharply when supplied with mismatched or unfiltered multi-country rule context, because they cannot reliably verify whether a given rule actually matches the visible scene. \method, in contrast, exhibits a markedly smaller gap across these conditions, indicating that the internalized knowledge provides a stable prior that is not easily overridden by noisy or incorrect external rules.

\paragraph{Algorithm.}
Algorithm~\ref{alg:driveopd} summarizes the training loop. At each step, we (i)~draw an on-policy rollout from the current student conditioned on the visual scene and question alone, (ii)~score the same trajectory under the frozen, rule-conditioned teacher, and (iii)~minimize a token-level forward KL between the two distributions along the rollout. Since the teacher is stationary, no parameter synchronization or replay buffer is required.

\begin{algorithm}[h]
\caption{\method training loop.}
\label{alg:driveopd}
\begin{algorithmic}[1]
\Require pre-trained VLM weights $\theta_0$; training pool $\mathcal{D}=\{(x,q,H_c)\}$; epochs $E$
\State Initialize student $\theta_S \leftarrow \theta_0$ and teacher $\theta_T \leftarrow \theta_0$ (frozen).
\For{$e = 1, \ldots, E$}
    \For{each minibatch $(x, q, H_c) \in \mathcal{D}$}
        \State Sample on-policy rollout $y_S \sim \pi_{\theta_S}(\cdot \mid x, q)$.
        \State Compute teacher distribution $\pi_{\theta_T}(\cdot \mid x, q, \mathrm{anon}(H_c))$ on tokens of $y_S$.
        \State $\mathcal{L} \leftarrow \tfrac{1}{|y_S|}\sum_{t} \mathrm{KL}\!\left[\pi_{\theta_T}(\cdot \mid y_S^{<t}) \,\|\, \pi_{\theta_S}(\cdot \mid y_S^{<t})\right]$.
        \State Update $\theta_S$ by gradient descent on $\mathcal{L}$.
    \EndFor
\EndFor
\State \Return student parameters $\theta_S$.
\end{algorithmic}
\end{algorithm}

\paragraph{Training data.}
We use the entire \bench (5{,}053 items spanning six countries and four task categories) as the distillation pool $\mathcal{D}$. Importantly, gold answer labels never enter the loss: the student is supervised solely by the teacher's output distribution, so this is not label leakage in the usual sense---the teacher's advantage at each step is its access to the (anonymized) country handbook, not the ground-truth answer. For every item, the student prompt contains only the image and question under a chain-of-thought template, while the teacher prompt augments this with the full country-$c$ handbook (with country names anonymized as above). The two prompts share an identical question and option block; they differ only in the rule context. Images are resized to a uniform resolution so that image-token counts remain comparable across backbones, and we cap prompt and completion lengths to keep the optimization tractable.

\paragraph{Optimization.}
Table~\ref{tab:driveopd-hparams} summarizes the optimization settings, which are identical across both backbones. Two design choices warrant explicit discussion. First, we adopt a \textbf{frozen teacher} ($\theta_T = \theta_0$) rather than an EMA-tracked teacher: this removes a per-step parameter synchronization cost and, more importantly, provides a stationary rule-grounded target throughout training, avoiding the drift that an EMA teacher would introduce as the student moves away from the rule-conditioned distribution. Second, we use \textbf{single-rollout on-policy sampling}---each minibatch performs one fresh student rollout, with no replay buffer and no PPO-style multi-sample estimator. We found this sufficient in practice because the forward KL is well-defined per token and does not require advantage normalization or variance reduction across multiple samples.

\begin{table}[h]
    \centering
    \small
    \caption{\method training hyperparameters, shared by \method$^{\dagger}$ (Qwen2.5-VL-7B) and \method$^{\ddagger}$ (InternVL3-8B).}
    \label{tab:driveopd-hparams}
    \begin{tabular}{ll}
        \toprule
        Hyperparameter & Value \\
        \midrule
        Optimizer                       & AdamW \\
        Peak learning rate              & $1\!\times\!10^{-5}$ \\
        LR schedule                     & cosine, $10\%$ warmup \\
        Epochs                          & 1 \\
        Effective batch size            & 8 \\
        Gradient clipping               & $1.0$ \\
        Mixed precision                 & bfloat16 \\
        \midrule
        Max prompt length               & $6{,}144$ tokens \\
        Max completion length           & $512$ tokens \\
        Image resolution                & $448\!\times\!448$ \\
        \midrule
        Teacher update                  & frozen ($\theta_T = \theta_0$) \\
        On-policy rollouts per step     & 1 \\
        \bottomrule
    \end{tabular}
\end{table}

\paragraph{Compute resources.}
All \method runs are performed on $4\!\times\!$NVIDIA A100/A800 (80\,GB) GPUs with sharded data-parallel training and a colocated on-policy sampler for student rollouts. Training a 7--8B backbone for one epoch over the full 5{,}053-item pool takes approximately 6--8 hours per checkpoint, with peak per-GPU memory around 60\,GB.

\subsection{Prompt Templates}
\label{sec:appendix-prompts}

We group the prompts used in our experiments into three families, each tied to a specific set of results in the main paper. The \textit{main evaluation} family (Direct, Reasoning, Rule-Given) underlies all results in Table~2; the \textit{rule-diagnosis} family (Wrong-Rule, Full-Handbook) probes how models interact with the supplied rule context (Fig.~4); and the \textit{robustness} family (No-Image, Image-Corruption) tests the degree to which models rely on visual grounding (Table~4). Across all templates, \texttt{\{q\}} denotes the question text, \texttt{\{opts\}} the multiple-choice option block, \texttt{\{rule\}} a country-specific rule snippet drawn from the handbook section cited by the item, and \texttt{\{handbook\}} the entire country handbook. To prevent the country identity from leaking through textual cues, every occurrence of a country name in \texttt{\{rule\}} or \texttt{\{handbook\}} is replaced with the placeholder ``this country.''

\paragraph{Main evaluation settings.}
The three main settings are designed to disentangle whether failures stem from a lack of reasoning, a lack of rule knowledge, or both. The \textbf{Direct} prompt requires the model to commit to a letter answer in a single forward pass without intermediate reasoning, isolating the model's immediate priors. The \textbf{Reasoning} prompt elicits a four-step chain-of-thought---geographic inference, cultural rule recall, visual reasoning, and answer selection---without providing any external rule context, thereby probing whether the model can recover the relevant rule from its own parametric knowledge. The \textbf{Rule-Given} prompt prepends the cited country-specific rule snippet to a four-step rule-application chain, providing an upper-bound estimate of accuracy when the correct rule is available at inference time.

\noindent\textbf{Direct.}
\begin{lstlisting}[basicstyle=\ttfamily\scriptsize, breaklines=true, frame=single, backgroundcolor=\color{backcolour}]
Answer the question using only the image. Choose exactly one option.
Question: {q}
Options:
{opts}
Output only the answer letter (A/B/C/D).
\end{lstlisting}

\noindent\textbf{Reasoning.}
\begin{lstlisting}[basicstyle=\ttfamily\scriptsize, breaklines=true, frame=single, backgroundcolor=\color{backcolour}]
Step 1 -- Geographic inference: Based on road infrastructure, traffic signs, lane markings, and environmental cues in the image, identify the most likely country or region where this scene was captured.
Step 2 -- Cultural traffic rule: State the relevant traffic rule(s) for that country that apply to this question.
Step 3 -- Visual reasoning: Describe what you observe in the image that is relevant to answering the question.
Step 4 -- Answer: Choose exactly one option.
Question: {q}
Options:
{opts}
Output your step-by-step reasoning followed by the final answer letter (A/B/C/D).
\end{lstlisting}

\noindent\textbf{Rule-Given.}
\begin{lstlisting}[basicstyle=\ttfamily\scriptsize, breaklines=true, frame=single, backgroundcolor=\color{backcolour}]
Relevant traffic rule(s):
{rule}

Using the traffic rule(s) above and the image, answer the question by following these steps:
Step 1 -- Rule understanding: Identify which part of the given traffic rule is relevant to this question.
Step 2 -- Visual observation: Describe what you observe in the image that relates to the question.
Step 3 -- Rule application: Apply the traffic rule to the observed scene to reason about the correct answer.
Step 4 -- Answer: Choose exactly one option.

Question: {q}
Options:
{opts}
Output your step-by-step reasoning followed by the final answer letter (A/B/C/D).
\end{lstlisting}

\paragraph{Rule-diagnosis settings.}
The two diagnostic settings hold the Rule-Given template fixed and vary only the rule context, isolating how robustly models verify external rules against the visible scene. \textbf{Wrong-Rule} replaces \texttt{\{rule\}} with a snippet drawn deterministically (seeded per item) from a different section of the \emph{same} country's handbook---one that does not govern the depicted scenario. This tests whether the model treats the provided rule as authoritative or cross-checks it against visual evidence; an ideal model should detect the mismatch and fall back on visual grounding. \textbf{Full-Handbook} substitutes the cited snippet with the entire country handbook, forcing the model to first retrieve the relevant clause from a long multi-clause document before applying it, and thereby reflecting a more realistic deployment scenario in which the precise rule reference is not pre-selected.

\noindent\textbf{Full-Handbook.}
\begin{lstlisting}[basicstyle=\ttfamily\scriptsize, breaklines=true, frame=single, backgroundcolor=\color{backcolour}]
Traffic handbook:
{handbook}

Using the traffic handbook above and the image, answer the question by following these steps:
Step 1 -- Rule understanding: Identify which rules in the handbook are relevant to this question.
Step 2 -- Visual observation: Describe what you observe in the image that relates to the question.
Step 3 -- Rule application: Apply the relevant rule(s) to the observed scene to reason about the correct answer.
Step 4 -- Answer: Choose exactly one option.

Question: {q}
Options:
{opts}
Output your step-by-step reasoning followed by the final answer letter (A/B/C/D).
\end{lstlisting}

\paragraph{Robustness settings.}
The robustness settings examine the extent to which apparent driving competence is supported by visual evidence rather than by language and action priors. The \textbf{No-Image} prompt explicitly informs the model that no image is provided, so the answer must be derived entirely from the question text and options---if accuracy remains substantially above chance, the benchmark is partially solvable from language priors alone. The \textbf{Image-Corruption} setting reuses the Direct prompt unchanged but feeds a Gaussian-blurred image (severity comparable to ImageNet-C level~3) to the visual encoder, degrading fine-grained perceptual details such as sign text and lane markings while preserving coarse scene layout. Reasoning and Rule-Given variants of both settings reuse the corresponding text templates without modification; only the pixel input changes.

\noindent\textbf{No-Image.}
\begin{lstlisting}[basicstyle=\ttfamily\scriptsize, breaklines=true, frame=single, backgroundcolor=\color{backcolour}]
Answer the question using only your general knowledge (no image is provided). Choose exactly one option.
Question: {q}
Options:
{opts}
Output only the answer letter (A/B/C/D).
\end{lstlisting}

\section{Additional Results}
\label{sec:appendix-results}

\subsection{Image Perturbation}
\label{sec:appendix-perturb}
Table~4 in the main paper reports the No-Image robustness condition for the three strongest base VLMs under the three prompt families. We extend that analysis in Table~\ref{tab:image-perturbation-full} to a broader grid: four base VLMs (LLaVA-1.6-7B, Gemma3-12B, InternVL3-8B, and Qwen3-VL-8B), three image conditions (\emph{Normal}, the original image; \emph{No Image}, the image removed entirely; and \emph{Image Corruption}, a Gaussian blur calibrated to ImageNet-C severity~3~\cite{hendrycks2019robustness}), and the three prompt settings (\emph{Direct}, \emph{Reasoning}, \emph{Rule-Given}).

The full grid reveals three regularities that sharpen the conclusions of the main paper. First, LLaVA-1.6-7B's accuracy is essentially flat across image conditions (within roughly two points), confirming that it relies primarily on language priors and barely engages with the visual input---an important sanity check that \bench is not trivially solvable from text alone, but also a reminder that aggregate accuracy can mask a lack of visual grounding. Second, the three stronger models lose 11--17 points under No-Image yet only 1--4 points under Image-Corruption, indicating that they read coarse scene structure but rarely depend on fine-grained visual details---consistent with the failure modes diagnosed in Fig.~5, where misreading specific signs and lane markings is a recurring source of error. Third, the Rule-Given column absorbs much of the No-Image penalty (the gap shrinks to 6--13 points), suggesting that the rule context partially substitutes for the missing visual scene by narrowing the set of plausible answers, even though it cannot fully recover the visual evidence.

\begin{table}[t]
\centering
\small
\setlength{\tabcolsep}{5pt}
\caption{Robustness to image perturbation on \bench (5{,}053 items). For each base VLM we report overall accuracy (\%) under three input conditions (\emph{Normal}: original image; \emph{No Image}: image removed; \emph{Image Corruption}: Gaussian blur at ImageNet-C severity~3) crossed with three prompting strategies (\emph{Direct}, \emph{Reasoning}, \emph{Rule-Given}). Within each (model, prompt) column triplet we \textbf{bold} the highest accuracy.}
\label{tab:image-perturbation-full}
\begin{tabular}{llccc}
\toprule
Model & Image Setting & Direct & Reasoning & Rule-Given \\
\midrule
\multirow{3}{*}{LLaVA-1.6-7B} & Normal           & 42.15           & \textbf{43.20}  & 57.11           \\
                              & No Image         & 40.53           & 36.93           & \textbf{57.27}  \\
                              & Image Corruption & \textbf{43.06}  & 39.96           & 56.11           \\
\midrule
\multirow{3}{*}{Gemma3-12B}   & Normal           & \textbf{60.12}  & \textbf{58.54}  & \textbf{72.17}  \\
                              & No Image         & 45.52           & 41.66           & 66.38           \\
                              & Image Corruption & 56.90           & 56.82           & 70.29           \\
\midrule
\multirow{3}{*}{InternVL3-8B} & Normal           & \textbf{61.01}  & \textbf{57.37}  & \textbf{77.18}  \\
                              & No Image         & 48.53           & 44.07           & 65.47           \\
                              & Image Corruption & 58.94           & 53.31           & 74.17           \\
\midrule
\multirow{3}{*}{Qwen3-VL-8B}  & Normal           & \textbf{58.82}  & 50.50           & \textbf{77.66}  \\
                              & No Image         & 47.16           & 37.01           & 64.60           \\
                              & Image Corruption & 56.98           & \textbf{50.85}  & 75.14           \\
\bottomrule
\end{tabular}
\end{table}

\subsection{Rule-Context Ablation per Task Category}
\label{sec:appendix-rule-ablation}
Fig.~4 in the main paper summarizes the rule-context ablation visually across the four task categories. Table~\ref{tab:ablation-rules-per-cat} reports the underlying numbers, broken down by task category for both the strongest base VLMs and our two \method checkpoints. We highlight two findings.

The Wrong-Rule column quantifies how much accuracy a model loses when forced to apply a non-applicable rule snippet drawn from the same country's handbook. For the base VLMs, the gap is largest on the Region task---where the answer hinges almost entirely on the cited rule rather than on visual evidence---with Wrong-Rule accuracy collapsing to 40--46\% relative to Rule-Given accuracy of 58--63\%. \method$^{\dagger}$ and \method$^{\ddagger}$ exhibit substantially smaller Wrong-Rule gaps across every category; for instance, \method$^{\ddagger}$ on Region drops only from 67\% to 63\%. This pattern is consistent with our interpretation that on-policy rule-conditioned distillation internalizes regional traffic knowledge into the model's parameters, allowing it to discount a misleading external rule when the visual scene contradicts it.

The Full-Handbook column reflects a more realistic deployment condition in which the precise rule clause is not pre-selected. Here, the \method checkpoints maintain accuracy comparable to their Rule-Given numbers, whereas the base VLMs degrade noticeably, especially on Prediction and Planning, since they must first locate the relevant clause within a long multi-section document before applying it. Together, these two columns indicate that test-time rule prompting is brittle along two distinct axes---noise in the rule itself and the burden of retrieval---both of which \method largely sidesteps by absorbing the rule knowledge during training.

\begin{table}[t]
\centering
\small
\setlength{\tabcolsep}{4pt}
\caption{Rule-context ablation per task category on \bench (5{,}053 items). Each cell reports overall accuracy (\%) for the given (task category, prompt setting, model) triplet. \textbf{Rule-Given} provides the cited country-specific rule; \textbf{Wrong-Rule} replaces it with a deterministically sampled non-applicable rule from the same country; \textbf{Full-Handbook} substitutes the cited snippet with the entire country handbook. \method$^{\dagger}$ and \method$^{\ddagger}$ are our distilled checkpoints over Qwen2.5-VL-7B and InternVL3-8B respectively. Companion to Fig.~4 in the main paper.}
\label{tab:ablation-rules-per-cat}
\begin{tabular}{llccccc}
\toprule
Category & Setting & Qwen3-VL-8B & InternVL3-8B & Gemma3-12B & \method$^{\dagger}$ & \method$^{\ddagger}$ \\
\midrule
\multirow{3}{*}{Perception} & Rule-Given    & 84.99 & \textbf{88.27} & 72.15 & 81.26 & 85.06 \\
                            & Wrong-Rule    & 60.94 & 62.65          & 51.70 & 82.11 & \textbf{83.16} \\
                            & Full-Handbook & 85.65 & \textbf{88.27} & 74.38 & 79.75 & 82.70 \\
\midrule
\multirow{3}{*}{Prediction} & Rule-Given    & \textbf{78.58} & 75.83 & 72.89 & 69.12 & 75.46 \\
                            & Wrong-Rule    & 59.47          & 59.83 & 55.51 & 58.64 & \textbf{72.15} \\
                            & Full-Handbook & 49.54          & 73.25 & 72.15 & 64.89 & \textbf{73.53} \\
\midrule
\multirow{3}{*}{Planning}   & Rule-Given    & 83.29 & 81.17 & \textbf{87.11} & 76.51 & 85.41 \\
                            & Wrong-Rule    & 69.13 & 65.22 & 78.03          & 71.50 & \textbf{81.17} \\
                            & Full-Handbook & 51.48 & 82.70 & \textbf{86.09} & 75.83 & 79.73 \\
\midrule
\multirow{3}{*}{Region}     & Rule-Given    & 62.70 & 61.19 & 57.62 & 57.38 & \textbf{67.06} \\
                            & Wrong-Rule    & 42.06 & 39.84 & 46.19 & 49.60 & \textbf{62.86} \\
                            & Full-Handbook & 51.83 & 62.30 & 60.95 & 56.75 & \textbf{63.73} \\
\bottomrule
\end{tabular}
\end{table}

\subsection{Error Analysis}
\label{sec:appendix-error-qwen}
\begin{figure}[!t]
    \centering
    \includegraphics[width=1.0\linewidth, trim=0cm 0.5cm 0cm 0cm, clip]{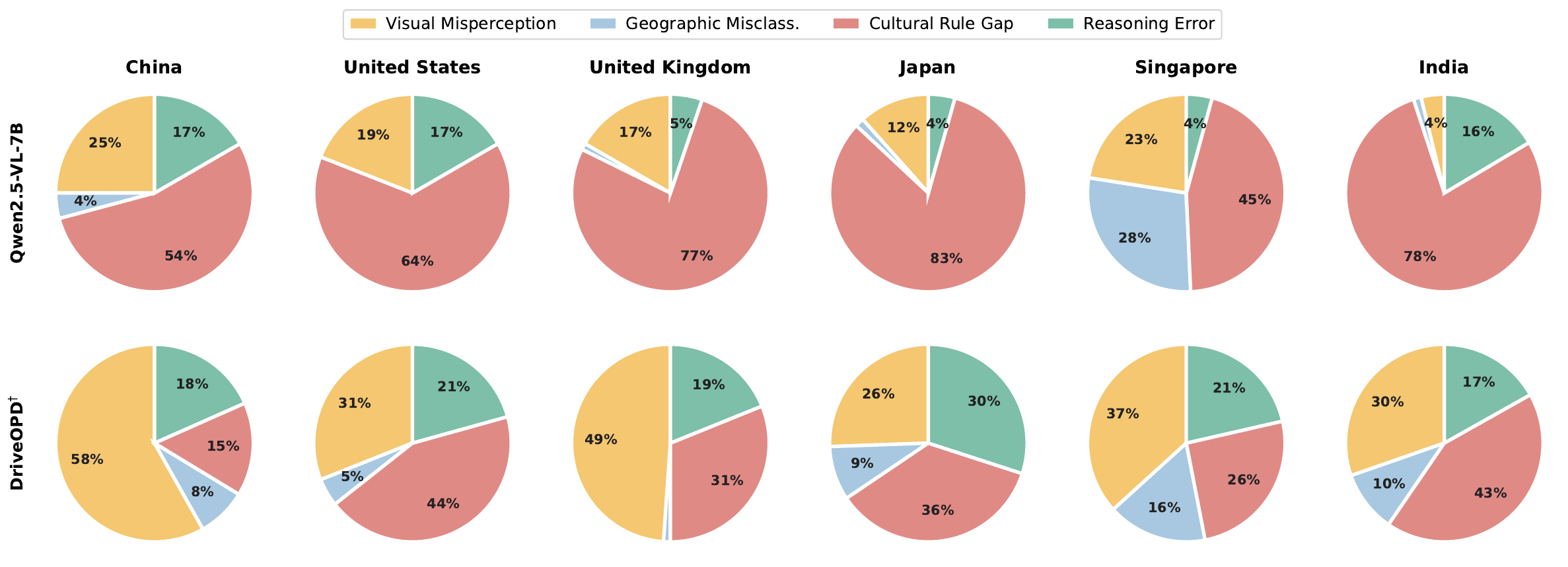}
    \vspace{-1em}
    \caption{Country-wise distribution of error types for Qwen2.5-VL-7B and \method$^{\dagger}$ under the Reasoning setting. Each pie chart shows the proportion of four error categories within a country: Visual Misperception, Geographic Misclassification, Cultural Rule Gap, and Reasoning Error. \looseness=-1}
    \label{fig:error_analysis_qwen}
    \vspace{-1em}
\end{figure}

Fig.~5 in the main paper visualizes the per-country error decomposition for InternVL3-8B and \method$^{\ddagger}$. To verify that the observed pattern is not specific to a single backbone, Fig.~\ref{fig:error_analysis_qwen} reports the same decomposition for the Qwen2.5-VL-7B / \method$^{\dagger}$ pair.

The two backbone families exhibit a consistent qualitative pattern. For both base models, \emph{Cultural Rule Gap} dominates the error mass, accounting for 35--84\% of sampled errors per country, with the highest concentrations on Japan---where left-hand traffic and Japan-specific signage diverge most strongly from common pretraining priors---and India, where dense mixed-traffic conventions are underrepresented in standard driving corpora. After distillation, both \method variants exhibit a substantial reduction in Cultural Rule Gap across every country, and the residual error mass shifts toward \emph{Visual Misperception} and \emph{Reasoning Error}. This shift is itself informative: once regional rule knowledge is internalized, fine-grained visual understanding becomes the more prominent bottleneck, suggesting a natural target for follow-up work---improving perceptual fidelity on region-specific signage and lane markings rather than further refining rule grounding.

% TODO: 2-3 sentence takeaway parallel to InternVL3 case in main paper:
%   - Cultural Rule Gap dominates Qwen2.5-VL errors in JP / IND
%   - DRIVEOPD$^{\dagger}$ shifts errors toward Visual Misperception /
%     Reasoning Error (rule gap closed, visual bottleneck remains).

% =============================================================================
\section{Benchmark Construction Details}
\label{sec:appendix-construction}
% =============================================================================

\subsection{Culture-Specific Traffic Scenarios}
\label{sec:appendix-13cats}

To turn ``cultural relevance'' into an operational filter rather than an intuitive judgment, we manually define 13 categories of culture-specific traffic situations, drawing on crowdsourced traffic regulations from Wikipedia and prior studies on cross-country driving behavior. A category is retained only if national traffic codes diverge along at least one of three axes: \textbf{(L) Legality} of the maneuver itself (e.g., \emph{turn-on-red}); \textbf{(E) Enforcement} strictness or prevailing compliance norms (e.g., \emph{pedestrian crossings}, \emph{bus lanes}); and \textbf{(I) Infrastructure} configuration of the regulatory setup (e.g., \emph{box junctions}, \emph{HOV lanes}, \emph{contraflow lanes}). Categories governed solely by universal driving common sense are excluded. The resulting set spans the full taxonomy of culturally divergent maneuvers we observed during pilot exploration of the six source datasets.

This taxonomy serves a dual purpose in our pipeline. First, the visual entities mentioned in each scenario description seed the keyword set used by the open-vocabulary detector in Stage~1 of the scenario-mining cascade (Fig.~2 in the main paper), enabling coarse retrieval of frames that are at least nominally relevant to a culturally divergent situation. Second, the per-country entries within each scenario---each recording a scene description, the governing rule, and the expected ego action---constitute the traffic rule handbook that supervises both Stage~2 semantic refinement and the QA-verification step in Section~3 of the main paper. By grounding both stages in a common handbook, we ensure that scene mining and QA validation reason about the \emph{same} notion of cultural relevance.

The remainder of this subsection enumerates the full per-country entry for each of the 13 scenarios. For every scenario, we list the visual-entity keyword used during coarse retrieval, followed by one block per country specifying the scenario instantiation, the relevant rule tag, the expected ego action, and a brief description of local compliance norms and infrastructure context. These entries form the shared supervision signal used by both the scenario miner and the QA verifier.

\paragraph{S1 -- Turn-on-Red / Signal Exception (L).}~\\
\begin{lstlisting}[basicstyle=\ttfamily\scriptsize, breaklines=true, frame=single, backgroundcolor=\color{backcolour}]
Scenario S1: Turn-on-Red / Signal Exception
Keyword: a car. a traffic light. a pedestrian. a crosswalk.

[UK] No Turn on Red (default)
  specific_scenario      : Signalized intersection, vehicle wants to turn left (left-side driving).
  expected_motion_action : Must stop and wait for green/arrow.
  detailed_description   : Default prohibition. Only permitted with explicit signal/arrow.

[US] Right Turn on Red (conditional)
  specific_scenario      : Signalized intersection, vehicle wants to turn right at red.
  expected_motion_action : Full stop → yield to pedestrians & cross traffic → turn if clear.
  detailed_description   : Legal in many states unless sign prohibits. Complete stop required.

[CN] Right Turn on Red (often allowed, city-dependent)
  specific_scenario      : Signalized intersection, vehicle wants to turn right at red.
  expected_motion_action : Slow/stop → yield to pedestrians & bikes → turn if safe.
  detailed_description   : Common practice allows right-turn at red, but strong pedestrian/bike conflict risk.

[JP] No Turn on Red (default)
  specific_scenario      : Signalized intersection, vehicle wants to turn left (left-side driving).
  expected_motion_action : Must stop and wait for green/arrow.
  detailed_description   : Similar to UK; default prohibition.

[SG] No Turn on Red (strict)
  specific_scenario      : Signalized intersection, vehicle wants to turn left (left-side driving).
  expected_motion_action : Must stop and wait for green arrow.
  detailed_description   : Highly regulated; strong enforcement, typically prohibited unless explicit.

[IND] No Turn on Red (default)
  specific_scenario      : Signalized intersection, vehicle wants to turn left (left-side driving).
  expected_motion_action : Must stop unless explicit free-left signage or green arrow.
  detailed_description   : Default prohibition under Motor Vehicles Act; some junctions have explicit free-left signage that lets drivers proceed if yielding to pedestrians.

\end{lstlisting}

\paragraph{S2 -- Unsignalized Pedestrian Crossing (E).}~\\
\begin{lstlisting}[basicstyle=\ttfamily\scriptsize, breaklines=true, frame=single, backgroundcolor=\color{backcolour}]
Scenario S2: Unsignalized Pedestrian Crossing
Keyword: a car. a pedestrian. a crosswalk. a bicycle. a traffic light.

[UK] Strong pedestrian priority
  specific_scenario      : Vehicle approaches zebra crossing without traffic light; pedestrian waiting.
  expected_motion_action : Slow early → stop before zebra → allow crossing.
  detailed_description   : Drivers expected to anticipate crossing even before step-in.

[US] Pedestrian priority (varies by state enforcement)
  specific_scenario      : Same scenario.
  expected_motion_action : Yield if pedestrian intends to cross; stop if entering crosswalk.
  detailed_description   : Legal obligation exists but behavioral variation across cities.

[CN] Pedestrian priority (mixed compliance)
  specific_scenario      : Same scenario with pedestrian + e-bike approaching.
  expected_motion_action : Yield; anticipate late crossing behavior.
  detailed_description   : Mixed vehicle–nonmotorized interaction increases uncertainty.

[JP] Strong pedestrian priority
  specific_scenario      : Same scenario on narrow urban street.
  expected_motion_action : Slow significantly → stop completely.
  detailed_description   : High social compliance; cautious driving norm.

[SG] Strict pedestrian priority
  specific_scenario      : Same scenario near school zone.
  expected_motion_action : Stop fully; high compliance expectation.
  detailed_description   : Strict enforcement, especially near schools.

[IND] Pedestrian priority (weak compliance)
  specific_scenario      : Vehicle approaches zebra crossing; pedestrian (and often auto-rickshaw / 2-wheeler) approaching.
  expected_motion_action : Yield by law; in practice slow and assert with mixed traffic.
  detailed_description   : Motor Vehicles Act mandates pedestrian priority and prohibits overtaking near crossings, but enforcement is inconsistent.

\end{lstlisting}

\paragraph{S3 -- Bus Lane / Time-Dependent Lane (I, E).}~\\
\begin{lstlisting}[basicstyle=\ttfamily\scriptsize, breaklines=true, frame=single, backgroundcolor=\color{backcolour}]
Scenario S3: Bus Lane / Time-Dependent Lane
Keyword: a car. a bus. a bus lane marking. a road sign. a traffic light. a pedestrian.

[UK] Time-dependent bus lane
  specific_scenario      : Urban road with bus lane active 7–10 AM.
  expected_motion_action : Avoid bus lane during active hours.
  detailed_description   : Lane becomes restricted during specified hours; sign-based.

[US] Bus-only lane (city-specific)
  specific_scenario      : Urban road with bus-only lane.
  expected_motion_action : Avoid unless permitted; merge carefully.
  detailed_description   : Rules vary by city; signage critical.

[CN] Bus-only lane (often peak-hour)
  specific_scenario      : Urban road with bus lane + electric bikes nearby.
  expected_motion_action : Avoid; manage bike conflicts.
  detailed_description   : Complex interaction with bikes and taxis.

[JP] Bus priority lane
  specific_scenario      : Urban arterial with designated bus lane.
  expected_motion_action : Avoid bus lane; merge safely.
  detailed_description   : Clear marking; disciplined traffic.

[SG] Strict time-dependent bus lane
  specific_scenario      : Bus lane active during peak hours.
  expected_motion_action : Strictly avoid during active hours; heavy enforcement.
  detailed_description   : Strong enforcement; automated fines common.

[IND] Bus priority lane (city-specific)
  specific_scenario      : Urban arterial with painted bus priority lane (e.g., Bengaluru Outer Ring Road).
  expected_motion_action : Stay clear of bus lane during operating hours; cars use designated cut-in points.
  detailed_description   : Patchy infrastructure: Delhi BRT dismantled in 2016, but Bengaluru / Ahmedabad / Pune retain Bus Priority Lanes with steep fines.

\end{lstlisting}

\paragraph{S4 -- Roundabout Entry Priority (I, E).}~\\
\begin{lstlisting}[basicstyle=\ttfamily\scriptsize, breaklines=true, frame=single, backgroundcolor=\color{backcolour}]
Scenario S4: Roundabout Entry Priority
Keyword: a car. a roundabout. a yield sign. a lane marking. a traffic light. a pedestrian.

[UK] Roundabout priority (circulating has priority)
  specific_scenario      : Multi-lane roundabout, vehicle entering, circulating traffic present.
  expected_motion_action : Yield at give-way line; enter only with safe gap; keep lane discipline.
  detailed_description   : Roundabouts common; strong norm: give way to right (circulating).

[US] Roundabout priority (often, but less consistent)
  specific_scenario      : Small modern roundabout, mixed driver behavior.
  expected_motion_action : Yield; be conservative to account for inconsistent behaviors.
  detailed_description   : Less frequent; drivers vary → planner must be robust.

[CN] Roundabout + mixed traffic
  specific_scenario      : Roundabout with scooters merging at entry.
  expected_motion_action : Yield; anticipate scooters cutting-in; slow down early.
  detailed_description   : Mixed flow increases uncertainty.

[JP] Roundabout priority
  specific_scenario      : Single-lane roundabout near residential area.
  expected_motion_action : Yield; cautious entry.
  detailed_description   : Less common but rules clear.

[SG] Strict roundabout yield
  specific_scenario      : Urban roundabout with clear yield markings.
  expected_motion_action : Yield strictly; follow lane arrows.
  detailed_description   : Strong signage and compliance.

[IND] Circulating priority (mixed compliance)
  specific_scenario      : Urban roundabout (chowk) with mixed traffic including 2/3-wheelers.
  expected_motion_action : Yield to circulating traffic; many chowks now signalized.
  detailed_description   : Officially priority-to-circulating per IRC:65, but yield compliance is weak so many large roundabouts have been signalized.

\end{lstlisting}

\paragraph{S5 -- Permissive Turn vs.\ Protected Arrow (L).}~\\
\begin{lstlisting}[basicstyle=\ttfamily\scriptsize, breaklines=true, frame=single, backgroundcolor=\color{backcolour}]
Scenario S5: Permissive Turn vs Protected Arrow
Keyword: a car. a traffic light. a left turn arrow. an oncoming car. a pedestrian.

[UK] Unprotected crossing turn
  specific_scenario      : Right turn (left-side driving) with green circular, oncoming straight traffic present.
  expected_motion_action : Proceed only if clear; yield to oncoming.
  detailed_description   : Mirror of US permissive left due to left-side driving.

[US] Permissive left (yield to oncoming)
  specific_scenario      : Left turn with green circular, oncoming straight traffic present.
  expected_motion_action : Yield to oncoming + pedestrians; take gap only when safe.
  detailed_description   : Classic failure case for imitation-only planners.

[CN] Permissive left + vulnerable users
  specific_scenario      : Left turn with green circular + bikes crossing.
  expected_motion_action : Yield to oncoming + bikes; often wait longer.
  detailed_description   : Bike interaction makes rule compliance nontrivial.

[JP] Unprotected crossing turn
  specific_scenario      : Right turn (left-side) with green circular + oncoming traffic.
  expected_motion_action : Yield; conservative gap acceptance.
  detailed_description   : Conservative culture.

[SG] Strict yield on permissive
  specific_scenario      : Right turn (left-side) with green circular at busy junction.
  expected_motion_action : Yield strictly; prefer waiting for arrow phase if present.
  detailed_description   : High enforcement; risk-averse behavior expected.

[IND] Unprotected crossing turn (mixed)
  specific_scenario      : Signalized intersection, vehicle wants to turn right across oncoming (left-side driving).
  expected_motion_action : Yield to oncoming; vulnerable users common in same conflict zone.
  detailed_description   : Permissive default unless protected arrow phase; arrow phases added at high-volume junctions.

\end{lstlisting}

\paragraph{S6 -- Box Junction / Keep-Clear (I).}~\\
\begin{lstlisting}[basicstyle=\ttfamily\scriptsize, breaklines=true, frame=single, backgroundcolor=\color{backcolour}]
Scenario S6: Box Junction / Keep Clear
Keyword: a car. a yellow box junction. a traffic light. a crosswalk. a pedestrian.

[UK] Box junction rule
  specific_scenario      : Approaching box junction; downstream lane congested.
  expected_motion_action : Do not enter unless exit is clear; wait before the box.
  detailed_description   : Very strong legality rule in UK.

[US] Keep intersection clear
  specific_scenario      : “Keep clear” marking near intersection.
  expected_motion_action : Avoid blocking intersection.
  detailed_description   : Not as standardized; still illegal to block.

[CN] Keep-clear marking
  specific_scenario      : Intersection with yellow grid marking; congestion.
  expected_motion_action : Do not enter if you’ll stop inside.
  detailed_description   : Common in dense cities.

[JP] Keep-clear rule
  specific_scenario      : Intersection with keep-clear marking near rail crossing.
  expected_motion_action : Wait; avoid blocking junction.
  detailed_description   : Often near rail/tram.

[SG] Strict keep-clear
  specific_scenario      : Junction with keep-clear box; camera enforcement.
  expected_motion_action : Wait strictly until exit is clear.
  detailed_description   : Strong enforcement.

[IND] Box junction (city-specific)
  specific_scenario      : Urban intersection with painted yellow box.
  expected_motion_action : Do not enter unless exit is clear.
  detailed_description   : Implemented in Mumbai and Bangalore; violations attract lane-violation challans.

\end{lstlisting}

\paragraph{S7 -- School Zone Time Window (E, I).}~\\
\begin{lstlisting}[basicstyle=\ttfamily\scriptsize, breaklines=true, frame=single, backgroundcolor=\color{backcolour}]
Scenario S7: School Zone Time Window
Keyword: a car. a zone sign. a children. a pedestrian. a traffic light. a school.

[UK] School zone speed rule
  specific_scenario      : Flashing sign indicates school time; kids near curb.
  expected_motion_action : Reduce speed; prepare to stop.
  detailed_description   : Cautious behavior expected.

[US] Time-dependent speed limit
  specific_scenario      : School zone sign with flashing lights.
  expected_motion_action : Obey reduced speed.
  detailed_description   : Clear “when flashing” trigger.

[CN] School zone caution
  specific_scenario      : School gate area with pickup traffic.
  expected_motion_action : Slow substantially.
  detailed_description   : High interaction complexity.

[JP] School zone caution
  specific_scenario      : School street with narrow lane + children.
  expected_motion_action : Slow and be ready to stop.
  detailed_description   : High caution norm.

[SG] Strict school-zone enforcement
  specific_scenario      : School zone with crossing guard.
  expected_motion_action : Slow + strict yielding.
  detailed_description   : Strong enforcement.

[IND] School zone speed limit (25 km/h)
  specific_scenario      : Approach school zone with children / pedestrian signage.
  expected_motion_action : Slow to no more than 25 km/h; school buses capped at 40 km/h.
  detailed_description   : MoRTH-mandated 25 km/h limit applies to all school zones (urban or rural); enforcement uneven, supplemented by awareness campaigns.

\end{lstlisting}

\paragraph{S8 -- Work Zone / Temporary Lane Shift (E, I).}~\\
\begin{lstlisting}[basicstyle=\ttfamily\scriptsize, breaklines=true, frame=single, backgroundcolor=\color{backcolour}]
Scenario S8: Work Zone / Temporary Lane Shift
Keyword: a car. a traffic cone. a construction sign. a lane arrow. a traffic light. a pedestrian.

[UK] Temporary traffic control
  specific_scenario      : Cones shift lane left; temporary speed limit.
  expected_motion_action : Follow cones/arrows; reduce speed.
  detailed_description   : Temporary rules override default markings.

[US] Work-zone compliance
  specific_scenario      : Work zone with flagger; lane reduced.
  expected_motion_action : Obey flagger; reduce speed; merge early.
  detailed_description   : Flaggers add human instruction.

[CN] Conflicting markings
  specific_scenario      : Cones + old lane markings visible.
  expected_motion_action : Prioritize temporary markings.
  detailed_description   : Conflicting cues challenge perception.

[JP] Work-zone detour
  specific_scenario      : Narrow work zone with detour.
  expected_motion_action : Follow detour.
  detailed_description   : Polite yielding culture.

[SG] Strict work-zone rules
  specific_scenario      : Structured work zone with clear arrows.
  expected_motion_action : Follow temporary guidance.
  detailed_description   : Strong compliance.

[IND] Work-zone control (IRC:SP:55)
  specific_scenario      : Approach urban or highway construction zone with traffic cones and signage.
  expected_motion_action : Slow down; follow taper signage and channelization through the zone.
  detailed_description   : Indian Roads Congress IRC:SP:55-2014 governs Work-zone Traffic Management Plans. Conflicting / faded markings common in practice.

\end{lstlisting}

\paragraph{S9 -- Emergency Vehicle Approaching (E).}~\\
\begin{lstlisting}[basicstyle=\ttfamily\scriptsize, breaklines=true, frame=single, backgroundcolor=\color{backcolour}]
Scenario S9: Emergency Vehicle Approaching
Keyword: a car. an ambulance. a police car. flashing lights. a traffic light. a pedestrian.

[UK] Emergency right-of-way
  specific_scenario      : Ambulance behind with siren.
  expected_motion_action : Pull left if safe.
  detailed_description   : Orderly yielding expected.

[US] Emergency pull-over
  specific_scenario      : Police car behind; intersection ahead.
  expected_motion_action : Pull over to right.
  detailed_description   : Do not block intersection.

[CN] Emergency + mixed traffic
  specific_scenario      : Ambulance approaching; scooters nearby.
  expected_motion_action : Create corridor.
  detailed_description   : Mixed traffic increases hazard.

[JP] Emergency narrow road
  specific_scenario      : Fire truck behind on narrow street.
  expected_motion_action : Stop/edge carefully.
  detailed_description   : Narrow streets require nuance.

[SG] Emergency compliance
  specific_scenario      : Ambulance behind; bus lane present.
  expected_motion_action : Make space quickly.
  detailed_description   : Disciplined behavior.

[IND] Emergency right-of-way (MV Act 2019)
  specific_scenario      : Ambulance with siren and lights approaches in dense urban traffic.
  expected_motion_action : Move left toward kerb; maintain at least 50 m distance.
  detailed_description   : Motor Vehicles (Amendment) Act 2019: Rs 10,000 fine and up to 6-month imprisonment for obstruction. Compliance complicated by dense mixed traffic.

\end{lstlisting}

\paragraph{S10 -- Bus Stop Re-entry Priority (E).}~\\
\begin{lstlisting}[basicstyle=\ttfamily\scriptsize, breaklines=true, frame=single, backgroundcolor=\color{backcolour}]
Scenario S10: Bus Stop Re-entry Priority
Keyword: a car. a bus. a bus stop. a turn signal. a traffic light. a pedestrian.

[UK] Bus priority etiquette
  specific_scenario      : Bus signaling to pull out from stop.
  expected_motion_action : Yield to bus.
  detailed_description   : Drivers expected to accommodate buses.

[US] Bus merge courtesy
  specific_scenario      : Bus leaving stop; no special sign.
  expected_motion_action : Cautious yield.
  detailed_description   : Less consistent behavior.

[CN] Bus + bike conflict
  specific_scenario      : Bus pulls out; bikes filtering.
  expected_motion_action : Slow and yield.
  detailed_description   : Bike conflict common.

[JP] Bus merge etiquette
  specific_scenario      : Bus re-enters traffic on arterial.
  expected_motion_action : Yield smoothly.
  detailed_description   : Predictable yielding.

[SG] Strong bus priority
  specific_scenario      : Bus leaving stop with priority markings.
  expected_motion_action : Yield.
  detailed_description   : High compliance.

[IND] Bus merge (mixed compliance)
  specific_scenario      : Bus signalling re-entry from bus stop; auto-rickshaws and 2-wheelers nearby.
  expected_motion_action : Yield courteously; watch for 2-wheelers / cycles passing on either side.
  detailed_description   : No formal priority etiquette; dense mixed traffic and parked vehicles near stops complicate the merge.

\end{lstlisting}

\paragraph{S11 -- One-Way Street with Contraflow Bike or Bus Lane (L, I).}~\\
\begin{lstlisting}[basicstyle=\ttfamily\scriptsize, breaklines=true, frame=single, backgroundcolor=\color{backcolour}]
Scenario S11: One-way + Contraflow Bike/Bus Lane
Keyword: a car. a bicycle. a one-way sign. a bike lane. a traffic light. a pedestrian.

[UK] Contraflow exception
  specific_scenario      : One-way street with contraflow bike lane.
  expected_motion_action : Follow one-way; avoid bike lane.
  detailed_description   : Signage important.

[US] One-way compliance
  specific_scenario      : Downtown one-way street.
  expected_motion_action : Follow direction.
  detailed_description   : Simple but common.

[CN] One-way ambiguity
  specific_scenario      : One-way with ambiguous signage.
  expected_motion_action : Slow and confirm direction.
  detailed_description   : Mixed traffic complexity.

[JP] One-way narrow street
  specific_scenario      : Narrow residential one-way.
  expected_motion_action : Follow direction slowly.
  detailed_description   : Narrow geometry interactions.

[SG] Strict signage
  specific_scenario      : One-way with explicit lane segregation.
  expected_motion_action : Follow signage strictly.
  detailed_description   : Clear segregation.

[IND] One-way (limited contraflow infrastructure)
  specific_scenario      : One-way urban street with a cyclist proceeding.
  expected_motion_action : Cyclists generally follow vehicular direction; formal contraflow bike lanes rare.
  detailed_description   : Dedicated cycle lanes exist in Delhi and Chandigarh, but formal contraflow bike-lane infrastructure on one-way streets is rare nationwide.

\end{lstlisting}

\paragraph{S12 -- U-turn at Median Opening (L, I).}~\\
\begin{lstlisting}[basicstyle=\ttfamily\scriptsize, breaklines=true, frame=single, backgroundcolor=\color{backcolour}]
Scenario S12: U-turn / Median Opening
Keyword: a car. a median. a U-turn sign. an oncoming car. a traffic light. a pedestrian.

[UK] U-turn restrictions
  specific_scenario      : Signalized intersection with U-turn signage.
  expected_motion_action : Only U-turn if allowed.
  detailed_description   : Conservative approach.

[US] U-turn permitted by sign
  specific_scenario      : Median opening with U-turn permitted.
  expected_motion_action : U-turn if safe.
  detailed_description   : Sign grounding critical.

[CN] Mixed compliance
  specific_scenario      : Wide intersection with scooters crossing.
  expected_motion_action : Prefer not unless clearly allowed.
  detailed_description   : High conflict scenario.

[JP] Conservative U-turn
  specific_scenario      : Narrow U-turn slot.
  expected_motion_action : Avoid unless designed.
  detailed_description   : Cautious driving culture.

[SG] Strict U-turn pocket
  specific_scenario      : Dedicated U-turn pocket.
  expected_motion_action : Use designated pocket.
  detailed_description   : Strong structure.

[IND] Median U-turn pockets (every ~2 km)
  specific_scenario      : Divided 4-lane highway with median opening; vehicle wants to U-turn.
  expected_motion_action : Use designated median openings; yield to oncoming traffic before crossing.
  detailed_description   : Median openings every approx. 2 km on 4-lane divided carriageways; treated as hazardous locations in safety studies.

\end{lstlisting}

\paragraph{S13 -- HOV / Occupancy-Dependent Lane (I).}~\\
\begin{lstlisting}[basicstyle=\ttfamily\scriptsize, breaklines=true, frame=single, backgroundcolor=\color{backcolour}]
Scenario S13: HOV / Occupancy-dependent Lane
Keyword: a car. a HOV lane marking. a road sign. a highway lane. a traffic light. a pedestrian.

[UK] Time-dependent lane access
  specific_scenario      : Express lane with time restriction.
  expected_motion_action : Enter only if permitted.
  detailed_description   : Sign-based control.

[US] HOV lane occupancy rule
  specific_scenario      : HOV lane requires 2+ occupants.
  expected_motion_action : Use lane only if occupancy satisfied.
  detailed_description   : Explicit dynamic rule.

[CN] Vehicle-class restriction
  specific_scenario      : Lane restricted by vehicle class/time.
  expected_motion_action : Follow restriction.
  detailed_description   : City policy variation.

[JP] Vehicle restriction lane
  specific_scenario      : Lane restricted by vehicle type/time.
  expected_motion_action : Follow restriction conservatively.
  detailed_description   : General restriction modeling.

[SG] Strict lane restriction
  specific_scenario      : Lane restricted by time/vehicle type.
  expected_motion_action : Follow restriction strictly.
  detailed_description   : Strong enforcement.

[IND] Heavy-vehicle lane restriction (no HOV)
  specific_scenario      : Multi-lane highway with vehicle-class lane restrictions for trucks / heavy vehicles.
  expected_motion_action : Heavy vehicles confined to designated lane(s); HGV speed limited.
  detailed_description   : India has no HOV / carpool lanes. Lane-discipline rules apply to heavy goods vehicles; e.g., Delhi enforces strict bus / goods-carrier lane rules.

\end{lstlisting}

\subsection{Traffic Rule Handbook}
\label{sec:appendix-handbook}
% TODO: 1 short paragraph describing the 20-section schema (S1-S20),
%       noting that S1-S15 cover topics that vary by country (so the
%       SAME section ID has DIFFERENT content per country) and S16-S20
%       are augmented sections (driver behavior, plates, signs, markings,
%       tolls) added later for coverage parity across countries.

For every country we compile a structured traffic-rule handbook of 20 numbered sections (S1--S20), drawing on crowdsourced traffic regulations from Wikipedia, the regional traffic descriptions released by LLaDA-AV~\cite{li2024driving}, and prior studies on cross-country driving behavior~\cite{dong2024towards,wang2025impact}. Sections S1--S15 cover topical anchors that the 13 culture-specific scenarios of Section~\ref{sec:appendix-13cats} can directly reference, including driving side and lane discipline, speed limits, traffic-light conventions, turn-on-red rules, pedestrian priority, roundabouts and merging, and box junctions and keep-clear markings. Sections S16--S20 cover five additional topics that provide parity coverage across countries on aspects frequently surfaced during scene mining: driver-behavior norms, license-plate schemes, road-sign conventions, lane and pavement markings, and tolls and electronic charging. Within each country, sections are ordered to follow the natural progression of that country's traffic code, so the same section ID can encode different topical content across countries; for instance, box-junction rules appear at S11 in the UK handbook but at a different position in the Chinese handbook. This design preserves the local logical structure of each country's regulations rather than forcing a single global ordering that would be unnatural for any individual country. The full per-country handbooks listed below serve as the supervision signal for both Rule-Given conditioning at evaluation time and teacher conditioning during \method training. Country names embedded in the text are anonymized to ``this country'' during training and inference, as described in Section~3.4 of the main paper, but are printed verbatim here for readability.

\paragraph{United Kingdom (UK).}~\\
\begin{lstlisting}[basicstyle=\ttfamily\scriptsize, breaklines=true, frame=single, backgroundcolor=\color{backcolour}]
Traffic-Rule Handbook -- United Kingdom (UK)
============================================

S1 Driving side & lane discipline: In the UK you drive on the left. As a default, keep left unless road markings/signs say otherwise; move out to overtake, to turn right, or to pass an obstruction, then return to the left when safe.

S2 Speed limits are in mph (not km/h). Unless signed otherwise, typical “national speed limit” defaults are 70 mph on motorways/dual carriageways and 60 mph on single carriageways, while built-up areas are commonly 30 mph (and widely 20 mph zones exist in many places). Always follow posted limit signs and the national speed limit symbol where used.

S3 Traffic lights (vehicles): Standard UK sequence is red → red+amber (get ready, still stop) → green → amber (stop unless unsafe). There is no general “turn on red” permission—treat a red signal as stop unless a separate permitted movement is explicitly shown.

S4 Green filter arrows at signals: A green arrow indicates a permitted movement for a specific lane/direction (a “filter”). Only enter that lane if you intend to go that way, and proceed only in the arrow’s direction when it is lit (even if other movements are held).

S5 Roundabouts: Modern UK-style roundabouts use the “priority rule”: traffic entering must give way to traffic already circulating. In left-hand traffic this generally means giving way to vehicles approaching from your right; choose the correct lane early and signal on exit.

S6 “Give Way” / “Stop” control: At many junctions you’ll see a GIVE WAY sign and/or an inverted triangle road marking with broken white lines—yield to traffic on the main road. A STOP sign and solid stop line require a full stop before proceeding.

S7 Zebra/parallel crossings (uncontrolled): Zebra crossings are marked with black-and-white stripes and Belisha beacons. Drivers should be prepared to slow/stop for pedestrians waiting, and must give way once a pedestrian has stepped onto the crossing. Do not overtake the vehicle nearest a crossing (especially if it has stopped).

S8 Signal-controlled pedestrian crossings: Pelican, puffin, toucan and related crossings use red/green “man” signals for pedestrians with push-buttons. Pelican crossings are distinctive because after the vehicle red phase, a flashing amber may allow vehicles to proceed if the crossing is clear; puffin/toucan crossings do not use a flashing amber phase and instead follow a standard traffic-light style sequence.

S9 Box junctions (“yellow boxes”): A criss-cross yellow box means “don’t enter unless your exit is clear.” Exception: you may enter and wait if you are turning right and are only blocked by oncoming traffic or other right-turning vehicles; at signalled roundabouts, do not enter the box unless you can fully clear it without stopping.

S10 Yellow-line parking/waiting rules: Double yellow lines mean no waiting at any time (subject to signed exceptions). Single yellow lines restrict waiting during times shown on nearby plates or zone-entry signs. Short yellow kerb marks (“blips”) add loading/unloading restrictions—check signs.

S11 Red routes (where used): Red lines along the kerb indicate stronger “no stopping” controls than yellow lines. Double red lines mean no stopping at any time; single red lines apply during signed times. Only stop in specifically marked bays/boxes when permitted by accompanying signs.

S12 Motorways basics: Motorways are controlled-access roads—no pedestrians/cyclists and no at-grade junctions; you join via a slip road and must give priority to motorway traffic. Use the left lane as your default cruising lane and overtake to the right; do not weave.

S13 Hard shoulder / emergency areas / overhead control: Do not use a hard shoulder except in an emergency or when directed by signs/authorities. On some motorways the hard shoulder may be opened as a running lane only when overhead signs show it is open; a red ‘X’ indicates a closed lane that must not be used. Use emergency areas only for emergencies and follow SOS signage.

S14 Bus lanes & bus gates: Many UK cities use bus lanes (and “bus gates”) reserved for buses (often also cycles and sometimes taxis) during signed hours; entering when not permitted is an offence. Look for carriageway text (e.g., “BUS LANE” / “BUS GATE”) and regulatory signs.

S15 Vulnerable road users & junction priority: Expect frequent interaction with pedestrians, cyclists, and (in some areas) horses. Give extra space when overtaking vulnerable road users; at junctions, give way to pedestrians crossing or waiting to cross the road you are turning into/from, and respect cyclist advanced stop lines at signals (stop behind the first line on red/amber).

S16 Driver behavior rules: It is illegal to hold and use a hand-held mobile phone or sat-nav while driving in the UK, even when stopped at traffic lights or in slow traffic; hands-free use is permitted only if it does not distract you. Motorcycle and moped riders, and pillion passengers, must wear an approved safety helmet nationwide. The drink-drive limit for ordinary private-car drivers is 80 mg of alcohol per 100 ml of blood in England, Wales and Northern Ireland, but 50 mg/100 ml in Scotland. Do not sound the horn while stationary on the road, and in built-up areas you must not use it between 11:30 pm and 7:00 am except when another road user is a danger. Use full beam only on dark, unlit roads; dip your headlights for oncoming traffic, when following another vehicle, and whenever full beam could dazzle. Daytime running lights are common on newer vehicles, but they do not replace dipped headlights in poor visibility.

S17 Vehicle registration plates: Standard UK private cars display black characters on a white reflective plate at the front and on a yellow reflective plate at the rear. The registration mark itself uses only Latin letters A–Z and Arabic numerals 0–9 in the prescribed Charles Wright style; other scripts are not used in legal registration marks. In Great Britain, most vehicles first registered since 1 September 2001 use the format AB12 CDE: the first two letters are the local memory tag, the two digits are the age identifier (changing on 1 March and 1 September, for example 24 and 74 in 2024), and the last three letters are random. Northern Ireland uses a separate format, typically ABC 1234, with no age identifier. Car plates are usually a single line, but a two-line plate is permitted where the mounting space is square or narrow. An optional left-hand band may show an approved national flag and identifier; for international use, only a UK identifier with the Union Flag replaces a separate country sticker.

S18 Sign system & visual cues: In the United Kingdom, most regulatory signs are circular: a red ring shows a prohibition or restriction (for example speed limits, no entry, no right turn), while a blue circle gives a mandatory instruction such as turn left or pass on a particular side. Warning signs are normally red-bordered triangles with a white background, not yellow diamonds; direction and information signs are instead usually rectangular, with blue used for motorways and green for primary routes. The STOP sign is the standard red octagon with white capital letters “STOP”, and the same word is usually painted on the road at the stop line. Give-way control is shown by an inverted white triangle with a red border, often reinforced by “GIVE WAY” road markings and a broken give-way line. School or children warnings use a red-bordered triangle showing children, and there is no unique nationwide “school zone” sign style; around schools, signed 20 mph limits or timed 20 mph zones are common, sometimes with flashing amber lamps.

S19 Road & pavement markings: In the UK, centre and lane markings are generally white, not yellow: opposing traffic is usually separated by broken white centre lines, while double white lines mean extra restriction. If the line nearest you is solid, you must not cross or straddle it except for limited Highway Code exceptions; if your side is broken, you may cross when safe. Lane and carriageway edges are also marked in white: lanes in the same direction are divided by short broken white lines, hazard warning lines use longer, closer dashes, and many higher-speed roads have a solid white edge line at the nearside. Do not confuse centre-line rules with kerbside double yellow lines, which mean no waiting at any time; UK “no overtaking” is primarily shown by double white centre lines. Special road text commonly includes BUS LANE, BUS STOP and cycle-lane symbols. Zebra crossings are wide black-and-white stripes, usually with Belisha beacons and zig-zag approach markings where stopping, parking and overtaking are prohibited.

S20 Tolls & special infrastructure: The United Kingdom has no nationwide electronic toll tag system: most roads are free, and charged crossings or local schemes use their own arrangements. A common example is the barrier-free Dart Charge at the Dartford Crossing, which must be paid by midnight the day after travel online, by phone, or via a local account; elsewhere, some toll bridges, tunnels, and congestion-charge areas use separate payment systems. Near schools, watch for 20 mph limits or 20 mph zones, amber flashing “wig-wag” lights at school times, and yellow zig-zag “SCHOOL KEEP CLEAR” markings; stopping on the zig-zags is prohibited during the times shown or while the lights operate. Bus lanes are marked with blue roadside signs and carriageway legends such as “BUS LANE,” with the sign stating permitted vehicle types and operating hours; outside those hours, general traffic may use the lane unless signing says otherwise. In built-up areas, give priority to buses signalling to pull away from stops.
\end{lstlisting}

\paragraph{United States (US).}~\\
\begin{lstlisting}[basicstyle=\ttfamily\scriptsize, breaklines=true, frame=single, backgroundcolor=\color{backcolour}]
Traffic-Rule Handbook -- United States (US)
===========================================

S1 Driving side & lane discipline: In the United States you drive on the right. As a default, keep right on two-way roads unless markings/signs say otherwise; move left to overtake, to position for a left turn, or to avoid an obstruction, then return to the right when safe.

S2 Speed limits are in mph (not km/h). There is no single nationwide default speed limit for all roads; limits vary by state, road type, and local conditions. Always follow posted limit signs, and expect lower signed speeds in school zones, work zones, residential streets, and other special areas.

S3 Traffic lights (vehicles): Standard signals are green, yellow, and red. Green permits movement subject to yielding where required; yellow means stop if you can do so safely; red means stop at the stop line, before the crosswalk, or before entering the intersection. A flashing red signal is treated like a STOP sign, while a flashing yellow signal means proceed with caution.

S4 Turn on red: A major U.S. feature is that right turn on red is generally permitted after a full stop unless a sign prohibits it, but this varies by state and locality. Even where allowed, the driver must yield to pedestrians, cyclists, and other traffic with the right-of-way. Do not assume that a red arrow allows the turn; red-arrow rules vary and are often more restrictive.

S5 All-way stop / four-way stop: All-way stop intersections are especially common in the U.S. Every approaching vehicle must come to a full stop. The first vehicle to arrive generally goes first; if two vehicles arrive at the same time, the vehicle on the right typically has priority. When opposite vehicles arrive together, a left-turning vehicle normally yields to an oncoming vehicle going straight.

S6 STOP / YIELD control: At a STOP sign, stop at the stop line; if there is no stop line, stop before the crosswalk or before entering the intersection where you can see safely. At a YIELD sign, slow down and give way to conflicting traffic and pedestrians before proceeding.

S7 Pedestrian crossings & turning priority: Drivers must yield to pedestrians in crosswalks. At many intersections, this includes marked crosswalks and also unmarked crosswalks implied by the intersection geometry under state law. Turning vehicles must not cut across pedestrians who are lawfully crossing.

S8 School buses: School-bus rules are a major U.S.-specific feature. Yellow flashing lights mean the bus is preparing to stop, so slow down and prepare to stop. Red flashing lights with an extended stop arm mean you must stop and remain stopped until the lights stop flashing, the stop arm is withdrawn, and the bus begins moving. On undivided roads, traffic in both directions generally must stop; on divided highways, opposing traffic often does not need to stop, depending on state law and roadway separation.

S9 Freeways / Interstates: U.S. freeways and Interstates are controlled-access roads. Use ramps to enter and exit, accelerate on the entrance ramp, and yield to freeway traffic when merging. Use the right lanes for ordinary travel and the left lanes mainly for overtaking, unless local signs or lane controls specify otherwise.

S10 HOV / carpool / managed lanes: Many U.S. urban highways use HOV or managed lanes reserved for vehicles meeting posted occupancy or toll requirements. These lanes are commonly marked by a white diamond symbol and/or HOV wording. Always check the posted occupancy requirement, hours of operation, and entry/exit restrictions.

S11 Lane-control signals / reversible lanes: Some roads use overhead lane-control signals. A green downward arrow means the lane is open for travel; a red X means the lane is closed and must not be used. On roads with reversible or actively managed lanes, follow overhead signals and signs rather than assuming the lane direction from memory.

S12 Two-way center left-turn lanes: A common U.S. road feature is the shared center left-turn lane on multi-lane roads. Use it only for left turns or, where permitted, short setup movements into driveways or side streets. Do not use it as a through lane, passing lane, or waiting lane for general travel.

S13 Railroad crossings: At railroad crossings, obey flashing lights, bells, gates, signs, and pavement markings. Stop behind the stop line or before the tracks when required, and do not enter unless you can clear the tracks completely. Never drive around or under a lowered gate.

S14 Emergency vehicles & roadside incidents: When an emergency vehicle approaches with siren/lights, pull to the right and stop unless directed otherwise. Many states also have “move over” or slow-down laws for stopped emergency, service, or disabled vehicles on the roadside, so check local law and follow posted requirements.

S15 Sign system note: U.S. road signs are largely standardized through the MUTCD. For visual recognition, common clues include the octagonal STOP sign, the triangular YIELD sign, yellow diamond warning signs, school-zone signs, white regulatory speed-limit signs in mph, and white diamond markings/signs for HOV or carpool lanes.

S16 Driver behavior rules: For ordinary drivers, the United States has no nationwide ban on mobile-phone use while driving; hand-held calling and texting are regulated mainly by state law, while hands-free use is usually allowed where restrictions exist. A specific federal rule applies to commercial motor vehicle drivers, who may not use hand-held mobile phones or text while driving. Motorcycle and moped helmet requirements are not nationwide: each state sets its own rules, and coverage may differ for motorcycles, mopeds, riders, and passengers; some states require helmets for all, others only for younger or less-experienced riders, and a few have no general motorcycle helmet mandate. The per se BAC limit for ordinary private-car drivers is 0.08% in every state except Utah, where it is 0.05%. State laws generally require a working horn and allow its use only when reasonably necessary for safety; unnecessary honking, and some local no-honking or quiet zones, can be penalized. High beams must be dimmed for oncoming traffic and when closely following another vehicle under state distance rules; daytime running lights are permitted but not federally required.

S17 Vehicle registration plates: The United States has no nationwide private-car plate design or numbering system. Plates are issued by each state and the District of Columbia, and standard passenger plates usually have a reflective background with dark letters, but colors, slogans, and graphics differ widely. If a state requires two plates, matching plates with the same number are displayed front and rear; other states require only a rear plate. The usual plate size is the North American 12 × 6 in (305 × 152 mm) single-line format, although motorcycles and some special classes use smaller plates. Passenger serials use only Latin letters and Arabic numerals, with state names and optional slogans in the Latin alphabet; ordinary plates do not use non-Latin scripts. Serial patterns are set by the issuing state, commonly 5 to 7 characters, such as ABC-1234, 123-ABC, or variants with separators. Most ordinary passenger plates do not include a mandatory region code in the serial itself, though some states print county names, use registration stickers, or reserve certain serial blocks for vehicle classes.

S18 Sign system & visual cues: U.S. signs follow the Manual on Uniform Traffic Control Devices (MUTCD). Regulatory signs—the “must do” or prohibition signs—are usually white with black lettering or symbols, most often rectangular, for example SPEED LIMIT, KEEP RIGHT, NO TURN ON RED, and lane-use-control signs; prohibitions may also use red, such as a red circle-and-slash symbol or the red-and-white DO NOT ENTER sign. Warning signs are generally yellow diamonds with black symbols or text (curve, merge, signal ahead, deer crossing), not the red-bordered triangular warning signs common in many countries. The STOP sign is always a red octagon with a white border and the word STOP in English. The YIELD sign is a downward-pointing triangle with a red border, white center, and red YIELD legend. School-area warnings typically use fluorescent yellow-green signs, especially the pentagon-shaped SCHOOL sign with child symbols; school speed limits are shown on white regulatory signs in mph and apply only under the posted conditions, such as stated times, flashing beacons, or when children are present.

S19 Road & pavement markings: In the United States, yellow centerlines separate opposing traffic and white lines separate lanes moving in the same direction. A broken yellow centerline may be crossed to pass when sight distance is clear; if the line on your side is solid, do not pass; a solid-and-broken pair allows passing only from the broken-line side; and double solid yellow means no passing in either direction, except to turn left into or out of a driveway, alley, or intersection where permitted. Lane-edge lines are typically solid white on the right edge of the roadway and solid yellow on the left edge of a divided highway or one-way roadway. Dashed white lines are used where lane lines or edge guidance continue through merges, exits, or lane drops. Double solid white lines separate same-direction lanes where lane changes are prohibited. Pavement words and symbols are usually white: ONLY marks turn-only lanes; BUS ONLY or BUS LANE marks transit lanes; and a white diamond marks HOV/carpool lanes. Crosswalks are marked in white, commonly as two transverse lines or high-visibility ladder/continental bars; no single zebra pattern is required nationwide.

S20 Tolls & special infrastructure: The United States has no single national toll tag. Electronic toll collection is common on toll roads, bridges, and tunnels; E‑ZPass is the main multi‑state system in the East and Midwest, while other regions use systems such as SunPass in Florida and FasTrak in California. Many facilities are now cashless, requiring either a compatible transponder or toll‑by‑plate billing sent to the registered owner. School zones are typically marked with fluorescent yellow‑green SCHOOL or school‑crossing signs and often a reduced speed limit; that lower limit applies only as stated on the sign, such as during posted hours, on school days, or when amber beacons flash. Pavement markings often include “SCHOOL” and marked crosswalks. Reserved bus or transit lanes are marked by signs and lane text such as “BUS ONLY” or “BUS LANE”; obey any posted operating hours, and do not drive, stop, or park in them except where signs permit entry, usually for turns. In states with yield‑to‑bus laws, let a bus signal and re‑enter traffic from a signed stop.
\end{lstlisting}

\paragraph{China, Mainland (CN).}~\\
\lstinputlisting[basicstyle=\ttfamily\scriptsize, breaklines=true, frame=single, backgroundcolor=\color{backcolour}]{table/prompts/handbook_cn.txt}

\paragraph{Japan (JP).}~\\
\lstinputlisting[basicstyle=\ttfamily\scriptsize, breaklines=true, frame=single, backgroundcolor=\color{backcolour}]{table/prompts/handbook_jp.txt}

\paragraph{Singapore (SG).}~\\
\lstinputlisting[basicstyle=\ttfamily\scriptsize, breaklines=true, frame=single, backgroundcolor=\color{backcolour}]{table/prompts/handbook_sg.txt}

\paragraph{India (IND).}~\\
\lstinputlisting[basicstyle=\ttfamily\scriptsize, breaklines=true, frame=single, backgroundcolor=\color{backcolour}]{table/prompts/handbook_ind.txt}

\subsection{Counterfactual Verification Protocol}
\label{sec:appendix-counterfactual}

A central design goal of \bench is that every retained QA pair must genuinely depend on region-specific reasoning rather than on universal driving common sense. To turn this requirement into an operational filter, every candidate QA pair undergoes a counterfactual verification pass before it can enter the benchmark. The pass holds the visual scene and the question text fixed and substitutes, one at a time, the homologous rule snippet from each of the other five countries in our handbook. For every substitution, the verifier is asked to (i)~apply the substituted rule---rather than the original country's rule---to the scene, and (ii)~select the option that becomes correct under that rule. A candidate is retained only if the resulting correct option \emph{differs from the original ground-truth answer under at least one of the five substitutions}, that is, if there exists a counterfactual country in which the same scene admits a different correct action. Items whose answer is invariant under every substitution are dropped, since they reflect universal driving conventions rather than region-specific reasoning.

This protocol provides a strong guarantee against a common failure mode in cross-cultural benchmarks: questions that nominally test cultural knowledge but are in fact resolvable by generic driving priors. By requiring at least one counterfactual to flip the answer, we ensure that surface visual features alone cannot resolve any item without engagement with the specific local rule. The verification pass is performed by a strong reasoning VLM conditioned on the structured scene state and each candidate rule snippet, with its per-country verdicts recorded for downstream analysis.

\paragraph{Verifier prompt.}~\\
\begin{lstlisting}[basicstyle=\ttfamily\scriptsize, breaklines=true, frame=single, backgroundcolor=\color{backcolour}]
You are a verifier for a culture-aware driving benchmark. Your task is
to decide whether a candidate (image, question, options) pair is genuinely
"culturally divergent" -- i.e., whether the correct option would actually
change if the SAME visual scene were governed by a DIFFERENT country's
traffic rules.

Inputs:
  scene_image       : the original driving frame
  scene_state       : structured description of the scene
  question          : the candidate question
  options           : multiple-choice options A/B/C/D
  origin_country    : the country whose rules generated the original GT
  origin_rule       : the cited rule snippet from {origin_country}'s handbook
  origin_gt         : the originally assigned correct option letter
  candidate_country : one of the other 5 countries
  candidate_rule    : the homologous rule snippet from {candidate_country}'s handbook

Procedure:
  1. Read scene_state to confirm what is actually visible.
  2. Apply candidate_rule (NOT origin_rule) to the scene.
  3. Pick the option that becomes correct under candidate_rule.
  4. Compare against origin_gt.

Output STRICT JSON, no commentary:
{
  "answer_under_candidate" : "A | B | C | D",
  "differs_from_origin"    : true | false,
  "reason"                 : "<one-sentence rationale grounded in candidate_rule>"
}

Decision: a candidate QA pair is RETAINED if `differs_from_origin == true`
for at least one of the five other countries (i.e., the scene admits at
least one cross-country answer flip). Items whose answer is invariant
under every substitution are DROPPED, since they reduce to universal
driving common sense rather than region-specific reasoning.
\end{lstlisting}

\subsection{Human Verification with Annotation Tool}
\label{sec:appendix-annotation-tool}

\begin{figure}[h]
    \centering
    \includegraphics[width=0.9\linewidth]{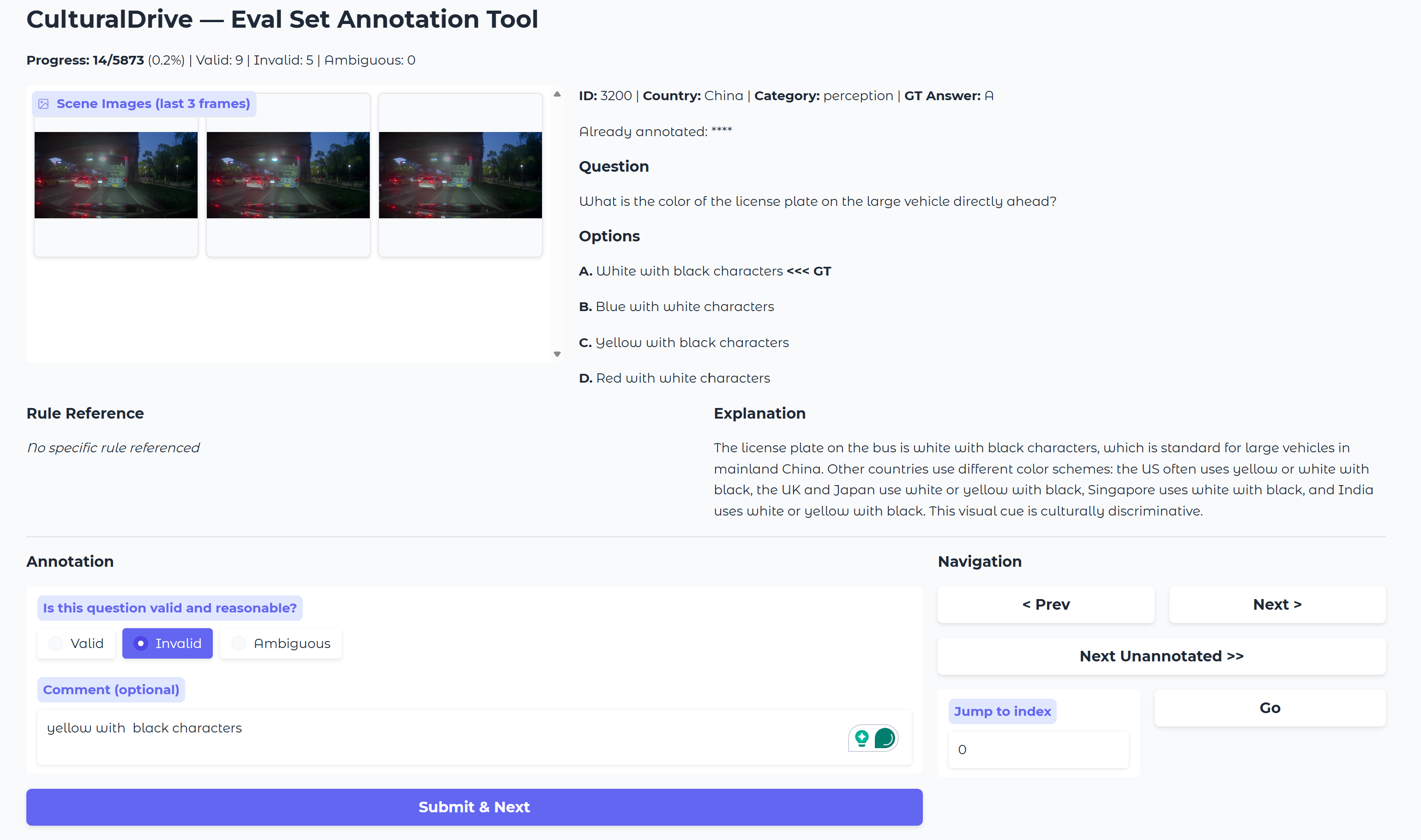}
    \caption{Web-based annotation tool used for human review.}
    \label{fig:annotation-tool}
\end{figure}

After counterfactual verification and basic quality filtering, the remaining candidate pool still contains two failure modes that surface only on careful inspection of the image: \textbf{(i)~ground-truth errors}, in which the QA-generation pipeline assigned an incorrect reference letter---most often on questions about color, sign identification, or signal state---and \textbf{(ii)~visual under-specification}, in which the available evidence does not uniquely determine a single option. To address both, we apply a two-stage human-calibrated verification protocol: a strong language-model judge first audits every candidate, after which human reviewers adjudicate the borderline verdicts through a custom web-based annotation tool (Figure~\ref{fig:annotation-tool}).

The tool is designed to make each item fully self-contained for the reviewer. The interface displays the last three frames of the driving scene alongside the cited rule reference, the country and task category, the four options with the proposed ground-truth letter highlighted, and the model's accompanying explanation. The reviewer marks each item as \textsc{Valid}, \textsc{Invalid}, or \textsc{Ambiguous}, and may attach a free-text comment recording a corrected option when the proposed answer is judged wrong. Borderline \textsc{Ambiguous} cases---those where visual evidence is partial but a defensible single answer remains recoverable---are retained, while clearly invalid items are dropped. Any corrected-answer suggestions from the reviewers are preserved for potential re-labeling in future versions of the benchmark.

\paragraph{Judge prompt.}~\\
\begin{lstlisting}[basicstyle=\ttfamily\scriptsize, breaklines=true, frame=single, backgroundcolor=\color{backcolour}]
You are an LLM-as-judge auditing a culture-aware driving benchmark for
ground-truth correctness. For each candidate (image, question, options,
GT_answer) you must:

  1. Look at the image carefully.
  2. Decide whether the provided GT_answer is supported by the image
     under the country-specific traffic context.
  3. Output a JSON record with the verdict, your confidence, a one-paragraph
     rationale, and (for INCORRECT verdicts) the option you believe is
     actually correct.

Inputs:
  scene_image  : the candidate driving frame
  country      : the country governing the rules for this item
  question     : the multiple-choice question
  options      : A / B / C / D options
  gt_answer    : the originally labeled correct option letter

Output STRICT JSON, no commentary:
{
  "verdict"       : "CORRECT | INCORRECT | AMBIGUOUS | PARSE_ERROR",
  "confidence"    : <float in [0, 1]>,
  "reason"        : "<one-paragraph rationale referencing visible evidence>",
  "suggested_fix" : "<the option the judge believes is correct, or null>"
}

Verdict semantics:
  CORRECT    : GT is correct given the image and country rules.
  INCORRECT  : GT is wrong; suggest a different option in `suggested_fix`.
  AMBIGUOUS  : Multiple options are plausible / image evidence insufficient.
  PARSE_ERROR: The question or options are malformed.
\end{lstlisting}

\paragraph{Outcome statistics.}
Table~\ref{tab:audit-funnel} summarizes the construction funnel from raw candidate to released benchmark. The QA-generation pipeline produced roughly 17{,}000 candidates; counterfactual and basic-quality filtering reduces this to 5{,}873 items. The language-model judge then labels 84.5\% of these as correct, 11.0\% as incorrect, and 4.3\% as ambiguous. Human review through the annotation tool retains all but the clearly incorrect items, yielding the final 5{,}053 QA pairs released in the benchmark. Among the items flagged as incorrect, the dominant failure modes are misreads of \emph{traffic-light or signal state} (42\%), \emph{road-sign identification} (18\%), \emph{lane count or lane markings} (14\%), and \emph{pedestrian or crosswalk presence and priority} (11\%); the remainder spans license-plate color, turning rules, and miscellaneous object identification. This breakdown indicates that the residual error mode at the end of our pipeline is dominated by fine-grained perceptual misreads rather than rule-application errors, consistent with the broader observation throughout the paper that perceptual grounding remains a non-trivial bottleneck even after rule knowledge is internalized.

\begin{table}[h]
\centering
\small
\setlength{\tabcolsep}{6pt}
\caption{Construction funnel from raw QA candidate to released benchmark. Verdict distribution rows correspond to the language-model judge audit; the final retained set additionally incorporates human review of borderline cases through the annotation tool.}
\label{tab:audit-funnel}
\begin{tabular}{lr}
\toprule
Stage & Count \\
\midrule
Raw candidates from the QA-generation pipeline                    & $\sim$17{,}000 \\
After counterfactual verification and quality filtering           & 5{,}873 \\
\midrule
\multicolumn{2}{l}{\emph{Language-model judge verdicts ($n = 5{,}873$)}} \\
\quad\textsc{Correct}                                             & 4{,}961 (84.5\%) \\
\quad\textsc{Incorrect}                                           & 643 (11.0\%) \\
\quad\textsc{Ambiguous}                                           & 254 (4.3\%) \\
\quad\textsc{Parse / request error}                               & 15 (0.3\%) \\
\midrule
Final retained after human review                                 & \textbf{5{,}053} \\
\bottomrule
\end{tabular}
\end{table}

% =============================================================================
\section{Extended Case Studies}
\label{sec:appendix-cases}
% =============================================================================

% Requires: \usepackage{tcolorbox}\tcbuselibrary{breakable,skins}
% Image paths assume the six case images live under ./misc/ alongside
% annotation_tool.png; adjust if your main paper uses a different
% asset directory.

We complement Fig.~6 in the main paper with six additional qualitative
comparisons between InternVL3-8B (base) and \method$^{\ddagger}$ under
the Reasoning setting. Each case fixes the (image, question, options)
triplet and shows the model's full 4-step chain-of-thought followed by
the predicted letter. The selected cases span three cultural cues that
do \emph{not} appear in the main-paper figure---license-plate color
schemes (Cases~1--3), all-way stop control (Case~4), and school-zone
sign conventions in countries other than the four shown in
Fig.~6 (Cases~5--6). Across all six cases the base model identifies the
correct country from visual cues but invokes an incorrect or generic
rule, while \method$^{\ddagger}$ explicitly cites the relevant
handbook section (S5, S17, or S18) and recovers the right answer.

\tcbset{
    casebox/.style={
        enhanced jigsaw, breakable,
        colback=gray!2!white, colframe=gray!55!black,
        coltitle=white, colbacktitle=gray!55!black,
        fonttitle=\bfseries, sharp corners=south,
        boxrule=0.4pt, left=4pt, right=4pt, top=4pt, bottom=4pt,
    },
    basebox/.style={
        colback=red!3!white, colframe=red!45!black,
        coltitle=white, colbacktitle=red!45!black,
        fonttitle=\bfseries\small, fontupper=\scriptsize,
        sharp corners=south, boxrule=0.4pt,
    },
    sdftbox/.style={
        colback=green!3!white, colframe=green!45!black,
        coltitle=white, colbacktitle=green!45!black,
        fonttitle=\bfseries\small, fontupper=\scriptsize,
        sharp corners=south, boxrule=0.4pt,
    },
}

% ---------------------------------------------------------------- Case 1 ----
\begin{tcolorbox}[casebox,
    title={Case~1: License plate appearance --- China (id=90004)}]
\begin{minipage}[t]{0.40\linewidth}
\vspace{0pt}
\includegraphics[width=\linewidth]{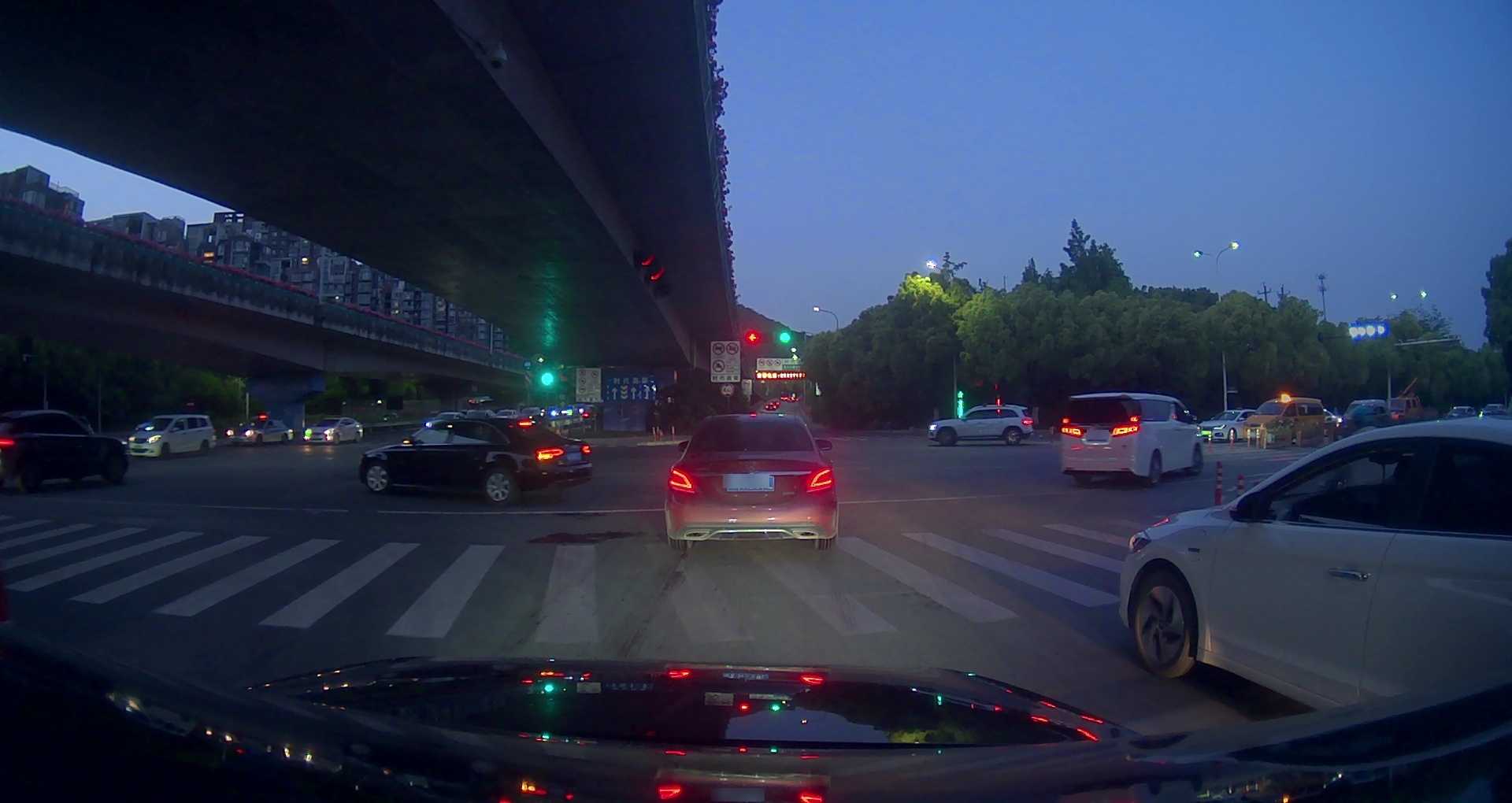}
\end{minipage}\hfill
\begin{minipage}[t]{0.57\linewidth}\vspace{0pt}\small
\textbf{Question.} For an ordinary private passenger car, which
registration-plate appearance is standard in the country shown?\\[2pt]
\textbf{Options.}\\
\hspace*{1em}A. Black plates with white/silver characters.\\
\hspace*{1em}B. No single national pattern; varies by state.\\
\hspace*{1em}C. White plates with green characters.\\
\hspace*{1em}D. Blue plates with white characters. \hfill\textbf{(GT)}
\end{minipage}

\smallskip
\begin{tcolorbox}[basebox, title={InternVL3-8B (base) --- predicted A \(\boldsymbol{\times}\)}]
\textbf{Step 1 (Geo).} Urban Asian setting; license plates appear white
with black characters; consistent with several East-Asian countries
including China.\\
\textbf{Step 2 (Rule).} In China, the standard for plates of private
passenger cars is white with black characters.\\
\textbf{Step 3 (Visual).} Plates in the image are white with black
characters, matching the recalled rule.\\
\textbf{Step 4 (Answer).} A.
\end{tcolorbox}

\begin{tcolorbox}[sdftbox, title={\method$^{\ddagger}$ --- predicted D \(\boldsymbol{\checkmark}\)}]
\textbf{Step 1 (Rule).} The relevant handbook section is \textbf{S17},
which describes vehicle registration plates for ordinary privately owned
small cars in the country.\\
\textbf{Step 2 (Visual).} Vehicles in the image carry \emph{blue} plates
with white characters on both front and rear.\\
\textbf{Step 3 (Apply).} S17 specifies blue plates with white characters
for ordinary privately owned small cars; the visible plates match.\\
\textbf{Step 4 (Answer).} D. Blue plates with white characters.
\end{tcolorbox}
\end{tcolorbox}

% ---------------------------------------------------------------- Case 2 ----
\begin{tcolorbox}[casebox,
    title={Case~2: License plate appearance --- United States (id=90222)}]
\begin{minipage}[t]{0.40\linewidth}\vspace{0pt}
\includegraphics[width=\linewidth]{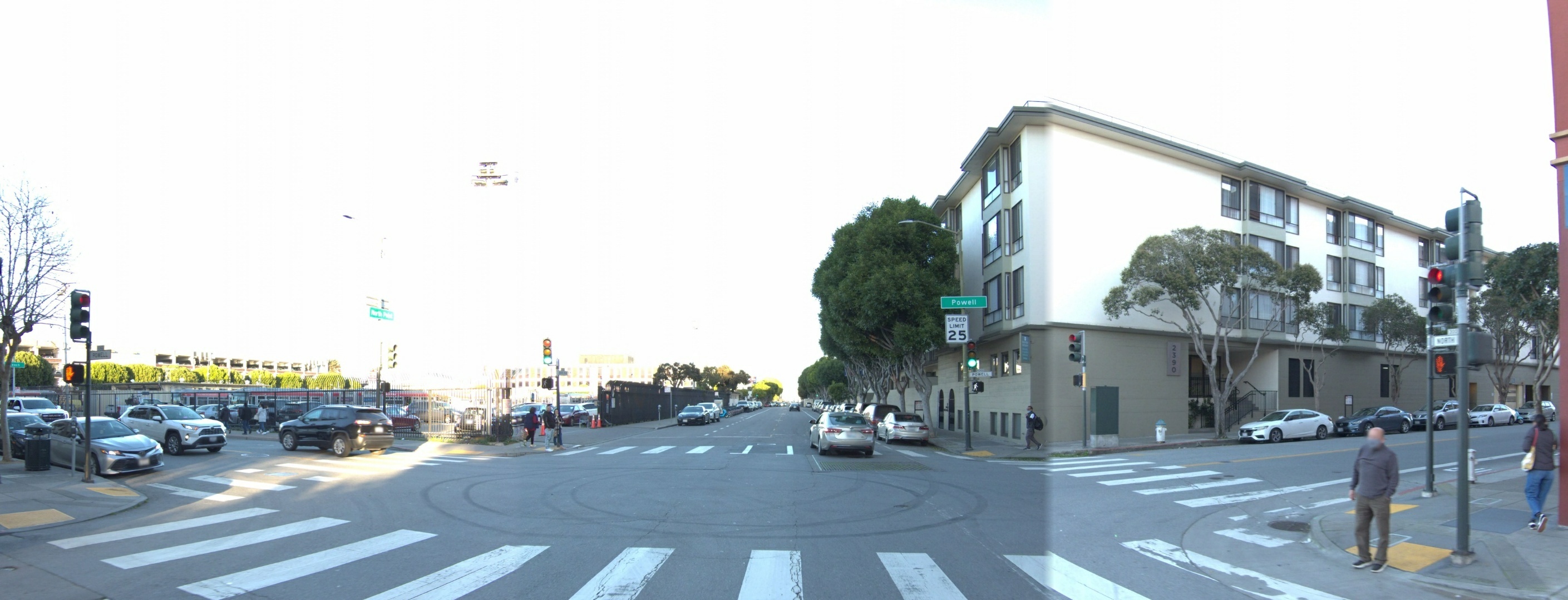}
\end{minipage}\hfill
\begin{minipage}[t]{0.57\linewidth}\vspace{0pt}\small
\textbf{Question.} (same as Case~1.)\\[2pt]
\textbf{Options.}\\
\hspace*{1em}A. No single national pattern; varies by state. \hfill\textbf{(GT)}\\
\hspace*{1em}B. Black plates with white/silver characters.\\
\hspace*{1em}C. White front, yellow rear, black characters.\\
\hspace*{1em}D. White plates with green characters.
\end{minipage}

\smallskip
\begin{tcolorbox}[basebox, title={InternVL3-8B (base) --- predicted B \(\boldsymbol{\times}\)}]
\textbf{Step 1 (Geo).} English signage (``Powell'', ``Speed Limit 25''),
North-American urban infrastructure $\Rightarrow$ United States.\\
\textbf{Step 2 (Rule).} ``In the United States $\ldots$ both front and
rear license plates. Typically, these plates are black with white or
silver characters.''\\
\textbf{Step 3 (Visual).} Plates not clearly visible; defaults to the
recalled rule.\\
\textbf{Step 4 (Answer).} B.
\end{tcolorbox}

\begin{tcolorbox}[sdftbox, title={\method$^{\ddagger}$ --- predicted A \(\boldsymbol{\checkmark}\)}]
\textbf{Step 1 (Rule).} \textbf{S17}: ``the U.S.\ has no nationwide
private-car plate design or numbering system; plates are issued by each
state and the District of Columbia, with variations in colors,
slogans, and formats.''\\
\textbf{Step 2 (Visual).} Urban U.S.\ scene; no specific plate clearly
visible, but no single colour scheme is implied by the context.\\
\textbf{Step 3 (Apply).} Per S17, many states display only a rear plate
and designs differ by state.\\
\textbf{Step 4 (Answer).} A.
\end{tcolorbox}
\end{tcolorbox}

% ---------------------------------------------------------------- Case 3 ----
\begin{tcolorbox}[casebox,
    title={Case~3: License plate appearance --- India (id=91050)}]
\begin{minipage}[t]{0.40\linewidth}\vspace{0pt}
\includegraphics[width=\linewidth]{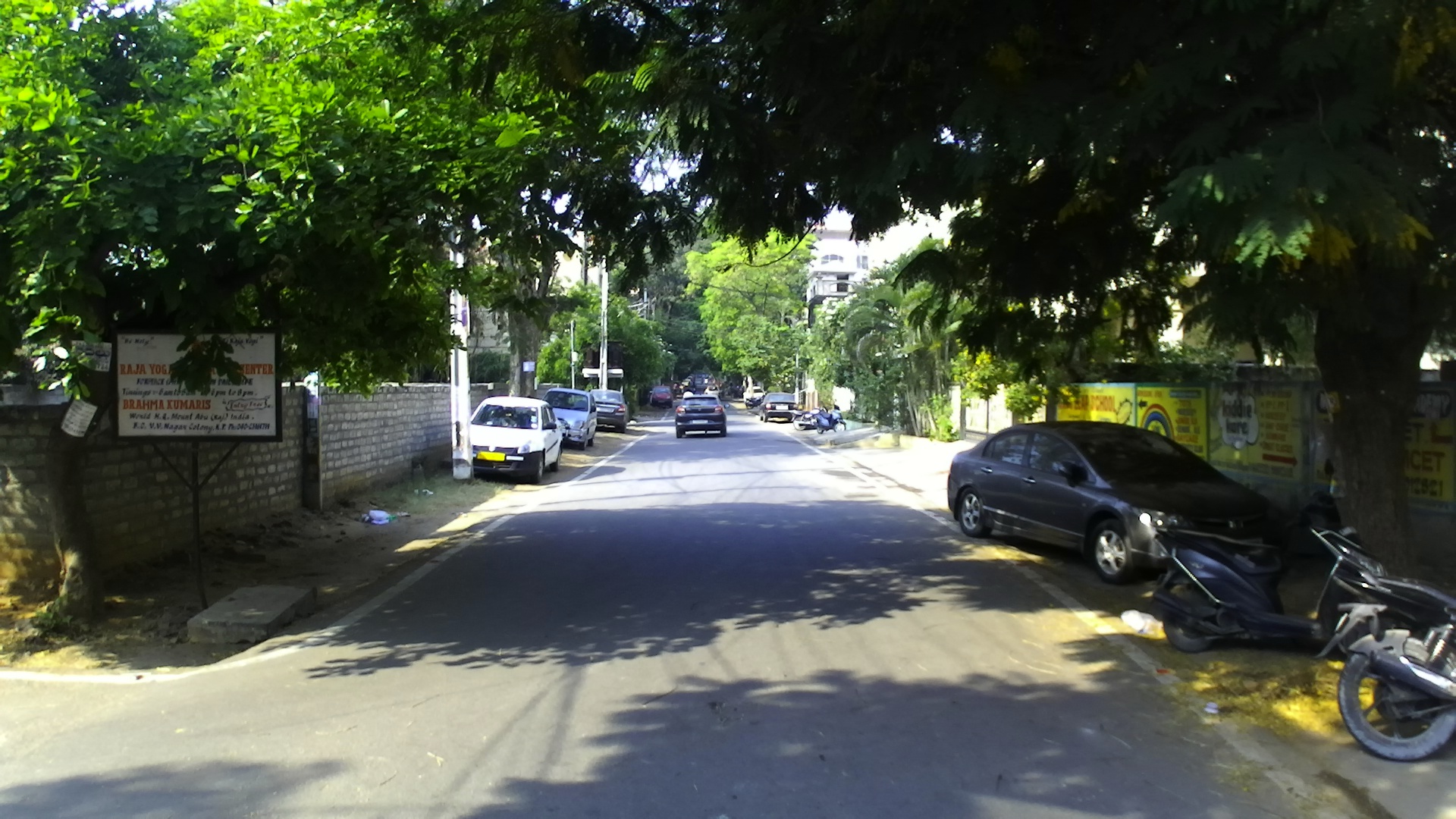}
\end{minipage}\hfill
\begin{minipage}[t]{0.57\linewidth}\vspace{0pt}\small
\textbf{Question.} (same as Case~1.)\\[2pt]
\textbf{Options.}\\
\hspace*{1em}A. White front, yellow rear, black characters.\\
\hspace*{1em}B. No single national pattern; varies by state.\\
\hspace*{1em}C. White plates with black characters for private vehicles. \hfill\textbf{(GT)}\\
\hspace*{1em}D. White plates with green characters.
\end{minipage}

\smallskip
\begin{tcolorbox}[basebox, title={InternVL3-8B (base) --- predicted A \(\boldsymbol{\times}\)}]
\textbf{Step 1 (Geo).} Trees, English signage, vehicle layout $\Rightarrow$
India. Plates appear white with black characters.\\
\textbf{Step 2 (Rule).} ``In India, private cars typically use white
plates with black characters.''\\
\textbf{Step 3 (Visual).} Plates in the image are indeed white with
black characters $\Rightarrow$ the front-white/rear-yellow option does
\emph{not} fit.\\
\textbf{Step 4 (Answer).} The reasoning oscillates and ultimately
defaults to A despite explicitly noting the contradiction.
\end{tcolorbox}

\begin{tcolorbox}[sdftbox, title={\method$^{\ddagger}$ --- predicted C \(\boldsymbol{\checkmark}\)}]
\textbf{Step 1 (Rule).} \textbf{S17}: private non-transport cars use
black characters on a white retro-reflective background, both front
and rear.\\
\textbf{Step 2 (Visual).} Plates of parked vehicles are white with
black characters, matching S17.\\
\textbf{Step 3 (Apply).} The yellow-rear pattern is reserved for
commercial / transport vehicles, not private cars.\\
\textbf{Step 4 (Answer).} C.
\end{tcolorbox}
\end{tcolorbox}

% ---------------------------------------------------------------- Case 4 ----
\begin{tcolorbox}[casebox,
    title={Case~4: All-way stop intersections --- United States (id=90215)}]
\begin{minipage}[t]{0.40\linewidth}\vspace{0pt}
\includegraphics[width=\linewidth]{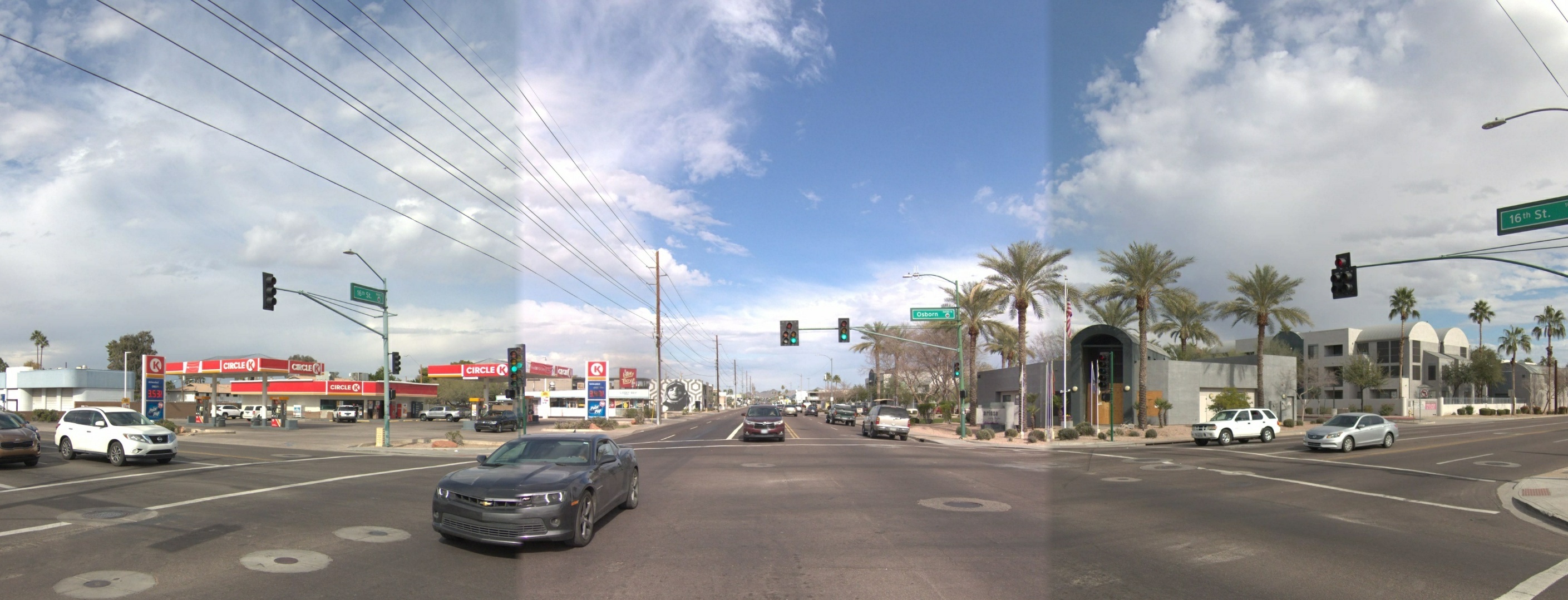}
\end{minipage}\hfill
\begin{minipage}[t]{0.57\linewidth}\vspace{0pt}\small
\textbf{Question.} Which intersection control is especially common in
the country shown, with every approach required to stop?\\[2pt]
\textbf{Options.}\\
\hspace*{1em}A. All-way (4-way) stop intersections; first to arrive
proceeds first. \hfill\textbf{(GT)}\\
\hspace*{1em}B. Four-way stops uncommon; rely on signals or give-way.\\
\hspace*{1em}C. Four-way stops uncommon; signals, stop signs, or
right-priority.\\
\hspace*{1em}D. Four-way stops uncommon; signals or yield signs.
\end{minipage}

\smallskip
\begin{tcolorbox}[basebox, title={InternVL3-8B (base) --- predicted D \(\boldsymbol{\times}\)}]
\textbf{Step 1 (Geo).} Circle K, palm trees, English signage
$\Rightarrow$ U.S.\ (likely southwest).\\
\textbf{Step 2 (Rule).} ``In the U.S., traffic lights are commonly used
at intersections; many intersections rely on signals.''\\
\textbf{Step 3 (Visual).} Signalized intersection in view; no stop
signs visible.\\
\textbf{Step 4 (Answer).} D.
\end{tcolorbox}

\begin{tcolorbox}[sdftbox, title={\method$^{\ddagger}$ --- predicted A \(\boldsymbol{\checkmark}\)}]
\textbf{Step 1 (Rule).} \textbf{S5}: all-way stop intersections require
every approach to come to a full stop; the first to arrive proceeds
first, with priority to vehicles on the right in case of a tie.\\
\textbf{Step 2 (Visual).} Standard urban intersection layout consistent
with the U.S.\\
\textbf{Step 3 (Apply).} All-way stops are a hallmark U.S.\ control,
distinct from signal-only or yield-only control.\\
\textbf{Step 4 (Answer).} A.
\end{tcolorbox}
\end{tcolorbox}

% ---------------------------------------------------------------- Case 5 ----
\begin{tcolorbox}[casebox,
    title={Case~5: School warning sign --- United States (id=90242)}]
\begin{minipage}[t]{0.40\linewidth}\vspace{0pt}
\includegraphics[width=\linewidth]{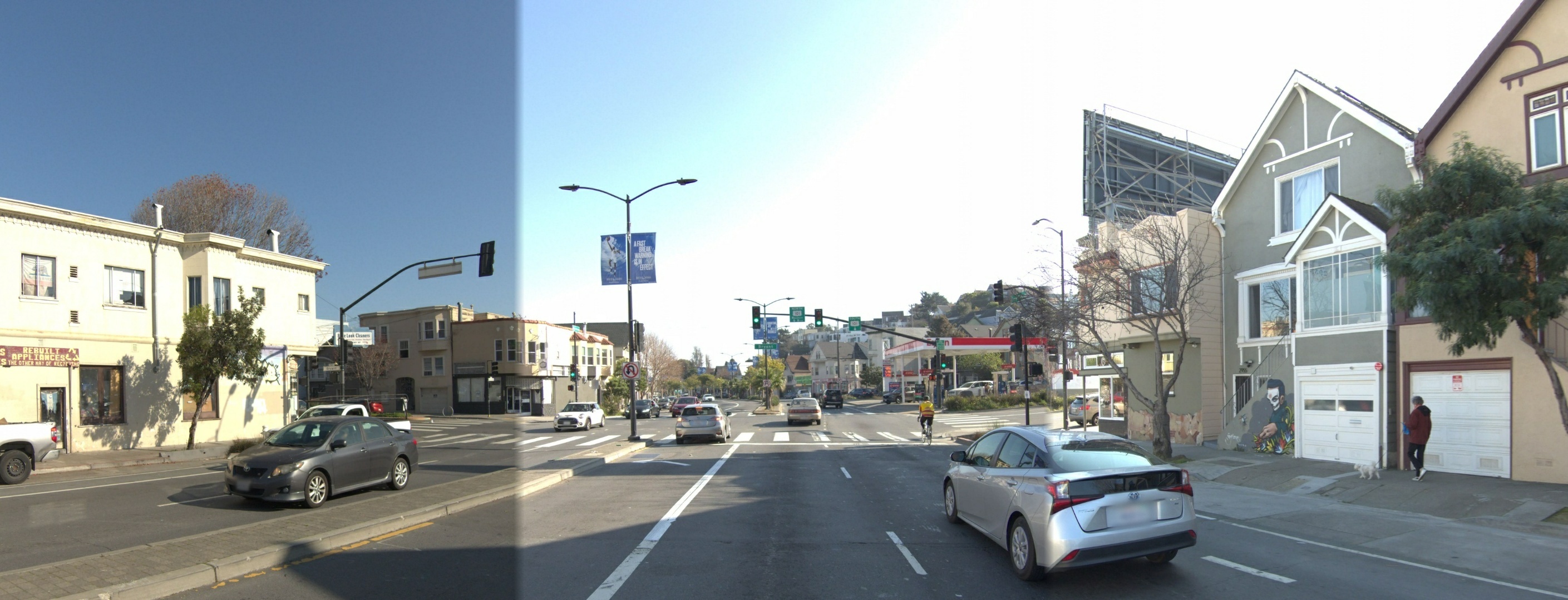}
\end{minipage}\hfill
\begin{minipage}[t]{0.57\linewidth}\vspace{0pt}\small
\textbf{Question.} What sign format warns that a school or children
are ahead in this country?\\[2pt]
\textbf{Options.}\\
\hspace*{1em}A. Red-bordered triangle with a children symbol.\\
\hspace*{1em}B. Yellow triangle with a black children symbol.\\
\hspace*{1em}C. Pentagon-shaped sign, fluorescent yellow-green. \hfill\textbf{(GT)}\\
\hspace*{1em}D. Yellow diamond with a black children/school symbol.
\end{minipage}

\smallskip
\begin{tcolorbox}[basebox, title={InternVL3-8B (base) --- predicted D \(\boldsymbol{\times}\)}]
\textbf{Step 1 (Geo).} Urban U.S.\ signage and infrastructure.\\
\textbf{Step 2 (Rule).} Default ``yellow diamond with black symbol''
template, recalled as the canonical U.S.\ warning sign.\\
\textbf{Step 3 (Visual).} The model imagines a yellow diamond children
sign; the actual school sign template is not retrieved.\\
\textbf{Step 4 (Answer).} D.
\end{tcolorbox}

\begin{tcolorbox}[sdftbox, title={\method$^{\ddagger}$ --- predicted C \(\boldsymbol{\checkmark}\)}]
\textbf{Step 1 (Rule).} \textbf{S18} (sign system \& visual cues):
school / pedestrian-school warning signs in the U.S.\ are pentagon
shaped and fluorescent yellow-green, distinct from the generic yellow
diamond used for most other warnings.\\
\textbf{Step 2 (Visual).} No school sign in this particular frame, but
the question fixes the country.\\
\textbf{Step 3 (Apply).} Pentagon + fluorescent yellow-green
$\Rightarrow$ option C.\\
\textbf{Step 4 (Answer).} C.
\end{tcolorbox}
\end{tcolorbox}

% ---------------------------------------------------------------- Case 6 ----
\begin{tcolorbox}[casebox,
    title={Case~6: School warning sign --- Singapore (id=90869)}]
\begin{minipage}[t]{0.40\linewidth}\vspace{0pt}
\includegraphics[width=\linewidth]{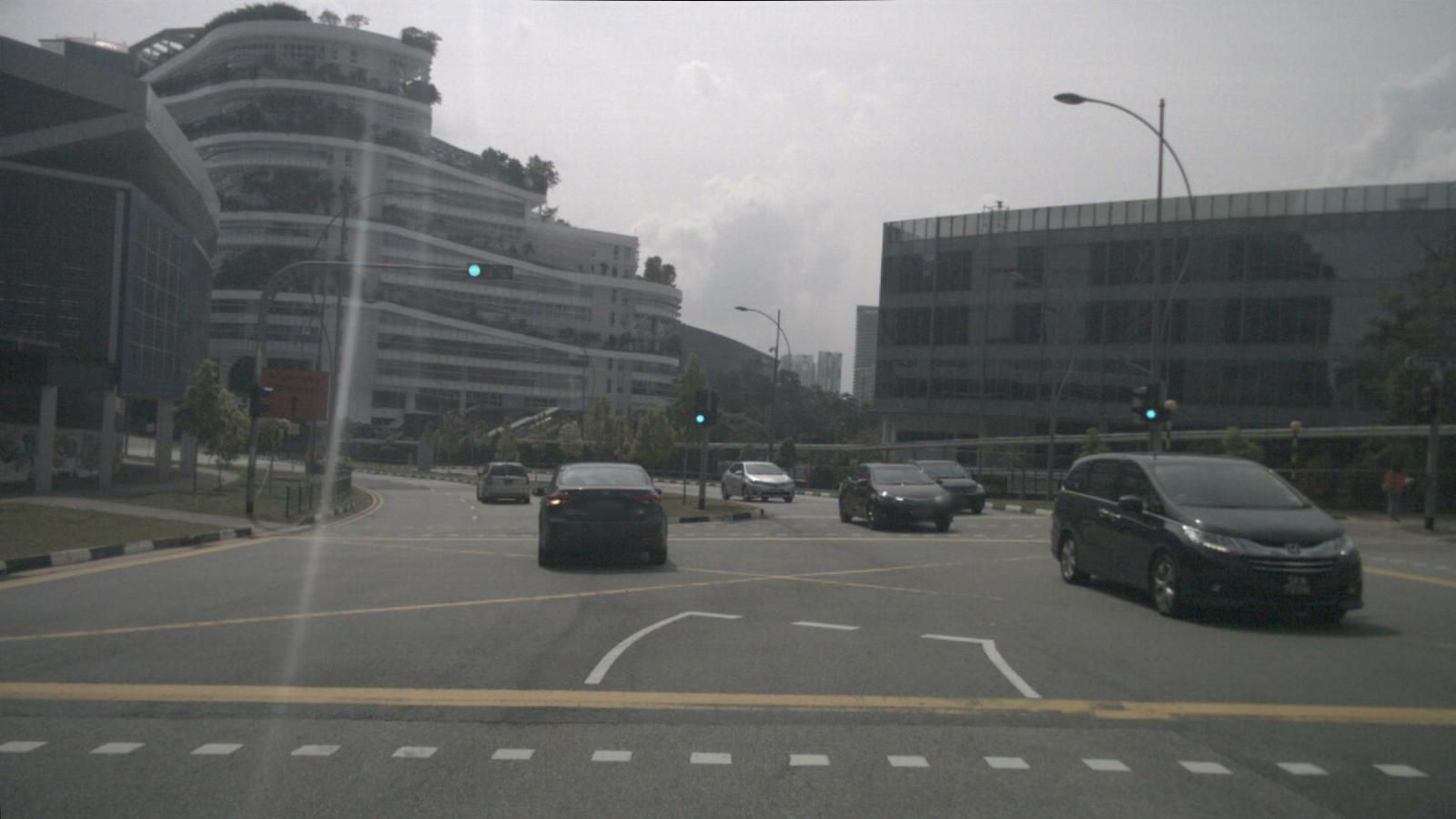}
\end{minipage}\hfill
\begin{minipage}[t]{0.57\linewidth}\vspace{0pt}\small
\textbf{Question.} What sign format warns that a school or children
are ahead in this country? (same as Case~5.)\\[2pt]
\textbf{Options.}\\
\hspace*{1em}A. Red-bordered triangle with a children symbol. \hfill\textbf{(GT)}\\
\hspace*{1em}B. Yellow triangle with a black children symbol.\\
\hspace*{1em}C. Pentagon-shaped sign, fluorescent yellow-green.\\
\hspace*{1em}D. Yellow diamond with a black children/school symbol.
\end{minipage}

\smallskip
\begin{tcolorbox}[basebox, title={InternVL3-8B (base) --- predicted D \(\boldsymbol{\times}\)}]
\textbf{Step 1 (Geo).} ``Modern urban setting $\ldots$ likely Asian
country.''\\
\textbf{Step 2 (Rule).} ``In many Asian countries (including
Singapore), warning signs are yellow with black symbols.''\\
\textbf{Step 3 (Visual).} Imagined yellow diamond template;
Vienna-Convention red-triangle convention not retrieved.\\
\textbf{Step 4 (Answer).} D.
\end{tcolorbox}

\begin{tcolorbox}[sdftbox, title={\method$^{\ddagger}$ --- predicted A \(\boldsymbol{\checkmark}\)}]
\textbf{Step 1 (Rule).} \textbf{S18}: Singapore school-area warnings
follow the Vienna Convention---red-bordered triangle on white
background with a children symbol.\\
\textbf{Step 2 (Visual).} No school sign visible in the frame; the
country fixes the convention.\\
\textbf{Step 3 (Apply).} Red-bordered triangle $\Rightarrow$ option A.\\
\textbf{Step 4 (Answer).} A.
\end{tcolorbox}
\end{tcolorbox}

\section{Broader Impact}
\label{sec:appendix-broader-impact}

This work focuses on evaluating and improving the geo-cultural reasoning capabilities of vision-language models in autonomous driving, with implications that extend beyond benchmark performance. \looseness=-1

\paragraph{Toward globally deployable driving foundation models.}
Most current driving VLMs are trained and evaluated on data from a small number of regions, predominantly the United States and Singapore, and our results show that this concentration produces models with substantial blind spots when deployed in regions whose visual cues and traffic conventions diverge from common pretraining priors. As VLM-based driving systems progress toward real-world deployment across countries with distinct traffic codes---left- versus right-hand traffic, country-specific sign conventions, and divergent right-of-way rules---the inability to ground decisions in local rules is no longer a benchmarking curiosity but a safety concern. \bench surfaces this failure mode quantitatively, and \method demonstrates that it can be partially mitigated through rule-conditioned self-distillation, providing a concrete path toward foundation models that generalize across regional traffic systems rather than overfitting to a dominant one.

\paragraph{Implications beyond autonomous driving.}
The pattern we identify---models that recognize a scene correctly but fail to apply the appropriate region-specific rule---is unlikely to be unique to driving. Any embodied or safety-critical VLM application that operates across jurisdictions or cultural contexts (medical guidelines, legal compliance assistants, robotic systems deployed across regions) faces an analogous challenge: surface visual or linguistic competence can mask a failure to ground decisions in the locally applicable norms. Our finding that unconstrained chain-of-thought reasoning can amplify rather than correct such failures is particularly relevant to deployment scenarios where reasoning traces may be mistaken for evidence of correctness. We hope \bench and \method encourage the broader community to design benchmarks and training procedures that explicitly probe contextual rule grounding, rather than aggregate accuracy alone.

\paragraph{Risks and responsible-use considerations.}
The benchmark itself is constructed from publicly released driving datasets under their original licenses, with no introduction of new identifying information beyond what is already present in the source data. We do not release human trajectories or any data that would expose individual drivers. Two responsible-use considerations nevertheless apply. First, models trained or evaluated on \bench should not be interpreted as certified for deployment in any jurisdiction; high accuracy on a multiple-choice benchmark does not equate to operational safety on real roads, and our results explicitly show that even our best models retain non-trivial visual perception errors. Second, the per-country traffic-rule handbooks we release reflect publicly available regulations as compiled at the time of dataset construction; users intending to deploy systems in production should verify the rules against current official sources, since traffic codes are periodically updated.

% \clearpage
% \input{checklist.tex}

\end{document}